\documentclass{article} 
\usepackage{iclr2027_conference,times}
\iclrfinalcopy

\usepackage{amsmath,amsfonts,bm}

\def\eqref#1{equation~\ref{#1}}

\def\1{\bm{1}}

\DeclareMathAlphabet{\mathsfit}{\encodingdefault}{\sfdefault}{m}{sl}
\SetMathAlphabet{\mathsfit}{bold}{\encodingdefault}{\sfdefault}{bx}{n}

\usepackage{url}
\usepackage{graphicx}
\usepackage{booktabs}
\usepackage{array}
\usepackage{multirow}
\usepackage{xcolor}
\usepackage{algorithm}
\usepackage{algpseudocode}
\usepackage{float}
\usepackage{hyperref}
\hypersetup{
  colorlinks=false,
  pdfborder={0 0 1},
  linkbordercolor={1 0 0},
  citebordercolor={0 1 0},
  urlbordercolor={0 1 1}
}
\definecolor{sandyellow}{RGB}{254,217,205}
\newcommand{\sand}[1]{\colorbox{sandyellow}{\makebox[1.45cm][c]{#1}}}
\newcommand{\rate}[2]{#1{\scriptsize$\,\pm\,#2$}}

\newcommand{\methodname}{ECL}

\title{Evidence-Guided Counterfactual Learning for Long-Tailed Robotic Policies}
\author{
Weiwei Chen$^{1}$ \quad Chenghao Xu$^{1}$ \quad Meiling Yu$^{1}$ \quad
Xu Yang$^{1}$ \quad Cheng Deng$^{1}$\thanks{Corresponding author.} \\
$^{1}$Hohai University \\
{\scriptsize\href{https://krlozhilderman287-creator.github.io/EVIDENCE-GUIDED-COUNTERFACTUAL-LEARNING-FOR-LONG-TAILED-ROBOTIC-POLICIES/}{\texttt{https://krlozhilderman287-creator.github.io/}}} \\
{\scriptsize\href{https://krlozhilderman287-creator.github.io/EVIDENCE-GUIDED-COUNTERFACTUAL-LEARNING-FOR-LONG-TAILED-ROBOTIC-POLICIES/}{\texttt{EVIDENCE-GUIDED-COUNTERFACTUAL-LEARNING-FOR-LONG-TAILED-ROBOTIC-POLICIES/}}}
}

\begin{document}
\raggedbottom

\maketitle

\begin{abstract}
In the practical deployment of vision--language--action policies, handling long-tailed task distributions is a common challenge due to the highly imbalanced composition of robot demonstration datasets. To investigate this issue, we study the head-task action bias in shared robot policies, particularly focusing on its relationship with task-specific visual evidence. Our experimental analysis shows that, when the visual evidence required by a tail task is insufficient, the policy increasingly falls back on action patterns reinforced by data-rich head tasks, resulting in degraded performance on under-represented tasks. Motivated by this observation, we propose Evidence-Guided Counterfactual Learning, a training framework for long-tailed robot policies. Our method identifies decision-relevant visual evidence using instruction-related and action-related signals, and constructs counterfactual observations by removing the selected evidence. The resulting counterfactual predictions are used to strengthen the association among task instructions, visual evidence, and actions, with the supervision adaptively adjusted according to the number of demonstrations available for each task. The proposed framework requires neither additional training data nor region-level annotations and incurs no additional computation during deployment. Extensive experiments across simulated benchmarks, different VLA architectures, and real-world robot platforms demonstrate that our method robustly and consistently improves long-tailed robot policy learning, particularly for tail tasks, while maintaining reliable performance on head tasks.
\end{abstract}

\section{Introduction}

Vision--language--action models (VLAs) have emerged as a promising foundation for generalist robot policies, integrating visual perception, language understanding, and action generation within a unified framework \citep{brohan2023rt,kim2024openvla,black2025pi0,MaSZHK26}. Robot demonstration datasets, however, can be highly imbalanced across manipulation skills \citep{openx2024}. Similar to the long-tailed category distributions observed in datasets such as iNaturalist \citep{vanhorn2018inaturalist}, robot task distributions can be skewed toward a few data-rich head tasks, leaving many tail tasks with limited demonstrations\citep{zhu2026beyond}.

Long-tailed learning addresses unequal supervision through approaches such as resampling, loss reweighting, and decoupled representation and classifier learning \citep{cui2019classbalanced,kang2020decoupling,ma2022delving}. However, most of these methods are developed for classification and are difficult to apply directly to VLAs due to differences in task formulation and model architecture.\citet{zhu2026beyond} propose Approaching-Phase Augmentation (APA) for long-tailed VLA learning, but it relies on additional tools, such as object detection and image inpainting, increasing the complexity of the approach.Moreover, APA primarily mitigates the problem through data augmentation, without directly constraining VLAs' reliance on head-task action bias.

However, comparing the results of policies trained on the balanced FULL dataset and the long-tailed LT dataset (see Appendix~\ref{app:dataset_distribution}), we find that tail-task performance drops substantially under LT, as shown in Fig.~\ref{fig1}(a). Interestingly, some head tasks perform worse under FULL than under LT: more complete and diverse training data do not yield uniform improvements across all tasks. This leads us to consider whether the long-tailed multi-task learning problem is related to how different tasks jointly influence a shared policy. We further visualize the model's attention during inference. Figures~\ref{fig1}(b) and (c) show that head tasks attend reasonably well to visual evidence associated with the target object, but may exhibit more dispersed visual attention due to cross-task interference. Tail tasks face a more severe problem: they struggle to establish reliable task-specific visual grounding. Through the toy experiment in Section~\ref{sec:toy_experiment}, we find that when such evidence is insufficient, the model must still produce an action, and the shared policy tends to fall back on frequent action patterns in the training data, giving rise to a \textbf{head-task action bias}. Thus, tail tasks do not merely suffer from insufficient data exposure; during decision-making, shared action priors associated with head tasks are more likely affect the visual evidence needed for the current task.

This finding raises a new question: \textbf{\emph{how can we preserve reliable decisions on head tasks while preventing tail tasks from relying excessively on head-task action bias?}}

\noindent
\begin{minipage}[t]{0.39\linewidth}
\vspace{0pt}
\setlength{\intextsep}{0pt}
\begin{figure}[H]
\centering
\includegraphics[width=\linewidth]{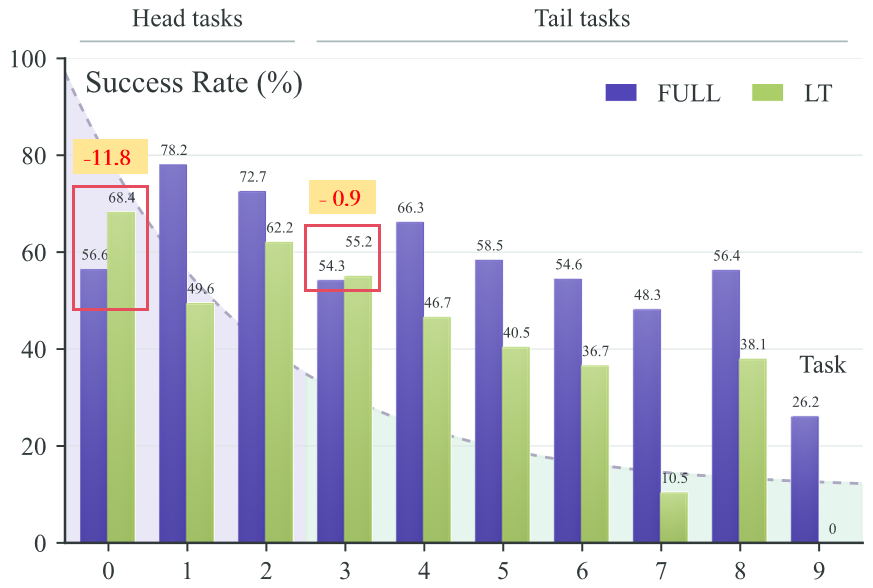}
\par\vspace{-2pt}
{\small\textbf{(a)} Task success rates}\par
\vspace{2pt}
\noindent\hspace*{0.06\linewidth}%
\begin{minipage}[t]{0.45\linewidth}
\centering
\includegraphics[width=\linewidth]{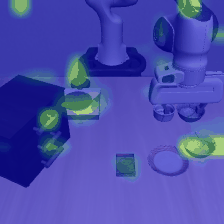}
\par\vspace{2pt}
{\small\textbf{(b)} Head task}
\end{minipage}\hfill
\begin{minipage}[t]{0.45\linewidth}
\centering
\includegraphics[width=\linewidth]{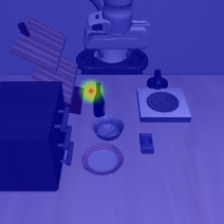}
\par\vspace{2pt}
{\small\textbf{(c)} Tail task}
\end{minipage}
\par\vspace{3pt}
\caption{\textbf{Head vs.tail task performance and intermediate-level visual grounding.} (a)Success rates for policies trained on FULL and LT. (b) Visual attention for ``Pick up the black bowl next to the plate and place it on the plate.'' (c) Visual attention for ``Put the wine bottle on the rack.''}
\label{fig1}
\end{figure}
\end{minipage}\hfill
\begin{minipage}[t]{0.57\linewidth}
\vspace{0pt}
Thus, we introduce \textbf{Evidence-Guided Counterfactual Learning (\methodname{})}, Without requiring additional annotations or training data, \methodname{} directly uses the original data and a relation-aware counterfactual branch to identify visual evidence relevant to decisions during rollouts. It constructs counterfactual samples by removing key relational tokens and strengthens the associations among task instructions, visual relations, and actions through counterfactual training. Crucially, evidence guides both counterfactual construction and supervision in \methodname{}, which adaptively assigns weaker regularization to head tasks and stronger supervision to tail tasks according to the number of demonstrations.

\par\vspace{0.7\baselineskip}\noindent
\textbf{Our contributions:} \textbf{1)}We reassess the conventional view that tail-task failures arise from data scarcity and show through a toy experiment that, when task-specific visual grounding is insufficient, tail-task predictions fall back on action bias shaped by head tasks. \textbf{2)} We propose \methodname{}, which suppresses tail tasks’ reliance on action priors via counterfactual supervised training to build their task-specific evidence, without requiring extra annotations or additional training data. \textbf{3)} We evaluate \methodname{} on two VLA policies in LIBERO-Core-LT and on two real-world robot platforms, Our method achieves state-of-the-art performance, with macro-averaged gains and task-specific outcomes reported.

\end{minipage}
\par\medskip

\section{Related Work}

\textbf{Long-Tailed Learning.} Long-tailed recognition studies class imbalance that biases models toward head classes \citep{zhang2023survey}. Representative approaches include class-balanced re-weighting \citep{cui2019classbalanced}, decoupled representation and classifier learning \citep{kang2020decoupling}, and prior-based logit adjustment \citep{menon2021logitadjustment}. LDAM adjusts classification margins according to class frequencies \citep{cao2019ldam}, Balanced Softmax addresses label distribution shift \citep{ren2020balanced}, and DisAlign combines adaptive score calibration with prior-based re-weighting \citep{zhang2021disalign}. Feature augmentation methods enrich tail classes through transferable class-generic variations \citep{chu2020feature}, learned semantic augmentation directions \citep{li2021metasaug}, or head-class feature displacements \citep{wang2021rsg}. Contrastive methods improve representation geometry: PaCo introduces learnable class-wise centers \citep{cui2021paco}, while BCL balances inter-class contributions \citep{zhu2022balanced}. These methods primarily address discrete class imbalance rather than long-tailed state and behavior distributions in sequential robotic manipulation.

\textbf{Robotic Foundation Models.} Robotic foundation models build on large vision and language models \citep{touvron2023llama2,zhai2023sigmoid,oquab2024dinov2}. PaLM-E integrates visual, state, and textual inputs for embodied reasoning and planning \citep{driess2023palme}. Robot policies span autoregressive, diffusion-based, and continuous generative approaches. RT-1 \citep{brohan2023rt1} scales Transformer-based behavior cloning through action tokenization, while RT-2 \citep{brohan2023rt} transfers Internet-scale vision-language knowledge to robotic control. OpenVLA \citep{kim2024openvla} combines a pretrained language model with SigLIP and DINOv2 features and trains on Open X-Embodiment demonstrations \citep{openx2024}. MiniVLA \citep{belkhale2024minivla} adopts a lightweight Qwen2.5-0.5B backbone, while OpenVLA-OFT improves adaptation and inference through parallel decoding, action chunking, and continuous regression \citep{kim2025oft}. Diffusion policies generate continuous actions through iterative denoising \citep{chi2023diffusion}. Octo \citep{ghosh2024octo} trains a Transformer-based diffusion policy on diverse robot trajectories for adaptation to new observation and action spaces. $\pi_0$ \citep{black2025pi0} combines flow matching with a pretrained vision-language model for cross-embodiment control, while $\pi_{0.5}$ \citep{black2025pi05} broadens task and environment coverage. SmolVLA uses a compact architecture and asynchronous inference for efficient training and deployment \citep{shukor2025smolvla}. Despite these advances, training data can remain imbalanced across tasks, states, and behaviors, motivating further investigation into how long-tailed distributions affect visuomotor representations and action prediction.

\section{Diagnosing and Mitigating Head Task Action Bias}
\label{sec:Diagnosing and Mitigating Head Task Action Bias}
\begingroup
\postdisplaypenalty=10000

\subsection{Toy Experiment on Head Task Action Bias}
\label{sec:toy_experiment}

We construct a two-task diagnostic to examine how task imbalance and visual masking affect action preferences under tail-task instructions. Two objects are positioned symmetrically to the left and right of a shared target container (Fig.~\ref{fig:toy_experiment}(a)). The tasks share the scene, robot initialization, and manipulation procedure, and differ in the instruction specifying the object to manipulate. We fix the number of head-task demonstrations and subsample the tail-task data to obtain frequency ratios $\gamma=N_H/N_T\in\{1,2,5,10,20\}$. This design measures the combined effect of increasing relative imbalance and reducing tail-data coverage.
\begin{figure}[!htbp]
    \centering

    \begin{minipage}[t]{0.22\linewidth}
        \centering
        \begin{minipage}[c][3.1cm][c]{\linewidth}
            \centering
            \includegraphics[
                width=\linewidth,
                height=2.9cm,
                keepaspectratio
            ]{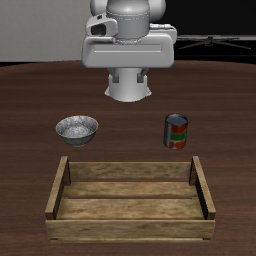}
        \end{minipage}
        \par\vspace{4pt}
        {\small \textbf{(a)} Task setup}
    \end{minipage}\hfill%
    \begin{minipage}[t]{0.47\linewidth}
        \centering
        \begin{minipage}[c][3.1cm][c]{\linewidth}
            \centering
            \includegraphics[
                width=\linewidth,
                height=2.9cm,
                keepaspectratio
            ]{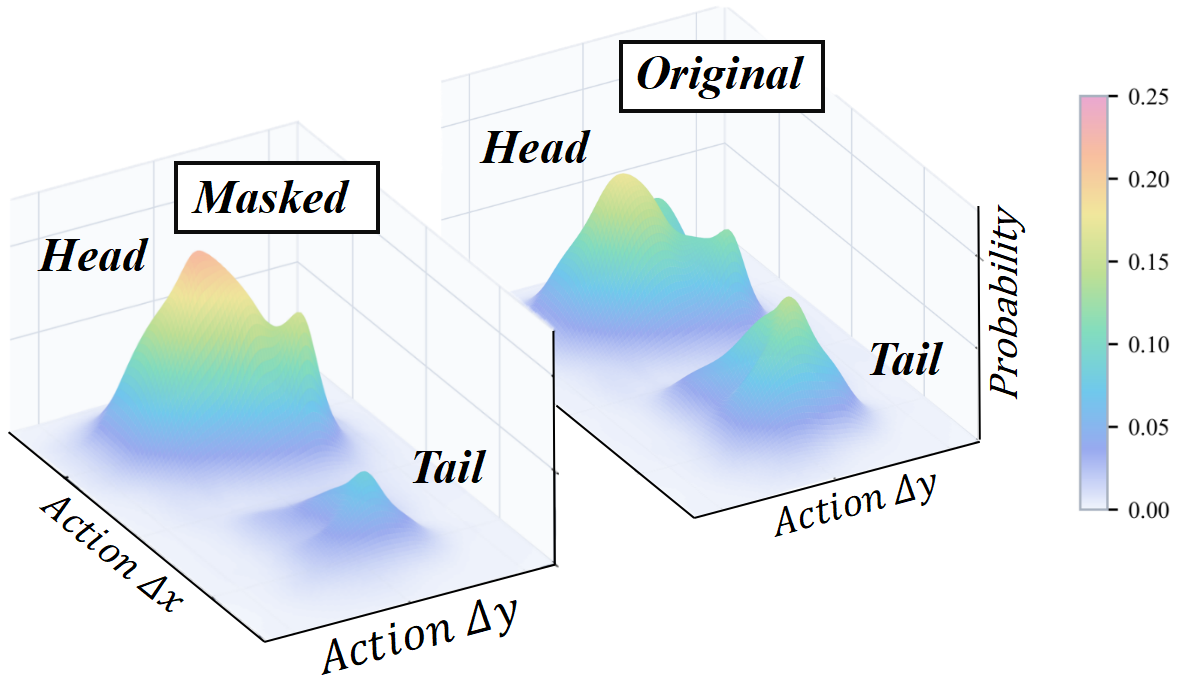}
        \end{minipage}
        \par\vspace{4pt}
        {\small \textbf{(b)} Action distributions}
    \end{minipage}\hfill%
    \begin{minipage}[t]{0.28\linewidth}
        \centering
        \begin{minipage}[c][3.1cm][c]{\linewidth}
            \centering
            \includegraphics[
                width=\linewidth,
                height=2.9cm,
                keepaspectratio
            ]{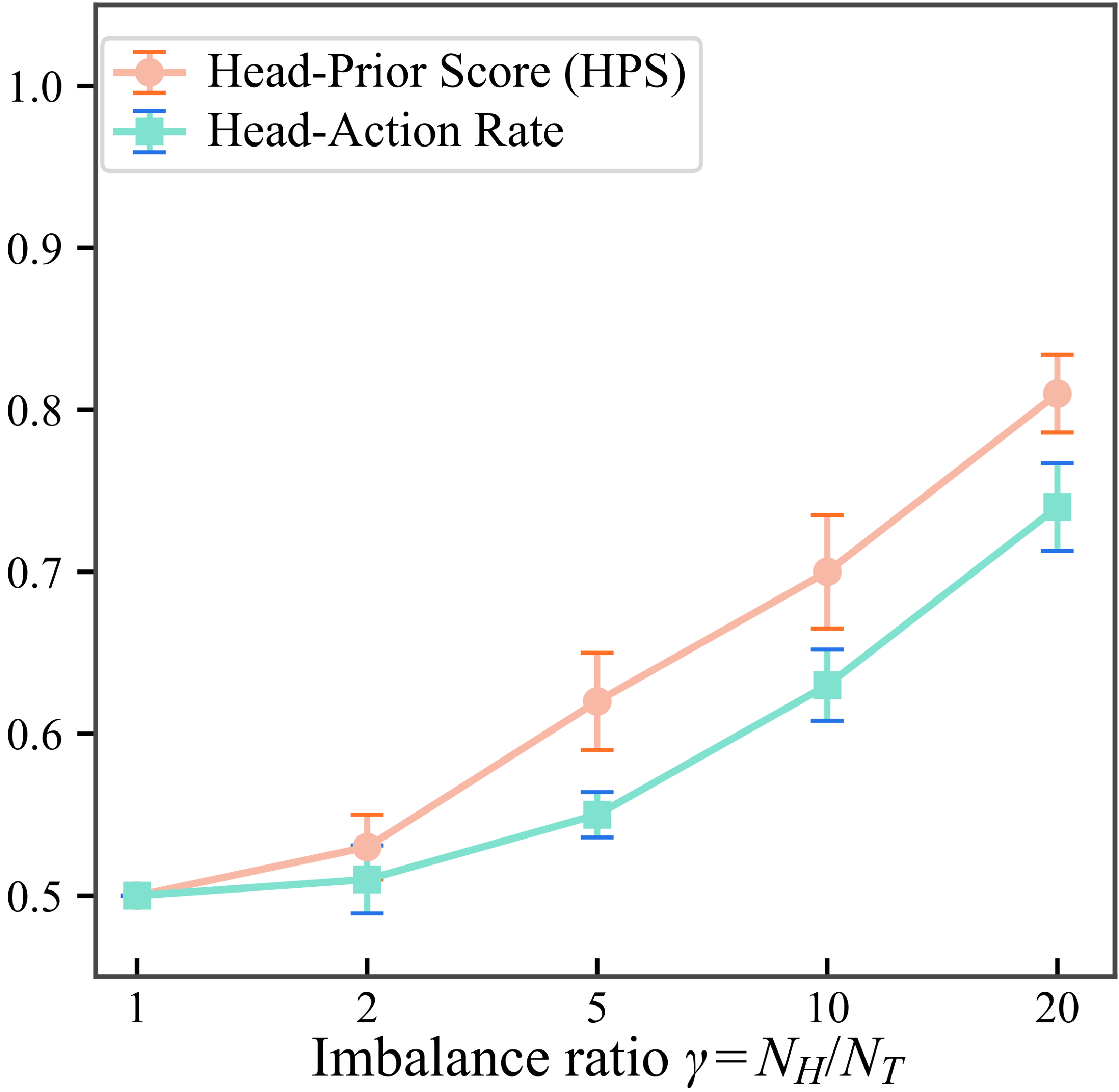}
        \end{minipage}
        \par\vspace{4pt}
        {\small \textbf{(c)} Head-task bias}
    \end{minipage}

    \par\vspace{6pt}
    \caption{
        \textbf{Diagnosing Head Task Action Bias.}
        (a) Two-task environment.
        (b) Action distributions before and after relational masking
        at the same state under a tail-task instruction.
        (c) Head-Prior Score and Head-Action Rate versus task imbalance.
    }
    \label{fig:toy_experiment}
\end{figure}

For each trained policy, we condition on the tail-task instruction at a common set of diagnostic states and inspect the predicted action-token distribution. We focus on horizontal displacement $\Delta x$, which distinguishes the two tasks, and exclude action dimensions shared by both. With the head task assigned to the left object and the tail task to the right, we partition directional tokens into a \emph{head-consistent set} $A_H$, comprising displacements $\Delta x<-\tau$, and a \emph{tail-consistent set} $A_T$, comprising displacements $\Delta x>\tau$. The threshold $\tau$ filters out negligible movements. Let $P_H(x)$ and $P_T(x)$ denote the probability mass assigned to $A_H$ and $A_T$, respectively. We define the \emph{Head-Prior Score (HPS)} as:
\begin{equation}
\operatorname{HPS}(x)=\frac{P_H(x)}{P_H(x)+P_T(x)}
\label{eq:head_prior_score}
\end{equation}
Here, $x$ indexes a diagnostic state. HPS is defined when $P_H(x)+P_T(x)>0$ and measures the relative preference between the two directional sets under a tail-task instruction. A value of $0.5$ indicates equal probability, while larger values indicate a stronger head-task bias. 

We compare predictions from the intact observation, $p(a\mid I,L_T)$, with predictions from the masked observation, $p(a\mid I_{\neg R},L_T)$, at the same diagnostic states. Here, $L_T$ is the tail-task instruction and $I_{\neg R}$ denotes the observation after masking selected task-relevant visual evidence. Masking shifts the action distribution further toward the head-task direction (Fig.~\ref{fig:toy_experiment}(b)). We also measure the \emph{Head-Action Rate}, the fraction of rollouts in which the initial movement is toward the head-task object despite a tail-task instruction. Both HPS and the Head-Action Rate increase with task imbalance rate (Fig.~\ref{fig:toy_experiment}(c)). Exchanging the head--tail assignment reverses the directional preference, which argues against a fixed left--right bias in this setting. 

\subsection{Frequency-Aware Counterfactual Supervision}
\label{sec:facs}
The above experiments show that bias induced by head tasks affects inference on tail tasks, reducing their performance. Based on this finding, We introduce \textbf{Evidence-Guided Counterfactual Learning (\methodname{})}, a counterfactual learning strategy that masks key visual regions for tail tasks. By constructing a counterfactual branch that disrupts the original visual information, FACS encourages the model to reduce its reliance on head tasks and re-establish connections to tail-task-specific evidence.Figure~\ref{fig:facs_overview} summarizes the framework.
\begin{figure}
    \centering
    \includegraphics[width=1\linewidth]{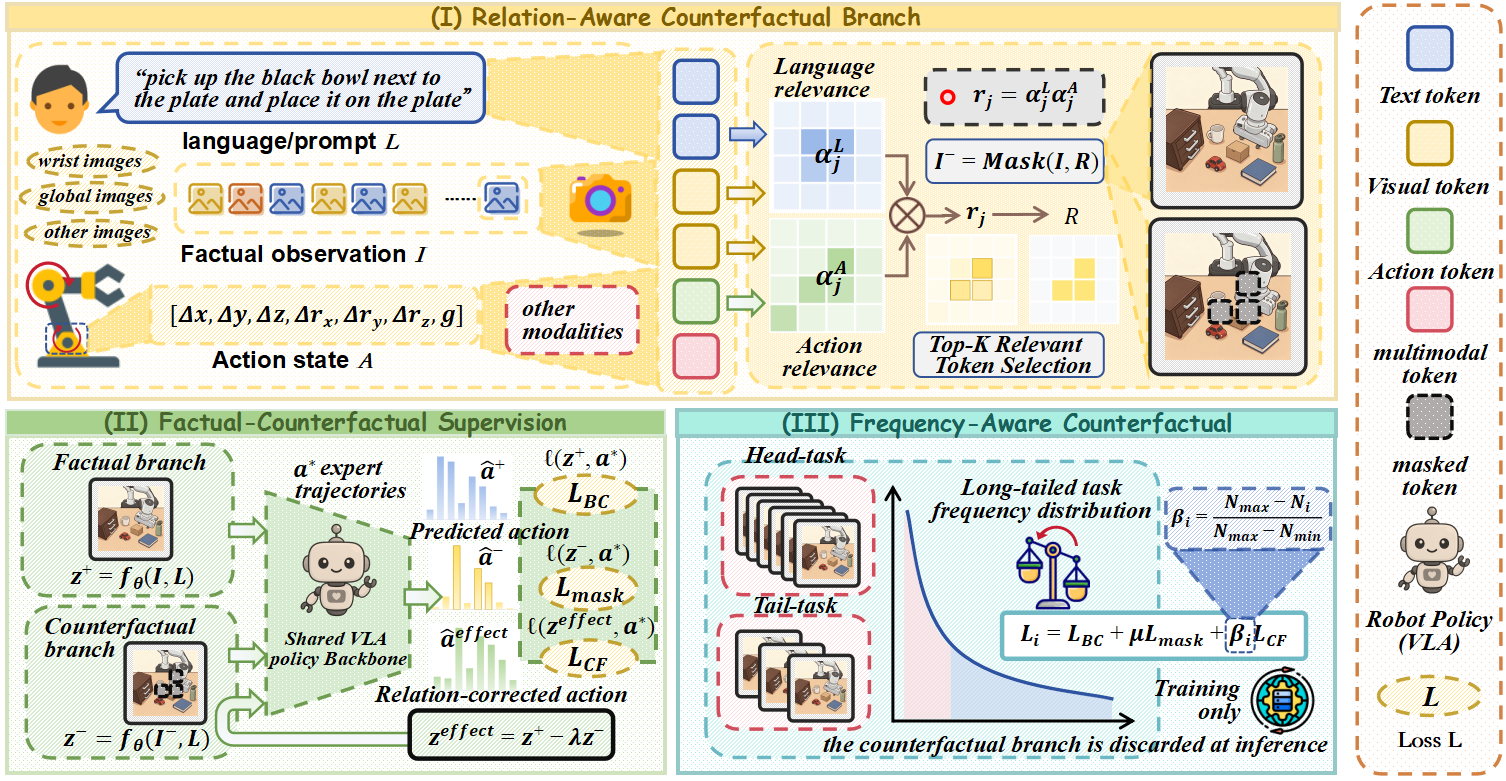}
    \caption{\textbf{\textbf{Overview of \methodname{}.}
\textbf{(I)} Joint language--action relevance selects top-$K$ visual tokens for masking.
\textbf{(II)} A shared policy produces factual, masked, and contrastive training outputs, supervised by $\mathcal{L}_{\mathrm{BC}}$, $\mathcal{L}_{\mathrm{mask}}$, and $\mathcal{L}_{\mathrm{CF}}$, respectively.
\textbf{(III)} Task frequency determines the per-sample weight of the counterfactual loss. At deployment, only the factual policy is used.}
\label{fig:facs_overview}}
    \label{fig:placeholder}
\end{figure}
\subsubsection{Relation-Aware Counterfactual Branch}
\label{sec:relational_mask}
Let $I$ denote the visual observation and $L$ the instruction.
For simplicity, we omit policy-specific conditioning variables and write the factual prediction from the intact observation as:
\begin{equation}
z^+=f_\theta(I,L)
\label{eq:factual_prediction}
\end{equation}
Here, $z^+$ denotes the prediction output on which the supervised action loss is evaluated. The factual and counterfactual branches share the same prediction space, allowing their contrast to be supervised using the same action loss.

To identify task-relevant visual evidence, we combine instruction-related and action-related attention as complementary relevance signals, while recognizing that attention weights alone do not reliably explain model predictions \citep{jain2019attention}. The former reflects how instruction-related queries attend to visual tokens, providing cues for locating the target and associated regions described by the instruction. The latter reflects how action-prediction queries attend to visual tokens, indicating which visual information the model attends to when generating actions. Relying solely on instruction-related attention may select semantically relevant regions that are not sufficiently utilized for action prediction. Conversely, relying solely on action-related attention may select regions shared across tasks that do not adequately distinguish the current task. We therefore combine both signals to prioritize visual tokens relevant to both task semantics and action prediction:
\begin{equation}
r_j=\alpha_j^L\alpha_j^A
\label{eq:relational_token_score}
\end{equation}
Here, $\alpha_j^L$ and $\alpha_j^A$ denote the instruction-related and action-related attention scores for visual token $v_j$, respectively.
The product suppresses tokens with a low score in either signal, emphasizing regions attended to by both. We select the top-$K$ tokens to form the set $R$, using a default selection ratio of 10\%, with ablations on this ratio presented in Section~\ref{sec:ablation Studies}.This joint score guides evidence selection without equating attention weights with the causal contributions of individual tokens.

We then keep the instruction unchanged, mask the selected visual tokens, and process the intact and masked observations using the same policy:
\begin{equation}
I^-=\operatorname{Mask}(I,R),
\label{eq:counterfactual_observation}
\end{equation}
\begin{equation}
z^-=f_\theta(I^-,L).
\label{eq:counterfactual_prediction}
\end{equation}
The masked branch addresses the following question: \textbf{\emph{what action predictions does the policy still produce when key visual evidence relevant to the current task is removed?}} Our diagnostic experiments show that insufficient task-specific evidence makes tail-task predictions more susceptible to head-task action bias. We therefore use $z^-$ as a reference under incomplete visual evidence for comparison with the prediction from the intact observation. This reference still reflects the influence of unmasked information and should not be equated with head-task action bias itself.

Inspired by the use of factual--counterfactual prediction differences to guide learning \citep{mothilal2020explaining,chiappa2019path,pmlr-v97-goyal19a,Sauer021}, we construct the following contrastive training output:
\begin{equation}
\boxed{z^{\mathrm{effect}}=z^+-\lambda z^-},
\label{eq:rcd_score}
\end{equation}
where $\lambda$ controls the contribution of the masked branch to the prediction contrast and is fixed at $0.3$ during training, with ablations presented in Section~\ref{sec:ablation Studies}. By applying action supervision to $z^{\mathrm{effect}}$, we encourage the policy to learn action responses supported by task-specific visual evidence in the intact observation, reducing its reliance on head-task action bias. The selected visual regions thus serve not only to construct masked samples but also to inform supervision through the prediction contrast between the two branches, strengthening the associations among task instructions, visual evidence, and actions. The counterfactual branch is used only during training; at deployment, the policy generates actions directly from the intact visual observation and language instruction.
\subsubsection{Frequency-Aware Counterfactual Supervision}
\label{sec:frequency_aware_supervision}
Let $\ell(z,a^*)$ denote the base policy's supervised action-prediction loss, where $a^*$ denotes the expert action target in the corresponding prediction space. We supervise the factual prediction, the masked prediction, and their counterfactual action effect:
\begin{equation}
\mathcal L_{\mathrm{BC}}=\ell(z^+,a^*),\qquad
\mathcal L_{\mathrm{mask}}=\ell(z^-,a^*),\qquad
\mathcal L_{\mathrm{CF}}=\ell(z^{\mathrm{effect}},a^*).
\label{eq:facs_losses}
\end{equation}
These objectives serve different roles. $\mathcal L_{\mathrm{BC}}$ retains the base policy's original supervision and directly optimizes the factual prediction used at deployment. $\mathcal L_{\mathrm{mask}}$ encourages the masked branch to predict actions using the remaining information, providing a meaningful reference under incomplete visual evidence. $\mathcal L_{\mathrm{CF}}$ supervises the counterfactual action effect derived from the two branches, encouraging the policy to associate task-specific visual evidence with actions and reduce its reliance on head-task action bias. The masked branch can still exploit task information that remains available and is not restricted to representing task-independent behavior.

Our diagnostic experiments show that tail tasks are more susceptible to head-task action bias when visual grounding is insufficient. We therefore allocate counterfactual supervision according to demonstration frequency, assigning stronger supervision to less frequent tasks while limiting additional constraints on frequent tasks. Let $N_i$ denote the number of training demonstrations for task $i$, and let $N_{\max}$ and $N_{\min}$ be the largest and smallest demonstration counts across tasks. For $N_{\max}>N_{\min}$, we define the normalized rarity of task $i$ as:
\begin{equation}
r_i=\frac{N_{\max}-N_i}{N_{\max}-N_{\min}},
\qquad r_i\in[0,1].
\label{eq:facs_task_rarity}
\end{equation}
We directly use $r_i$ as the the weight of the counterfactual supervision loss, yielding the training objective for a sample from task $i$:
\begin{equation}
\boxed{
\mathcal L_i=
\mathcal L_{\mathrm{BC}}
+\mu\mathcal L_{\mathrm{mask}}
+r_i\mathcal L_{\mathrm{CF}}
}.
\label{eq:facs_total_loss}
\end{equation}

Here, $\mu$ controls the auxiliary supervision strength of the masked branch.
For the most frequent task, $r_i=0$, retaining only factual and masked prediction supervision. For the least frequent task, $r_i=1$, applying the full weight to counterfactual supervision. Intermediate tasks receive progressively larger weights as their demonstration counts decrease. This weighting operates per sample; each task's overall contribution to optimization also depends on its sampling frequency.

The frequency-based schedule uses demonstration counts as a proxy for the amount of supervision available to each task.
The combined objective aims to balance action prediction and task-specific visual grounding, with its effectiveness assessed through factual-policy performance and loss-component ablations.
\endgroup
\begin{table}[!htbp]
\centering
\caption{Per-task success rates (\%, mean $\pm$ binomial standard deviation of the rate) on the ten-task LIBERO-Core benchmark. The table reports four training configurations for MiniVLA and $\pi_{0.5}$.The Average standard deviation is computed across the three seed-wise mean success rates.}
\label{tab:per_task_success}
\resizebox{\linewidth}{!}{%
\begin{tabular}{l|cccc|cccc}
\toprule
& \multicolumn{4}{c|}{MiniVLA Success Rate (\%)} & \multicolumn{4}{c}{$\pi_{0.5}$ Success Rate (\%, synthetic)} \\
Task & FULL & LT(Baseline) & APA & \sand{\methodname{}(Ours)} & FULL & LT(Baseline) & APA & \sand{\methodname{}(Ours)} \\
\midrule
Task 0 & \rate{56.6}{4.0} & \rate{68.4}{2.3} & \rate{72.6}{2.3} & \sand{\rate{80.0}{2.2}} & \rate{56.7}{2.4} & \rate{72.4}{3.4} & \rate{69.5}{1.6} & \sand{\rate{75.6}{2.5}} \\
Task 1 & \rate{78.2}{4.0} & \rate{49.6}{3.6} & \rate{67.3}{1.4} & \sand{\rate{68.9}{1.3}} & \rate{61.5}{3.5} & \rate{64.4}{1.6} & \rate{65.8}{1.9} & \sand{\rate{71.7}{1.8}} \\
Task 2 & \rate{72.7}{3.8} & \rate{62.2}{3.2} & \rate{66.1}{1.8} & \sand{\rate{74.7}{1.8}} & \rate{72.6}{2.4} & \rate{58.7}{2.4} & \rate{57.8}{1.5} & \sand{\rate{66.4}{0.7}} \\
Task 3 & \rate{50.3}{3.8} & \rate{55.2}{3.5} & \rate{70.0}{2.1} & \sand{\rate{83.5}{1.8}} & \rate{50.2}{1.6} & \rate{52.6}{2.6} & \rate{55.4}{1.9} & \sand{\rate{59.6}{0.4}} \\
Task 4 & \rate{66.3}{4.2} & \rate{46.7}{3.6} & \rate{56.5}{1.6} & \sand{\rate{64.7}{1.4}} & \rate{68.3}{2.0} & \rate{45.8}{1.5} & \rate{43.3}{3.4} & \sand{\rate{52.1}{3.4}} \\
Task 5 & \rate{58.5}{4.2} & \rate{40.5}{3.4} & \rate{44.5}{1.6} & \sand{\rate{48.0}{1.3}} & \rate{58.8}{2.3} & \rate{36.6}{2.7} & \rate{40.3}{1.7} & \sand{\rate{44.4}{2.4}} \\
Task 6 & \rate{54.6}{4.0} & \rate{36.7}{3.4} & \rate{37.2}{1.8} & \sand{\rate{39.7}{1.1}} & \rate{53.9}{2.1} & \rate{29.8}{3.5} & \rate{27.9}{1.6} & \sand{\rate{37.5}{3.0}} \\
Task 7 & \rate{48.3}{4.2} & \rate{10.5}{4.0} & \rate{11.8}{1.6} & \sand{\rate{13.7}{0.9}} & \rate{46.5}{2.1} & \rate{23.1}{0.8} & \rate{27.6}{2.2} & \sand{\rate{30.7}{1.2}} \\
Task 8 & \rate{56.4}{4.2} & \rate{38.1}{3.4} & \rate{20.9}{1.3} & \sand{\rate{54.3}{1.1}} & \rate{58.2}{2.3} & \rate{15.3}{3.2} & \rate{19.4}{0.9} & \sand{\rate{22.8}{2.4}} \\
Task 9 & \rate{26.2}{4.0} & \rate{0.0}{0.0} & \rate{1.3}{0.9} & \sand{\rate{2.4}{0.7}} & \rate{26.2}{2.2} & \rate{4.3}{0.7} & \rate{10.1}{0.8} & \sand{\rate{11.3}{3.7}} \\
\midrule
Average & \rate{57.2}{4.0} & \rate{40.8}{3.0} & \rate{44.8}{1.6} & \sand{\rate{53.0}{1.4}} & \rate{55.3}{2.0} & \rate{40.3}{1.1} & \rate{41.7}{1.6} & \sand{\rate{47.2}{0.6}} \\
\bottomrule
\end{tabular}%
}
\end{table}
\section{Experiments}
This section is organized around the following research questions:

\noindent\fbox{%
\begin{minipage}{0.96\linewidth}
\setlength{\parindent}{0pt}%
\setlength{\parskip}{0pt}%
\setlength{\baselineskip}{1.05\baselineskip}%
\small\itshape
1) Does \methodname{} improve long-tail policy performance in simulation and the real world?\\
2) How do different design choices affect the performance of \methodname{}?\\
3) How robust is \methodname{} to different forms of long-tailed distributions?
\end{minipage}%
}

We address the first question in Sections~\ref{sec:simulation_experiments}
and~\ref{sec:real_world_experiments}, the second in
Section~\ref{sec:ablation_studies}, and the third in
Section~\ref{sec:context_robustness}. Task definitions and demonstration
frequencies are provided in Appendix~\ref{app:dataset_distribution}.

\subsection{Simulation Experiments}
\label{sec:simulation_experiments}
We conduct simulation experiments in LIBERO \citep{liu2023libero}, using the 10 manipulation tasks in the LIBERO-Core benchmark \citep{zhu2026beyond}. The tasks cover diverse manipulation skills, including object picking, transport, placement, pushing, and container interaction. All tasks are evaluated in closed loop with a 7-DoF Franka Panda manipulator.

\textbf{Experimental Setup.} Following \citet{zhu2026beyond}, we use LIBERO-Core-FULL and LIBERO-Core-LT and evaluate MiniVLA \citep{belkhale2024minivla} and $\pi_{0.5}$ \citep{black2025pi05}. MiniVLA is initialized from a checkpoint pretrained on LIBERO-90, and $\pi_{0.5}$ is initialized from $\pi_{0.5}$-Base. We compare four configurations: FULL, trained on LIBERO-Core-FULL; LT, trained directly on LIBERO-Core-LT; APA, using the augmentation strategy of \citet{zhu2026beyond}; and \methodname{}, trained on LIBERO-Core-LT with the proposed auxiliary objectives. All reported main results use factual-only inference: the masked branch, the contrastive output, and task-frequency weights are not used during action generation.

\textbf{Results and Analysis.}
Table~\ref{tab:per_task_success} reports success rates on the ten LIBERO-Core tasks. \methodname{} increases macro-averaged success from 40.8\% to 53.0\% for MiniVLA and from 40.3\% to 47.2\% for $\pi_{0.5}$, corresponding to gains of 12.2 and 6.9 percentage points, or relative improvements of approximately 29.9\% and 17.1\%, respectively. These results indicate that \methodname{} provides consistent overall performance gains across different VLA policies. Compared with APA~\cite{zhu2026beyond}, \methodname{} improves macro-averaged success from 44.8\% to 53.0\% for MiniVLA and from 41.7\% to 47.2\% for $\pi_{0.5}$. On tail tasks (Tasks 7-9), average success increases from 11.3\% to 23.5\% for MiniVLA and from 19.0\% to 21.6\% for $\pi_{0.5}$.These gains are achieved using only the existing training data, without requiring additional data construction. Notably, ECL improves performance on tail tasks while also benefiting head tasks, avoiding the cross-task bias observed with FULL.
\setcounter{figure}{2}
\begin{figure}[t]
\centering
\includegraphics[width=\linewidth]{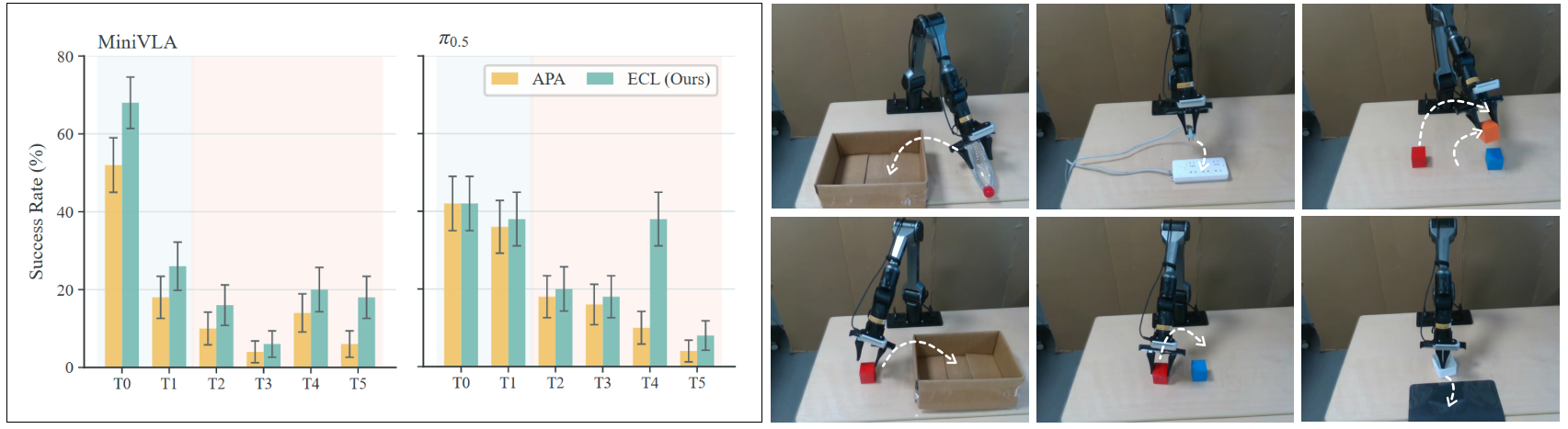}
\caption{Real-world evaluation of \methodname{} with MiniVLA and $\pi_{0.5}$ on AGILEX PiPER under long-tailed task distributions. Evaluation amd results on xArm6 are provided in Appendix~\ref{app:qualitative_rollouts}.}
\end{figure}
\begin{table}[t]
\centering
\caption{Real-world per-task success rates (\%) on AGILEX PIPER and xArm6. For each platform, APA and \methodname{} are evaluated with MiniVLA and $\pi_{0.5}$ on six tasks ordered from head tasks (Task 0-1) to tail tasks (Task 2-5).}
\label{tab:real_success}
\setlength{\tabcolsep}{8pt}      
\resizebox{\linewidth}{!}{%
\begin{tabular}{lcccccccc}
\toprule
& \multicolumn{4}{c}{\textbf{AGILEX PIPER}}
& \multicolumn{4}{c}{\textbf{xArm6}} \\
\cmidrule(lr){2-5}\cmidrule(lr){6-9}
& \multicolumn{2}{c}{MiniVLA}
& \multicolumn{2}{c}{$\pi_{0.5}$}
& \multicolumn{2}{c}{MiniVLA}
& \multicolumn{2}{c}{$\pi_{0.5}$} \\
\cmidrule(lr){2-3}\cmidrule(lr){4-5}
\cmidrule(lr){6-7}\cmidrule(lr){8-9}
Task
& APA & \sand{\methodname{}(Ours)}
& APA & \sand{\methodname{}(Ours)}
& APA & \sand{\methodname{}(Ours)}
& APA & \sand{\methodname{}(Ours)} \\
\midrule
Task 0
& \rate{52}{7.1} & \sand{\rate{68}{6.6}}
& \rate{42}{7.0} & \sand{\rate{42}{7.0}}
& \rate{60}{6.9} & \sand{\rate{62}{6.9}}
& \rate{64}{6.8} & \sand{\rate{68}{7.0}} \\
Task 1
& \rate{18}{5.4} & \sand{\rate{26}{6.2}}
& \rate{36}{6.8} & \sand{\rate{38}{6.9}}
& \rate{20}{5.7} & \sand{\rate{38}{6.9}}
& \rate{20}{5.7} & \sand{\rate{40}{6.9}} \\
Task 2
& \rate{10}{4.2} & \sand{\rate{16}{5.2}}
& \rate{18}{5.4} & \sand{\rate{20}{5.7}}
& \rate{12}{4.6} & \sand{\rate{20}{5.7}}
& \rate{22}{5.9} & \sand{\rate{24}{6.0}} \\
Task 3
& \rate{4}{2.8} & \sand{\rate{6}{3.4}}
& \rate{16}{5.2} & \sand{\rate{18}{5.4}}
& \rate{4}{2.8} & \sand{\rate{6}{2.8}}
& \rate{14}{4.9} & \sand{\rate{20}{5.7}} \\
Task 4
& \rate{14}{4.9} & \sand{\rate{20}{5.7}}
& \rate{10}{4.2} & \sand{\rate{38}{6.9}}
& \rate{18}{5.4} & \sand{\rate{18}{5.4}}
& \rate{20}{5.7} & \sand{\rate{20}{5.7}} \\
Task 5
& \rate{6}{3.4} & \sand{\rate{18}{5.4}}
& \rate{4}{2.8} & \sand{\rate{8}{3.8}}
& \rate{2}{2.0} & \sand{\rate{12}{4.6}}
& \rate{4}{2.8} & \sand{\rate{12}{4.6}} \\
\midrule
Average
& 17.3 & \sand{25.7}
& 21.0 & \sand{27.3}
& 19.3 & \sand{26.0}
& 24.0 & \sand{30.7} \\
\bottomrule
\end{tabular}%
}
\end{table}
\subsection{Real-World Experiments}
\label{sec:real_world_experiments}
We further evaluate \methodname{} on two real robotic platforms, AGILEX PIPER and xArm6, having established ECL's improvements over the LT baseline in simulation, we focus the real-world evaluation on a direct comparison with APA~\cite{zhu2026beyond}.

\textbf{Experimental Setup.}
We fine-tune MiniVLA and $\pi_{0.5}$ on a six-task real-world manipulation benchmark with imbalanced demonstration frequencies. Both policies are evaluated separately on two above manipulators, The sample counts and detailed descriptions of the six tasks are provided in Appendix~\ref{app:dataset_distribution}. The per-task outcomes for each platform--policy combination are reported in Table~\ref{tab:real_success}.

\textbf{Results and Analysis.}
Table~\ref{tab:real_success} reports higher macro-averaged success for ECL across all four platform-policy combinations. On AGILEX PIPER, success increases from 17.3\% to 25.7\% for MiniVLA and from 21.0\% to 27.3\% for $\pi_{0.5}$; on xArm6, it increases from 19.3\% to 26.0\% and from 24.0\% to 30.7\%, respectively. ECL improves tail tasks success rates over APA by an average of 9.0\% on AGILEX PIPER and 4.3\% on xArm6. APA~\cite{zhu2026beyond} constructs augmented demonstrations through object replacement, for which its real-world pipeline proposes object localization using YOLOv8~\cite{varghese2024yolov8}, followed by image inpainting.  ECL leverages existing training data without constructing object-replacement demonstrations. These results demonstrate ECL's effectiveness across policy models and robot platforms, highlighting its applicability to real-world robotic manipulation.

\setcounter{figure}{3}
\begin{figure}[t]
\centering
\includegraphics[width=\linewidth]{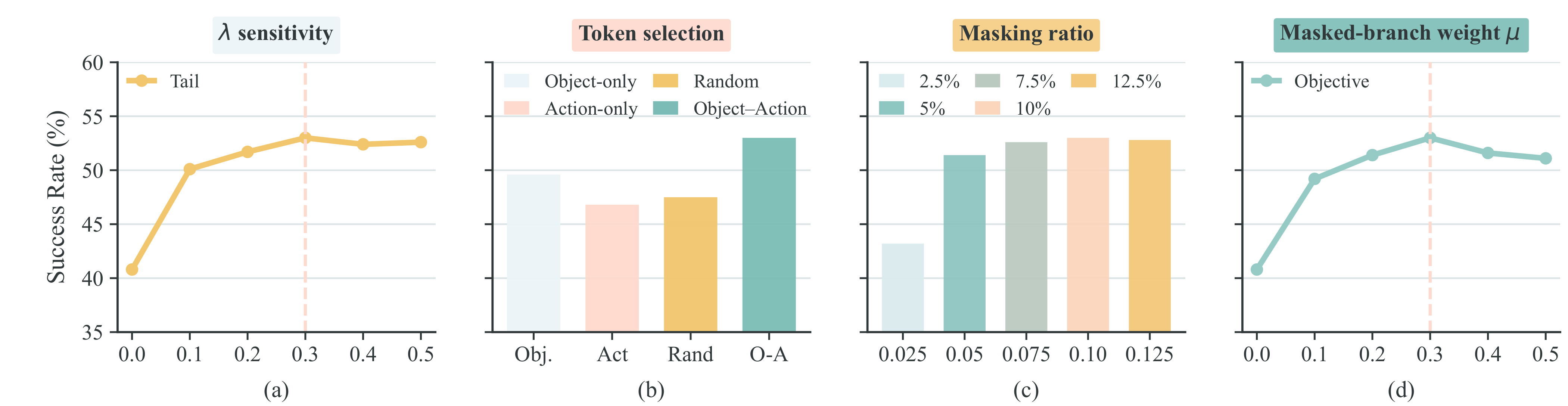}
\caption{Ablation studies of \methodname{} on \textbf{(a)} the training coefficient $\lambda$, \textbf{(b)} visual-token selection strategies, \textbf{(c)} the counterfactual masking ratio $K$, and \textbf{(d)} the masked-branch supervision weight $\mu$. Performance is measured by the average success rate (\%) across 10 tasks in LIBERO-Core-LT using MiniVLA.}
\label{fig:placeholder5}
\end{figure}
\subsection{Ablation Studies}
\label{sec:ablation_studies}
\label{sec:ablation Studies}
We conduct ablation studies to examine the contribution and sensitivity of the main design choices in \methodname{}. Unless otherwise specified, the training and evaluation settings are the same as those in the simulation experiments.

\textbf{Training coefficient $\lambda$.} We examine $\lambda\in\{0,0.1,0.2,0.3,0.4,0.5\}$ for the head and tail task groups and measure performance in terms of average success rate. As shown in Fig.~\ref{fig:placeholder5}(a), success rate is highest at $\lambda=0.3$ among the tested values, and decreases as $\lambda$ increases further. These results suggest that an appropriate weighting of the ECL objective benefits overall task performance, whereas an excessively large coefficient limits these gains. 

\textbf{Visual-token selection.} To assess how the masked tokens are selected, we compare \textbf{Object-only}, \textbf{Action-only}, \textbf{Random}, and the full \textbf{Object--Action Relation} strategy at the same masking ratio. Further details are provided in Appendices~\ref{app:mask_selection_visualization}.The first two strategies use only object-related or action-related attention, respectively; Random masks tokens without using either relevance score. The full strategy ranks tokens by the product of the two scores. In Fig.~\ref{fig:placeholder5}(b), this joint strategy achieves the highest reported success rate 53.0\%. 

\textbf{Counterfactual masking ratio.} We vary the proportion of visual tokens selected for masking by the top-$K$ strategy over $\{0.025, 0.05, 0.075, 0.10, 0.125\}$ while keeping the token-ranking criterion fixed. As shown in Fig.~\ref{fig:placeholder5}(c), the highest average success rate is achieved at a masking ratio of 0.10, which we adopt in the main experiments. A smaller masking ratio may leave much of the selected evidence available, whereas a larger ratio may also remove information needed for action prediction.

\textbf{Masked-branch supervision weight $\mu$.} We investigate the effect of $\mu$ by varying its value over $\{0,0.1,0.2,0.3,0.4,0.5\}$ while keeping all other experimental settings fixed, and report the average success rate across tasks. When $\mu=0$, performance remains close to that of the standard long-tailed training baseline. Without direct supervision on the masked branch, $z^{-}$ may become uninformative, weakening its role as a counterfactual reference. In particular, if $z^{-}$ approaches zero, $z^{\mathrm{effect}}$ becomes nearly identical to $z^{+}$, making counterfactual supervision largely redundant with factual prediction supervision. This provides a possible explanation for the limited improvement at $\mu=0$ and highlights the importance of supervising the masked branch to obtain a meaningful counterfactual reference.

\setcounter{figure}{4}
\begin{figure}[t]
\centering
\includegraphics[width=\linewidth]{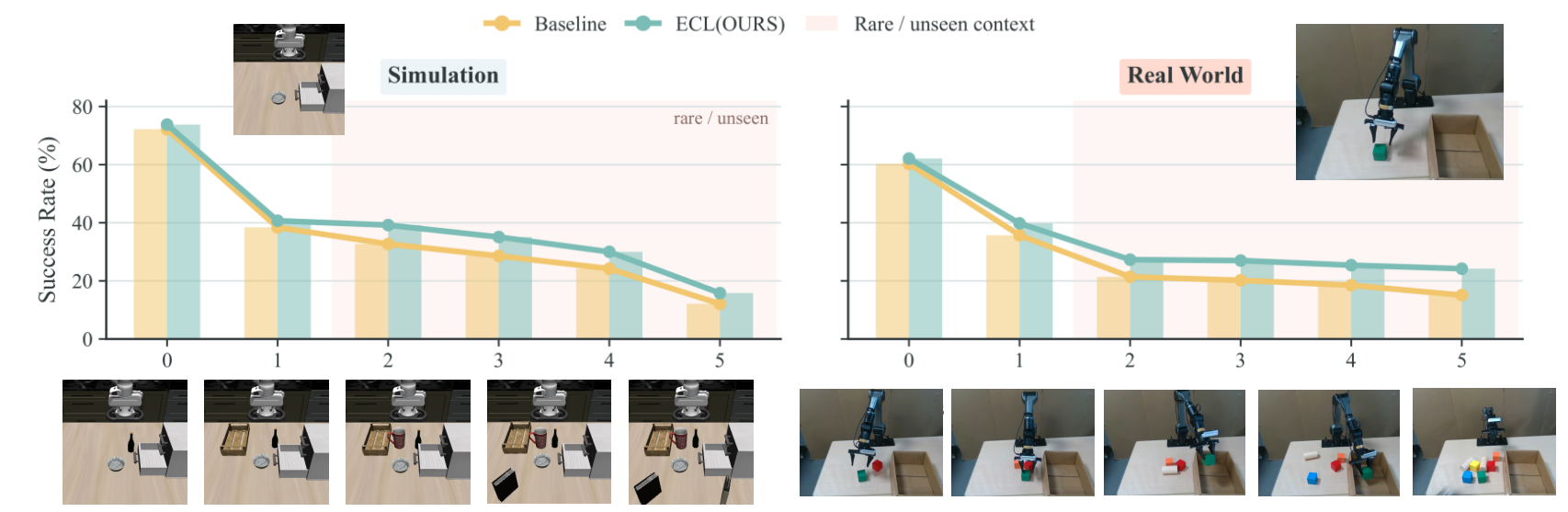}
\caption{Robustness of \methodname{} to context-level long-tailed distributions in simulation (left) and the real world (right). Contexts 1-5 correspond to the scenes below each plot. Yellow and teal lines and bars show success rates (\%) for the LTC baseline and \methodname{} respectively. Pink shading marks rare or unseen contexts (2-5). }
\label{fig:placeholder3}
\end{figure}

\subsection{Robustness to Context-Level Long-Tailed Distributions}
\label{sec:context_robustness}

The preceding experiments consider imbalance across different tasks. A related challenge is unequal coverage of scene configurations within a task. We therefore examine performance under context-level distribution shifts, where the task instruction and manipulation goal are held fixed while non-target objects vary.

To investigate this question, we construct a Long-Tailed Context (LTC) setting. We fix the task semantics, target object, and manipulation goal, while varying only the scene context. Specifically, we create scenes with different levels of complexity and control their numbers in the training set to induce a long-tailed distribution. This setting simulates the common situation in real-world robotic datasets where some contexts are well covered while low-frequency contexts have limited samples.We conduct experiments in both simulation and the real world using the MiniVLA policy, with the original LTC serving as the baseline. Detailed definitions of the context configurations and their corresponding training-sample counts are provided in Appendix~\ref{app:dataset_distribution}.

The results in Fig.~\ref{fig:placeholder3} show that success rates generally decline as evaluation moves toward tail contexts, while \methodname{} consistently outperforms the baseline in both simulation and real-world.\methodname{} increases average success from 34.7\% to 37.6\% in simulation and from 28.5\% to 34.3\% in the real world.These results demonstrate that \methodname{} effectively alleviates performance degradation caused by context distribution shifts, exhibiting strong generalization and decision robustness.

\section{Conclusion, Limitations and Future Work}
In this work, we investigate how long-tailed task distributions affect vision-language-action policies and identify weak task-specific visual grounding as a factor that leaves rare tasks susceptible to head action bias. Evidence-Guided Counterfactual Learning (\methodname{}) identifies visual evidence jointly relevant to language instructions and action generation, uses this evidence to guide counterfactual observation construction through targeted token masking, and adapts the strength of counterfactual supervision to task frequency, encouraging task-specific grounding without requiring additional demonstrations or relation-level annotations.Therefore, we propose a lightweight, simple, and elegant solution to the long-tail problem in robotic policy learning. The source code and demonstrations are available in our   \href{https://anonymous.4open.science/r/EVIDENCE-GUIDED-COUNTERFACTUAL-LEARNING-FOR-LONG-TAILED-ROBOTIC-POLICIES-440F/}{project repository}.

Despite its effectiveness, \methodname{} introduces additional training overhead through counterfactual sample construction and extra forward passes. Under a fixed wall-clock training budget, this overhead creates an opportunity cost: computation devoted to improving tail-task performance could otherwise support further optimization of head tasks. Consequently, preserving head-task performance does not necessarily imply an optimal allocation of the available compute. Future work will explore more efficient counterfactual supervision and adaptive computation allocation to improve the trade-off between training cost and performance across the task distribution.

\section{AI Use Statement}
During the preparation of this manuscript, an AI-based language tool was utilized to refine the prose and enhance overall clarity. The authors have rigorously examined all AI-generated suggestions and assume full accountability for the final submitted version, encompassing all textual content, scientific assertions, and research artifacts.

\bibliography{iclr2027_conference}
\bibliographystyle{iclr2027_conference}

\clearpage
\appendix
\section{Dataset Construction and Task Distribution}
\label{app:dataset_distribution}

\subsection{Simulation Task-Level Long-Tailed Dataset}
\label{app:simulation_task_dataset}

Following APA \citep{zhu2026beyond}, we use the same dataset construction protocol and Pareto-based task-frequency distribution to obtain LIBERO-Core-LT from the ten-task LIBERO-Core-FULL benchmark, built on LIBERO \citep{liu2023libero}. Table~\ref{tab:app_tasks} lists the task instructions and the demonstration counts used in our experiments. Tasks 0--2 form the head group, while all remaining tasks (3--9) form the tail group. The LT counts determine the frequency-aware correction strength in Eq.~(\ref{eq:facs_task_rarity}).
\begin{table}[H]
\centering
\caption{LIBERO-Core task definitions and demonstration counts. ``FULL'' and
``LT'' denote the full and long-tailed training configurations, respectively.
Task IDs follow the ordering in the project benchmark map.}
\label{tab:app_tasks}
\normalsize
\setlength{\tabcolsep}{5pt}
\renewcommand{\arraystretch}{1.12}
\begin{tabular*}{0.96\linewidth}{@{\extracolsep{\fill}}c p{0.66\linewidth} r r}
\toprule
Task & Language instruction & FULL & LT \\
\midrule
0 & Pick up the black bowl next to the plate and place it on the plate & 46 & 46 \\
1 & Pick up the black bowl next to the cookie box and place it on the plate & 47 & 28 \\
2 & Pick up the black bowl on the cookie box and place it on the plate & 45 & 19 \\
3 & Pick up the ketchup and place it in the basket & 42 & 15 \\
4 & Pick up the alphabet soup and place it in the basket & 47 & 11 \\
5 & Push the plate to the front of the stove & 39 & 9 \\
6 & Put the bowl on top of the cabinet & 47 & 8 \\
7 & Put the cream cheese in the bowl & 39 & 7 \\
8 & Put the wine bottle on top of the cabinet & 45 & 6 \\
9 & Put the wine bottle on the rack & 38 & 5 \\
\midrule
\multicolumn{2}{r}{Total demonstrations} & 435 & 154 \\
\bottomrule
\end{tabular*}
\end{table}

\subsection{Real-World Task-Level Long-Tailed Dataset}
\label{app:real_world_task_dataset}

We define a six-task manipulation benchmark for real-world
evaluation on AGILEX PIPER and xArm6. The tasks involve grasping
and placing objects onto supports or into receptacles. The AGILEX PIPER
demonstrations contain synchronized wrist-camera and global-camera views,
whereas the xArm6 demonstrations contain global-camera views. For each robot,
the FULL and LT demonstration counts follow the per-task allocations in
Table~\ref{tab:app_real_tasks}. Tasks are ordered
by decreasing demonstration count in the long-tailed
configuration (LT). The reference configuration (FULL)
allocates 45--50 demonstrations per task, whereas LT uses
50, 46, 29, 13, 10, and 9 demonstrations, respectively.
The resulting allocation contains 285 demonstrations for
FULL and 157 for LT, with an LT maximum-to-minimum
frequency ratio of approximately $5.56{:}1$.

\begin{table}[H]
\centering
\caption{
Real-world task definitions and demonstration
allocation. FULL denotes the approximately balanced
reference configuration, and LT denotes the long-tailed
configuration. Task IDs follow the order used in the
real-world success-rate table.
}
\label{tab:app_real_tasks}
\normalsize
\setlength{\tabcolsep}{5pt}
\renewcommand{\arraystretch}{1.12}
\begin{tabular*}{0.96\linewidth}{@{\extracolsep{\fill}}c p{0.66\linewidth} r r}
\toprule
Task & Language instruction & FULL & LT \\
\midrule
0 & Pick up the block and place it in the box
  & 50 & 50 \\
1 & Stack the red block on top of the blue block
  & 48 & 46 \\
2 & Stack the two blocks on the left on top of the block on the right
  & 47 & 29 \\
3 & Insert the three-prong plug into the socket
  & 49 & 13 \\
4 & Push the box to the designated location
  & 46 & 10 \\
5 & Pick up the water bottle and place it in the box
  & 45 & 9 \\
\midrule
\multicolumn{2}{r}{Total demonstrations}
  & 285 & 157 \\
\bottomrule
\end{tabular*}
\end{table}

\begin{figure}[H]
\centering
\includegraphics[width=0.7\linewidth]{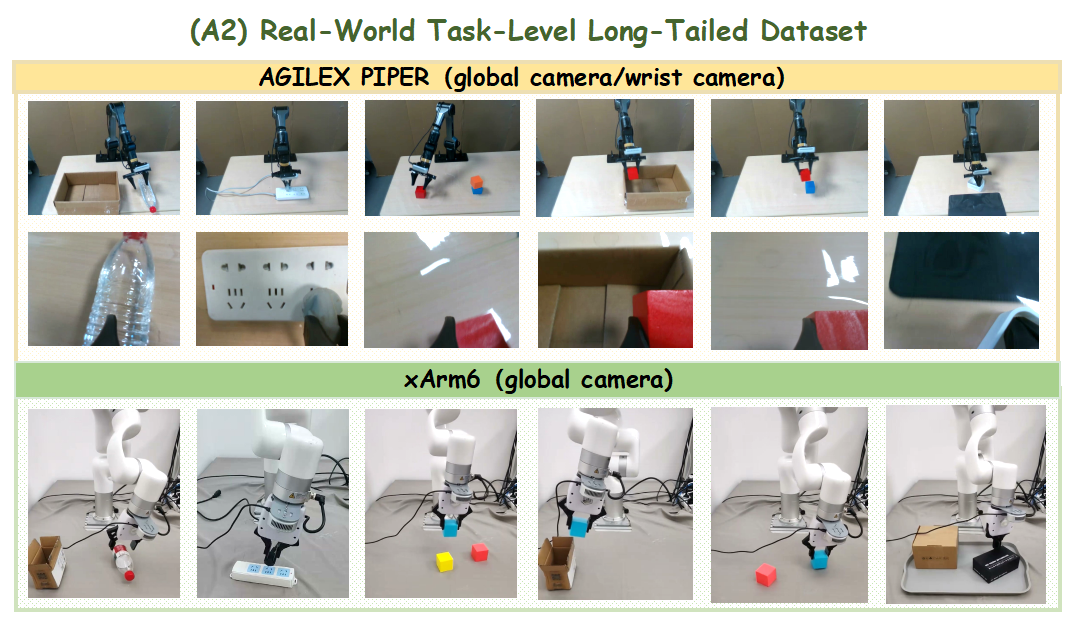}
\caption{Examples from the real-world task-level long-tailed dataset. AGILEX
PIPER provides global-camera and wrist-camera observations, while xArm6 provides
global-camera observations. The columns show the six manipulation tasks listed
in Table~\ref{tab:app_real_tasks}.}
\label{fig:app_real_world_dataset}
\end{figure}
\vspace{-14pt}
\begin{figure}[!htbp]
\centering
\includegraphics[width=0.7\linewidth]{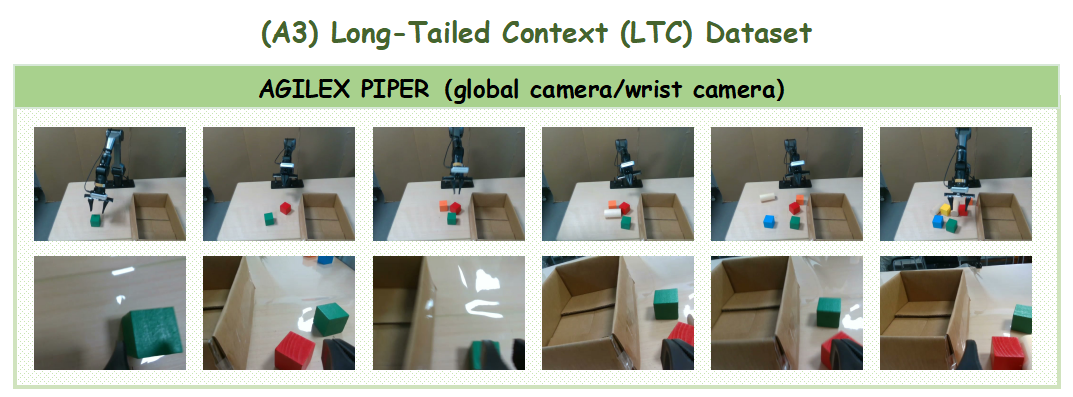}
\caption{Examples from the Long-Tailed Context (LTC) dataset. Columns correspond
to Scenes 0--5 in Table~\ref{tab:app_context_tasks}; the top and bottom rows show
the global-camera and wrist-camera observations, respectively.}
\label{fig:app_ltc_dataset}
\end{figure}

\subsection{Long-Tailed Context (LTC) Dataset}
\label{app:context_level_dataset}

The Long-Tailed Context (LTC) dataset used in Section~4.4 keeps the language
instruction and manipulation goal fixed while varying the scene configuration. Scene IDs
0--5 denote increasingly rare context configurations. This dataset is collected
on AGILEX PIPER and contains synchronized wrist-camera and global-camera views.
Figure~\ref{fig:app_ltc_dataset} visualizes the six scene configurations and
both camera views. Table~\ref{tab:app_context_tasks} reports the balanced FULL
allocation and the long-tailed LT allocation.

\begin{table}[!htbp]
\centering
\caption{Context-level task definition and demonstration counts used in
Section~4.4. All scenes share the same language instruction. FULL contains 50
demonstrations per scene, while LT progressively reduces the number of
demonstrations for rarer scene configurations.}
\label{tab:app_context_tasks}
\normalsize
\setlength{\tabcolsep}{5pt}
\renewcommand{\arraystretch}{1.12}
\begin{tabular*}{0.96\linewidth}{@{\extracolsep{\fill}}p{0.58\linewidth} c r r}
\toprule
Language instruction & Scene & FULL & LT \\
\midrule
\multirow{6}{*}{Put the green block into the box}
  & 0 & 50 & 50 \\
  & 1 & 50 & 25 \\
  & 2 & 50 & 20 \\
  & 3 & 50 & 15 \\
  & 4 & 50 & 10 \\
  & 5 & 50 & 5 \\
\midrule
\multicolumn{2}{r}{Total demonstrations} & 300 & 125 \\
\bottomrule
\end{tabular*}
\end{table}

\clearpage
\section{Experimental Rollout Results}
\label{app:qualitative_rollouts}

This appendix complements the aggregate success rates with representative
closed-loop trajectories. Section~\ref{app:simulation_rollouts} contains one
rollout for each of two representative tasks and each policy backbone (four
comparisons in total). Section~\ref{app:real_rollouts} reports real-world
results on AGILEX PIPER and xArm6. The simulation tables show frames sampled at
fixed evaluation steps. The real-world tables use the selected PIPER
global-camera and wrist-camera frames and the selected xArm6 global-camera frames.

\newcommand{\realframe}[1]{\fbox{\parbox[c][0.038\textheight][c]{0.079\linewidth}{\centering\tiny #1}}}
\newcommand{\realrow}[1]{\realframe{#1} & \realframe{#1} & \realframe{#1} & \realframe{#1} & \realframe{#1} & \realframe{#1} & \realframe{#1} & \realframe{#1}}

\expandafter\newcommand\csname piperrow-minivla-APA-third\endcsname{\parbox[c][0.038\textheight][c]{0.079\linewidth}{\centering\includegraphics[width=\linewidth]{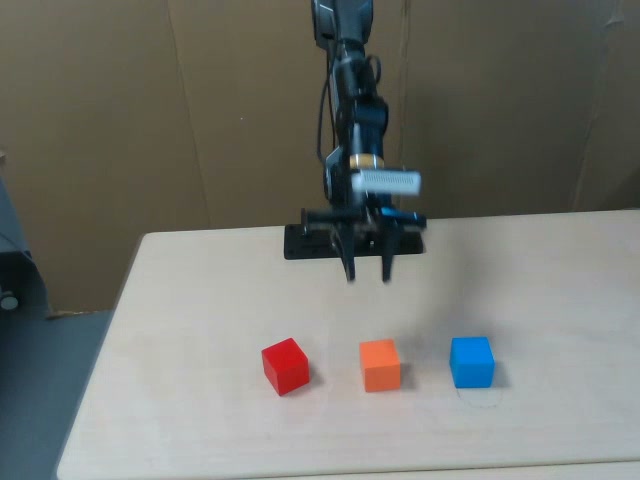}} & \parbox[c][0.038\textheight][c]{0.079\linewidth}{\centering\includegraphics[width=\linewidth]{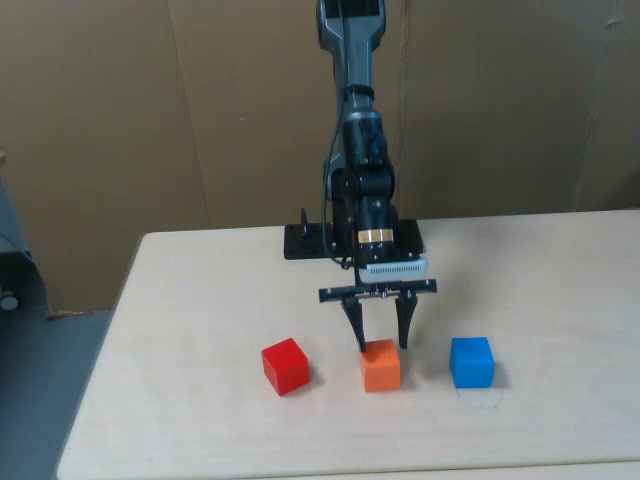}} & \parbox[c][0.038\textheight][c]{0.079\linewidth}{\centering\includegraphics[width=\linewidth]{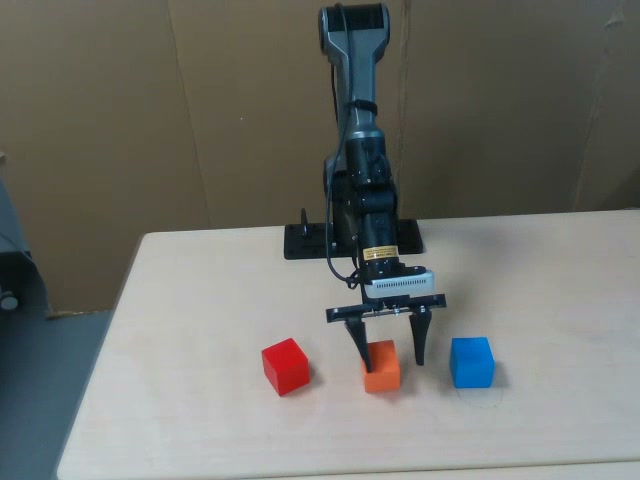}} & \parbox[c][0.038\textheight][c]{0.079\linewidth}{\centering\includegraphics[width=\linewidth]{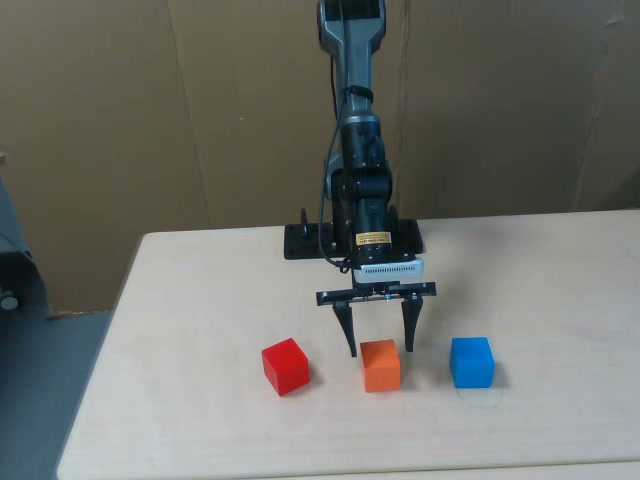}} & \parbox[c][0.038\textheight][c]{0.079\linewidth}{\centering\includegraphics[width=\linewidth]{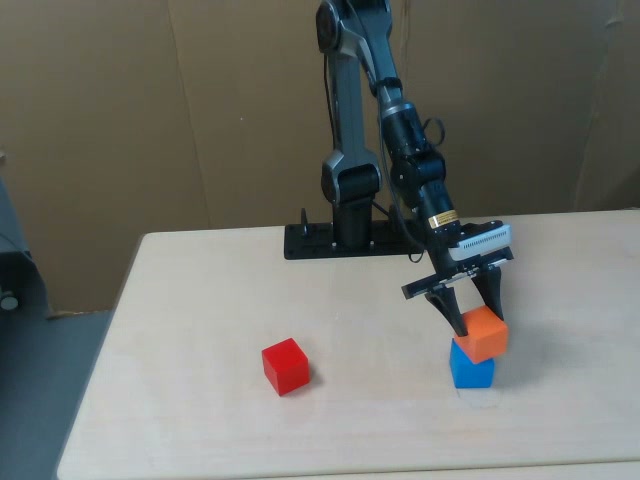}} & \parbox[c][0.038\textheight][c]{0.079\linewidth}{\centering\includegraphics[width=\linewidth]{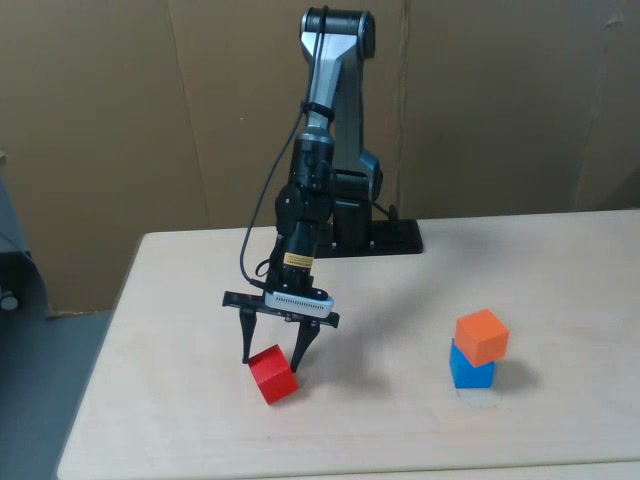}} & \parbox[c][0.038\textheight][c]{0.079\linewidth}{\centering\includegraphics[width=\linewidth]{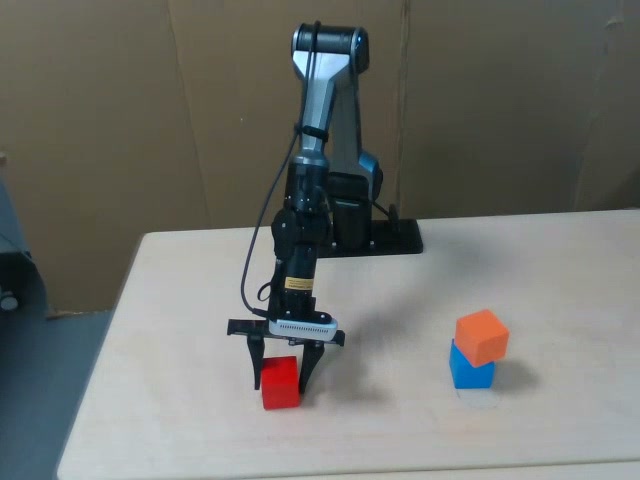}} & \parbox[c][0.038\textheight][c]{0.079\linewidth}{\centering\includegraphics[width=\linewidth]{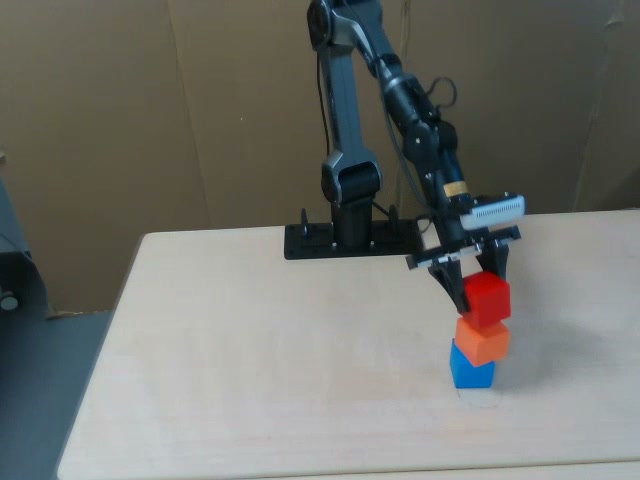}}}
\expandafter\newcommand\csname piperrow-minivla-APA-wrist\endcsname{\parbox[c][0.038\textheight][c]{0.079\linewidth}{\centering\includegraphics[width=\linewidth]{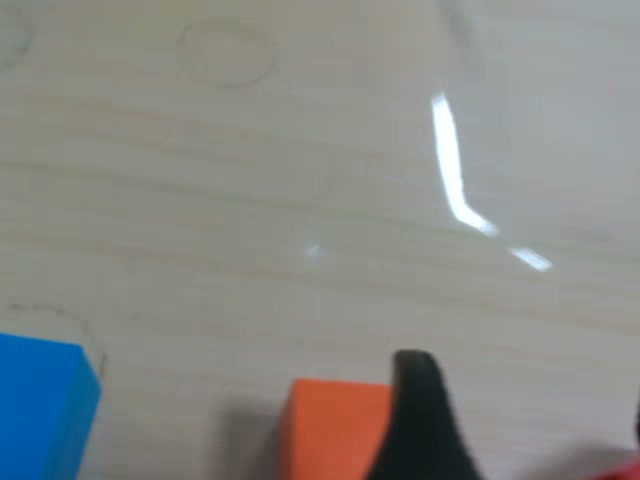}} & \parbox[c][0.038\textheight][c]{0.079\linewidth}{\centering\includegraphics[width=\linewidth]{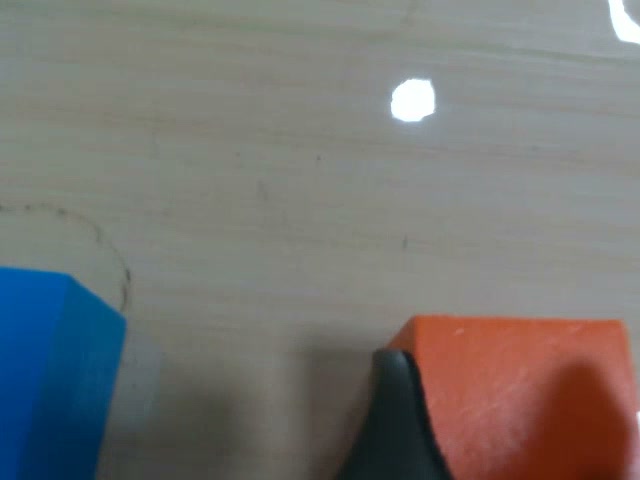}} & \parbox[c][0.038\textheight][c]{0.079\linewidth}{\centering\includegraphics[width=\linewidth]{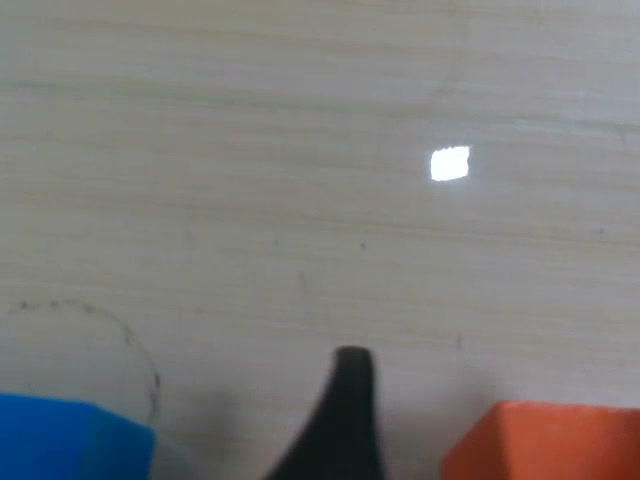}} & \parbox[c][0.038\textheight][c]{0.079\linewidth}{\centering\includegraphics[width=\linewidth]{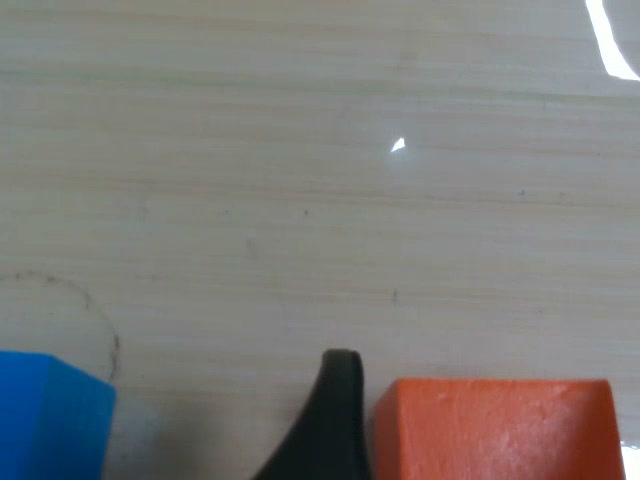}} & \parbox[c][0.038\textheight][c]{0.079\linewidth}{\centering\includegraphics[width=\linewidth]{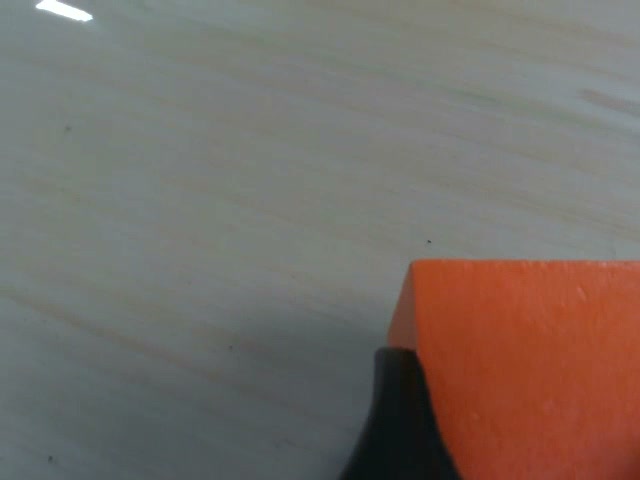}} & \parbox[c][0.038\textheight][c]{0.079\linewidth}{\centering\includegraphics[width=\linewidth]{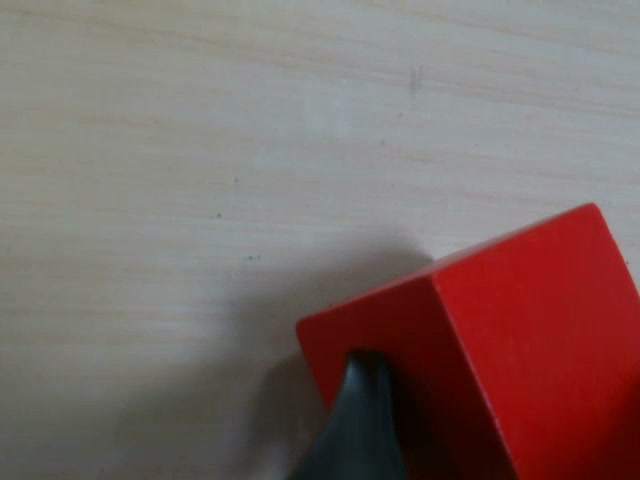}} & \parbox[c][0.038\textheight][c]{0.079\linewidth}{\centering\includegraphics[width=\linewidth]{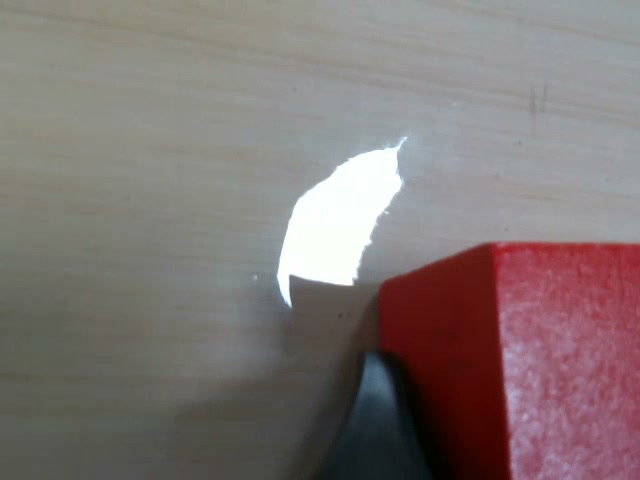}} & \parbox[c][0.038\textheight][c]{0.079\linewidth}{\centering\includegraphics[width=\linewidth]{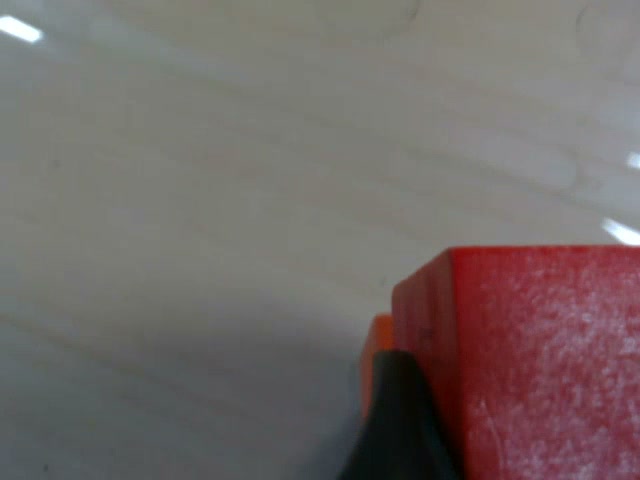}}}
\expandafter\newcommand\csname piperrow-minivla-ECL-third\endcsname{\parbox[c][0.038\textheight][c]{0.079\linewidth}{\centering\includegraphics[width=\linewidth]{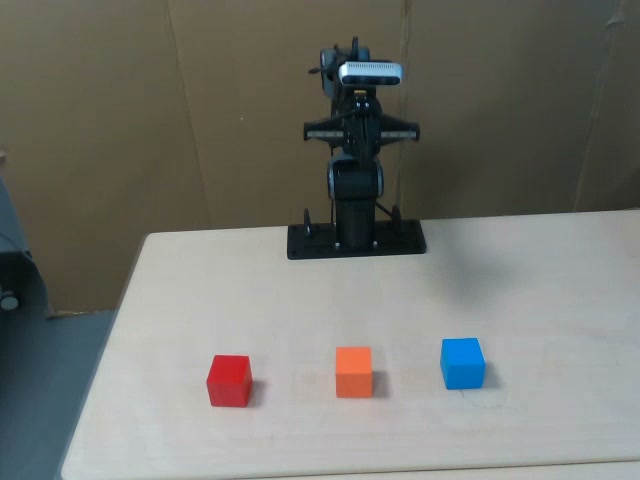}} & \parbox[c][0.038\textheight][c]{0.079\linewidth}{\centering\includegraphics[width=\linewidth]{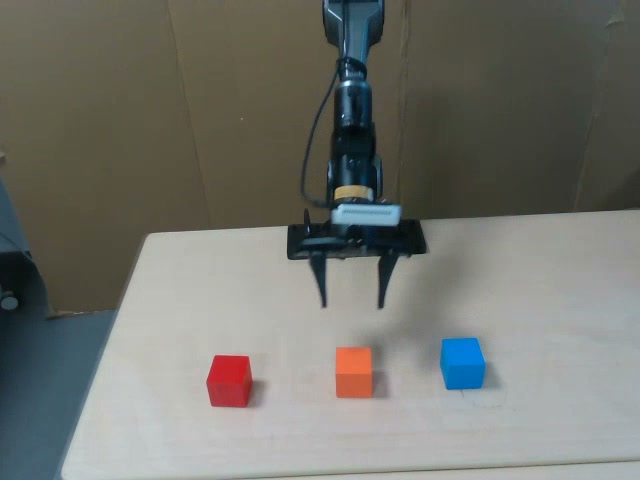}} & \parbox[c][0.038\textheight][c]{0.079\linewidth}{\centering\includegraphics[width=\linewidth]{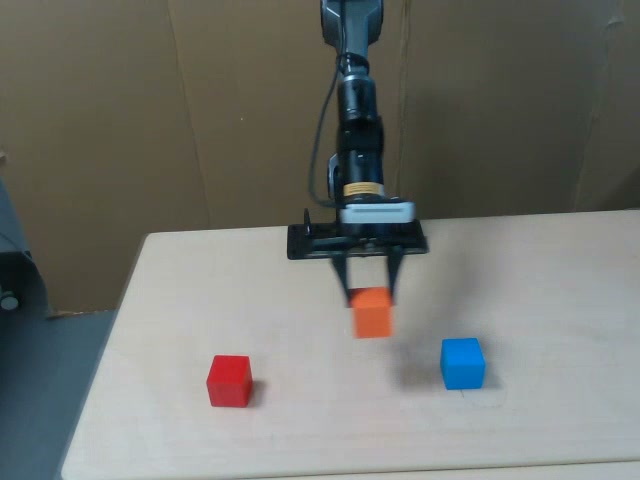}} & \parbox[c][0.038\textheight][c]{0.079\linewidth}{\centering\includegraphics[width=\linewidth]{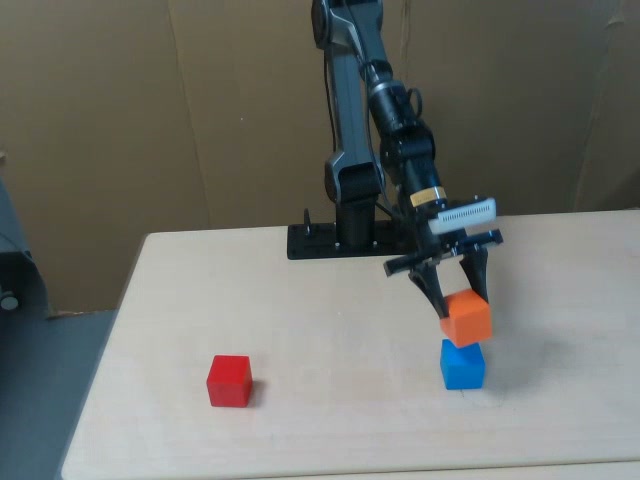}} & \parbox[c][0.038\textheight][c]{0.079\linewidth}{\centering\includegraphics[width=\linewidth]{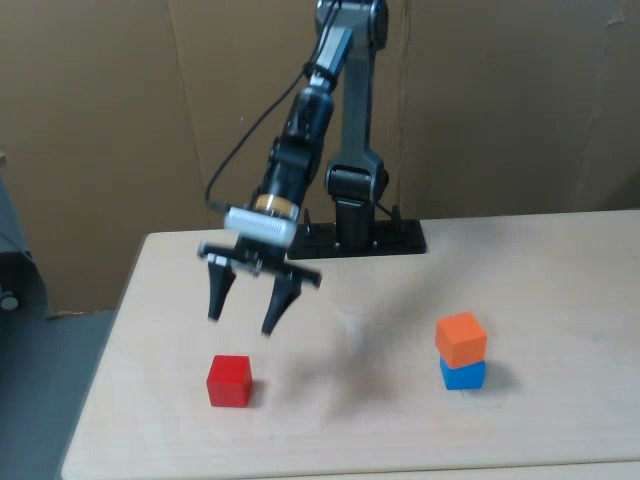}} & \parbox[c][0.038\textheight][c]{0.079\linewidth}{\centering\includegraphics[width=\linewidth]{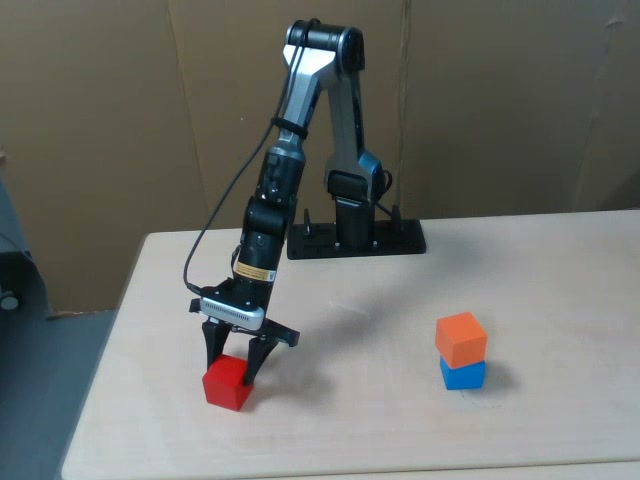}} & \parbox[c][0.038\textheight][c]{0.079\linewidth}{\centering\includegraphics[width=\linewidth]{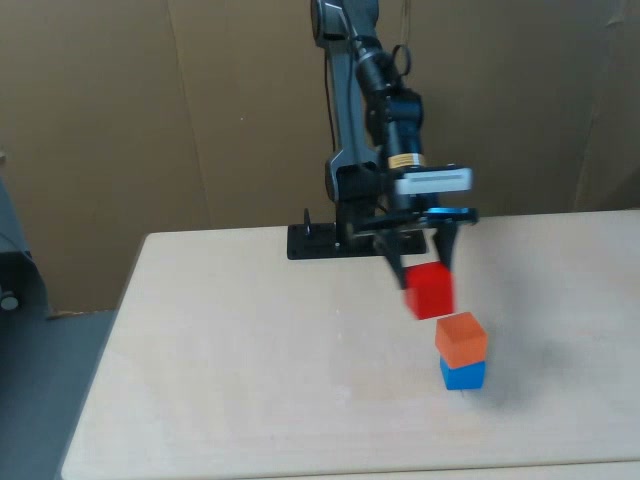}} & \parbox[c][0.038\textheight][c]{0.079\linewidth}{\centering\includegraphics[width=\linewidth]{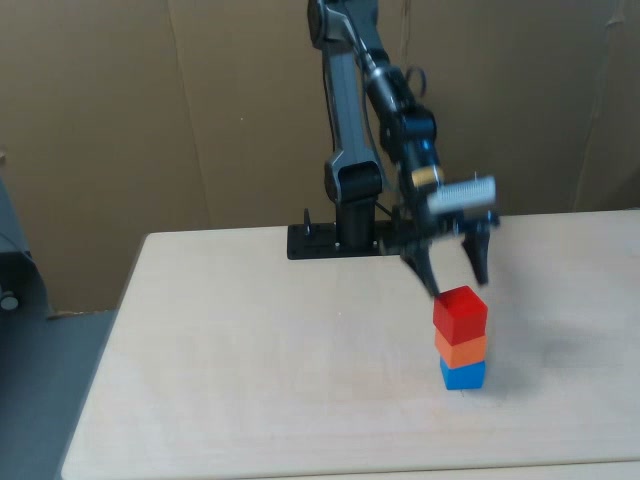}}}
\expandafter\newcommand\csname piperrow-minivla-ECL-wrist\endcsname{\parbox[c][0.038\textheight][c]{0.079\linewidth}{\centering\includegraphics[width=\linewidth]{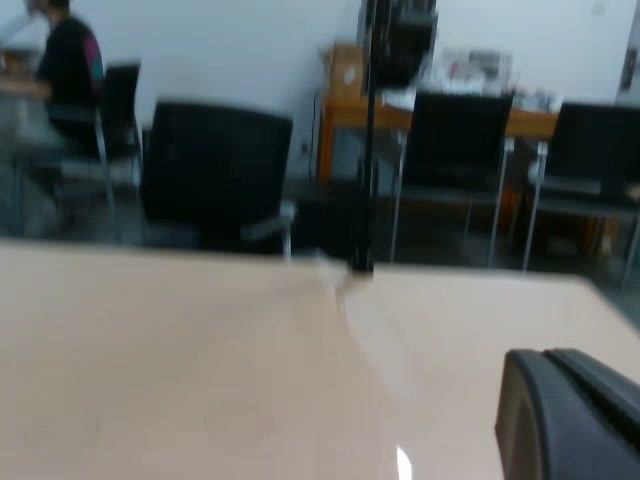}} & \parbox[c][0.038\textheight][c]{0.079\linewidth}{\centering\includegraphics[width=\linewidth]{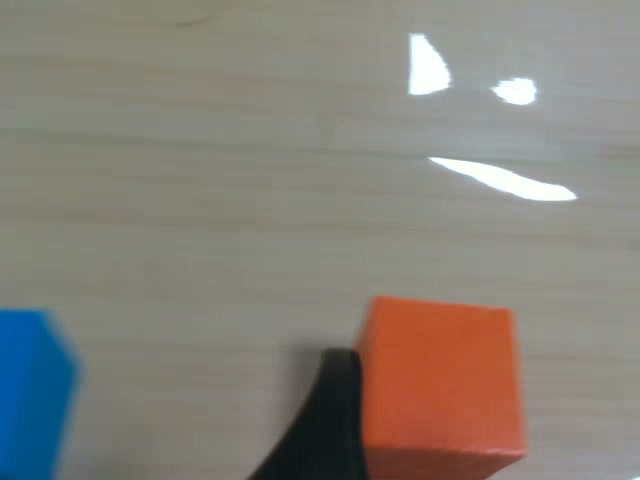}} & \parbox[c][0.038\textheight][c]{0.079\linewidth}{\centering\includegraphics[width=\linewidth]{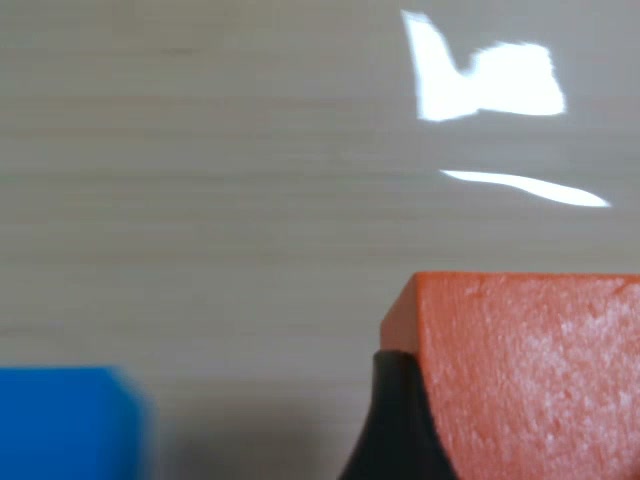}} & \parbox[c][0.038\textheight][c]{0.079\linewidth}{\centering\includegraphics[width=\linewidth]{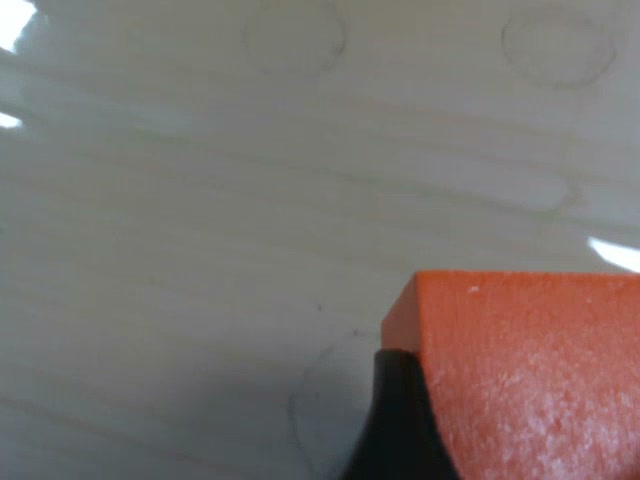}} & \parbox[c][0.038\textheight][c]{0.079\linewidth}{\centering\includegraphics[width=\linewidth]{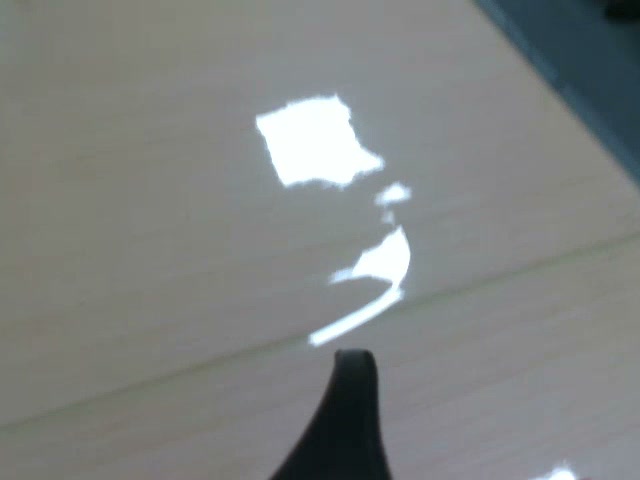}} & \parbox[c][0.038\textheight][c]{0.079\linewidth}{\centering\includegraphics[width=\linewidth]{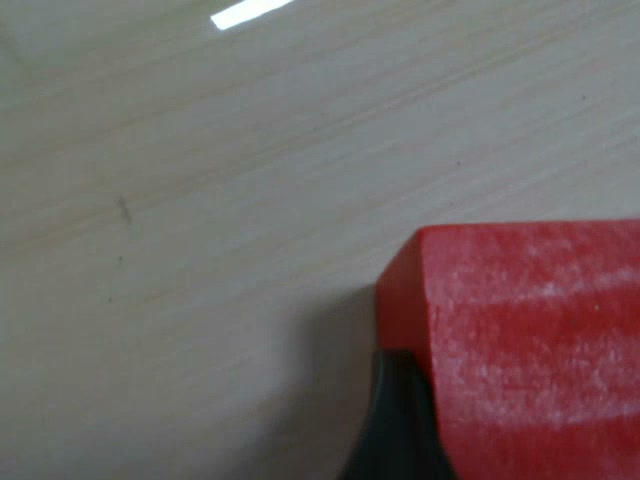}} & \parbox[c][0.038\textheight][c]{0.079\linewidth}{\centering\includegraphics[width=\linewidth]{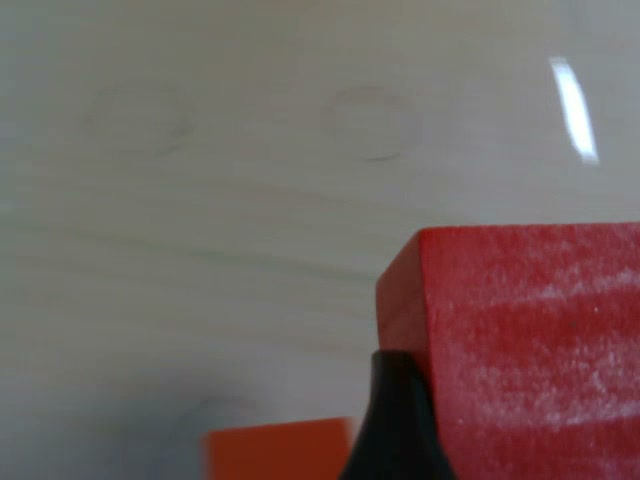}} & \parbox[c][0.038\textheight][c]{0.079\linewidth}{\centering\includegraphics[width=\linewidth]{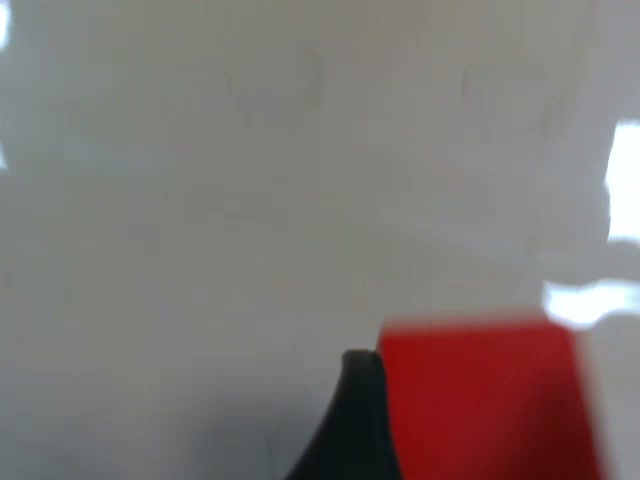}}}
\expandafter\newcommand\csname xarmrow-minivla-APA\endcsname{\parbox[c][0.064\textheight][c]{0.089\linewidth}{\centering\includegraphics[width=\linewidth]{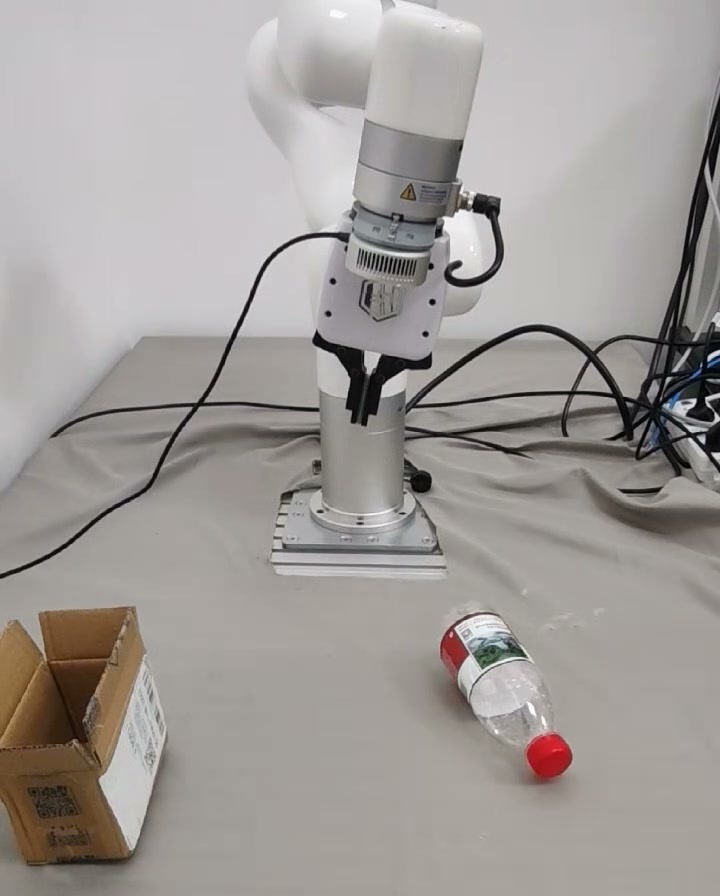}} & \parbox[c][0.064\textheight][c]{0.089\linewidth}{\centering\includegraphics[width=\linewidth]{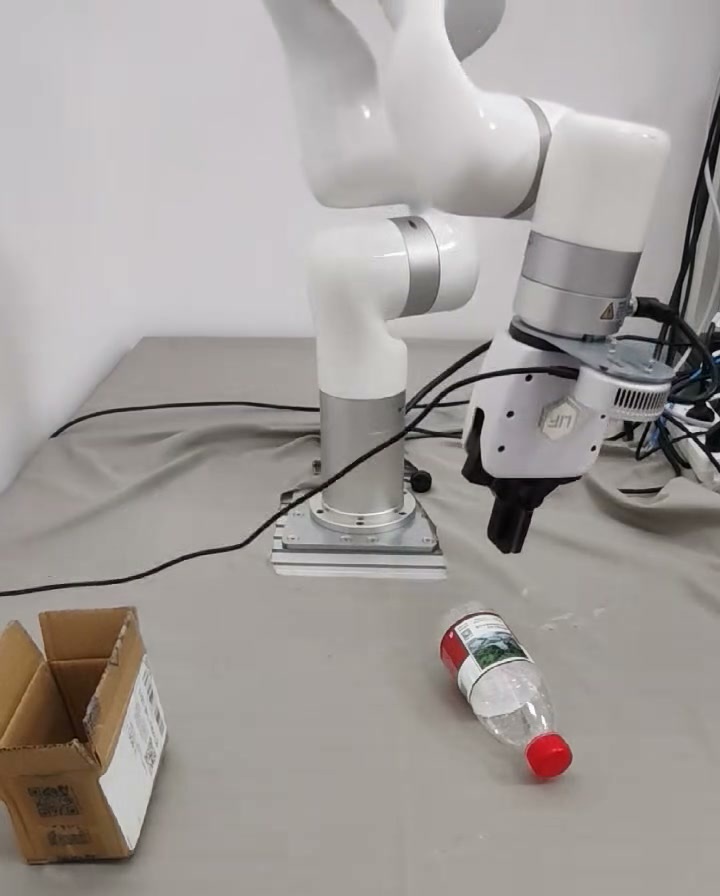}} & \parbox[c][0.064\textheight][c]{0.089\linewidth}{\centering\includegraphics[width=\linewidth]{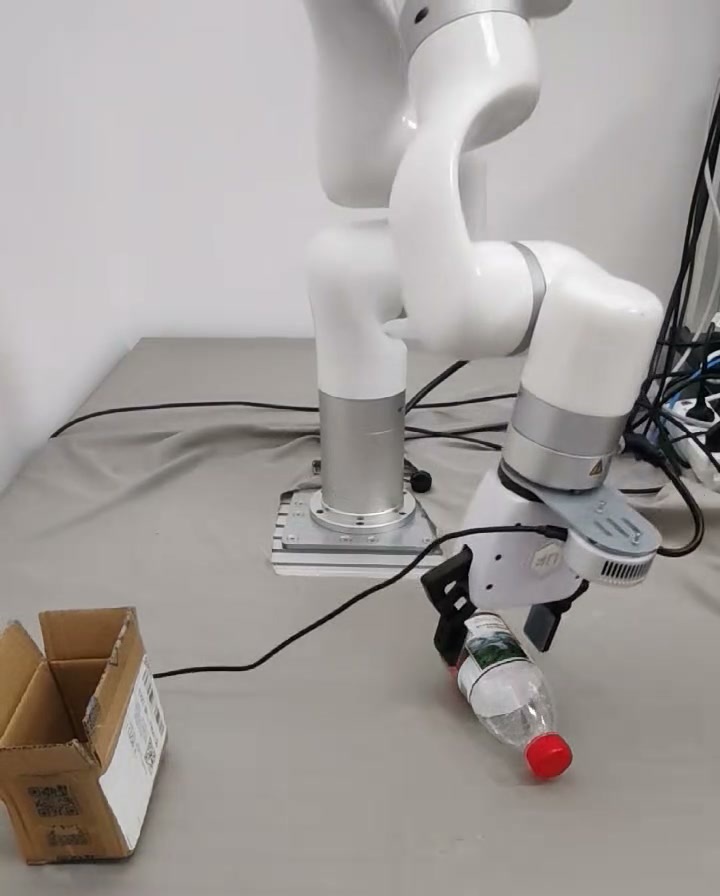}} & \parbox[c][0.064\textheight][c]{0.089\linewidth}{\centering\includegraphics[width=\linewidth]{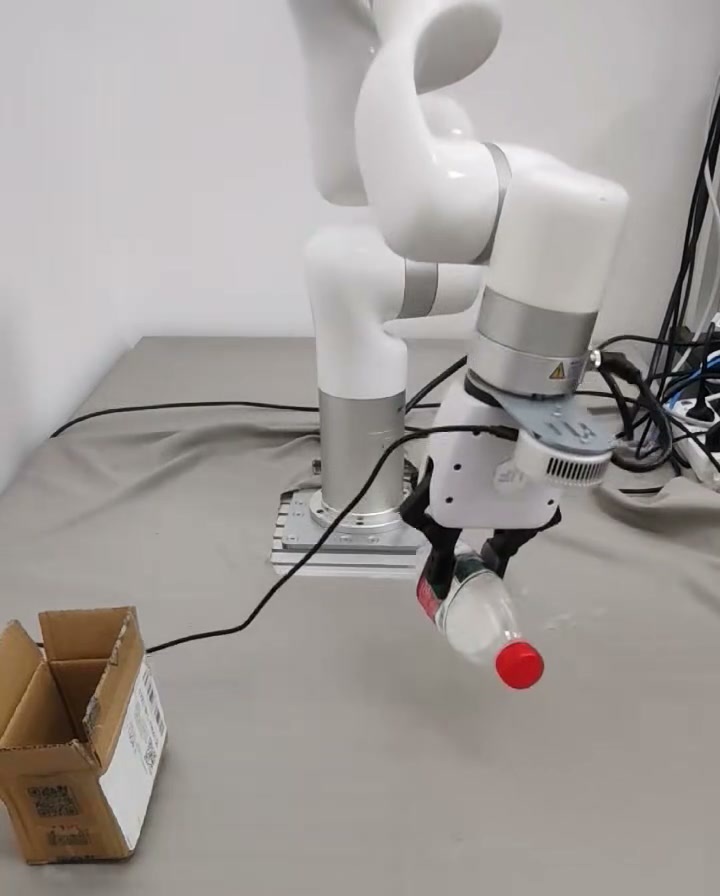}} & \parbox[c][0.064\textheight][c]{0.089\linewidth}{\centering\includegraphics[width=\linewidth]{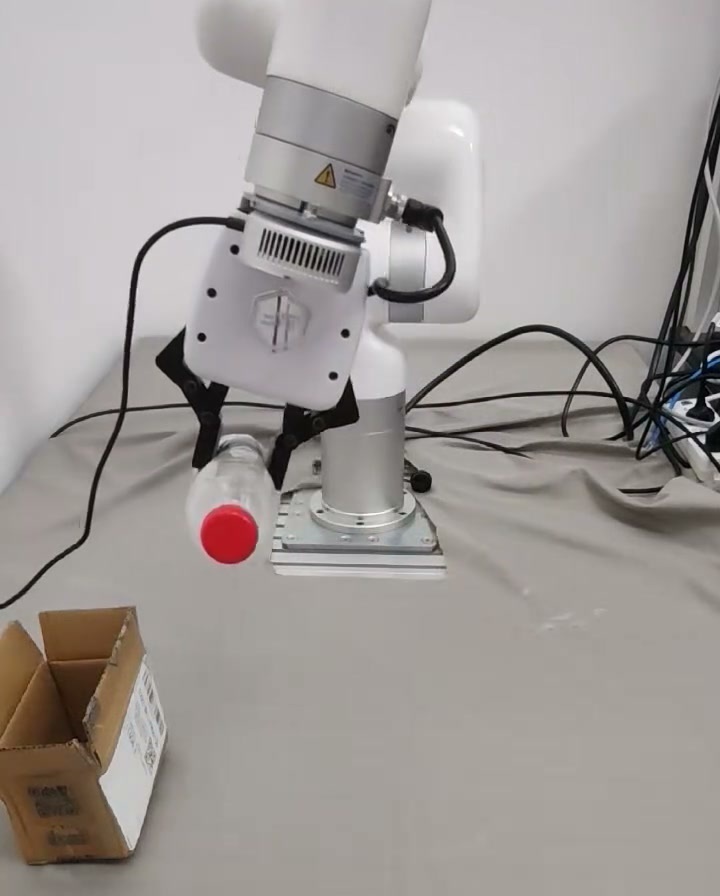}} & \parbox[c][0.064\textheight][c]{0.089\linewidth}{\centering\includegraphics[width=\linewidth]{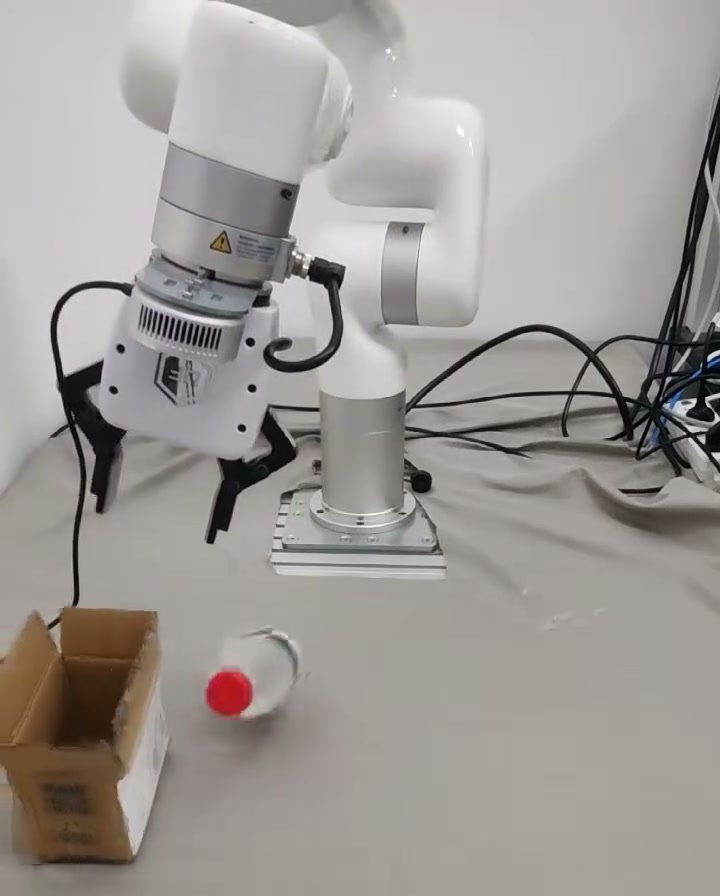}} & \parbox[c][0.064\textheight][c]{0.089\linewidth}{\centering\includegraphics[width=\linewidth]{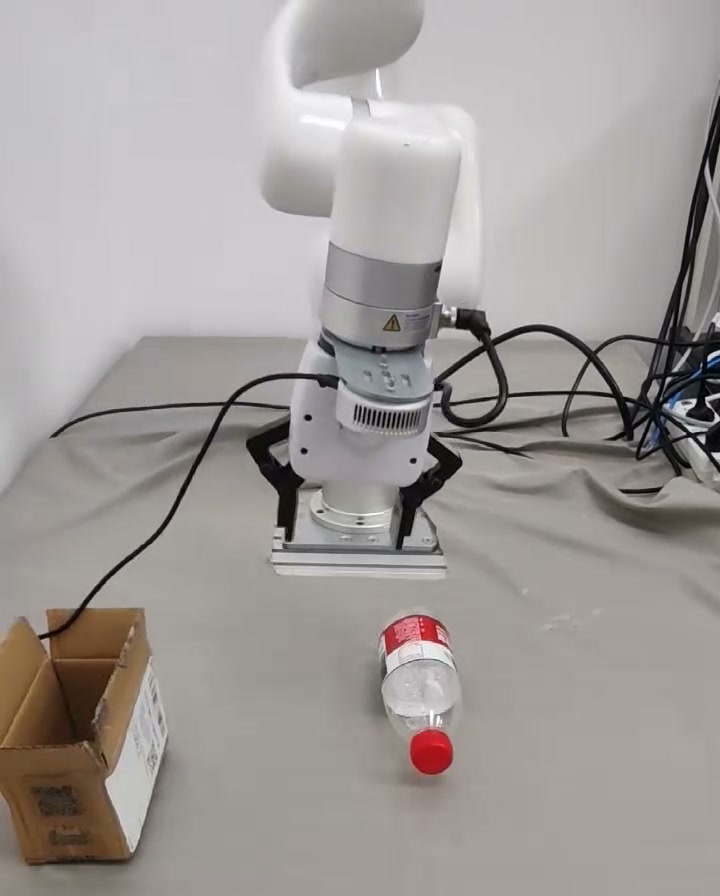}} & \parbox[c][0.064\textheight][c]{0.089\linewidth}{\centering\includegraphics[width=\linewidth]{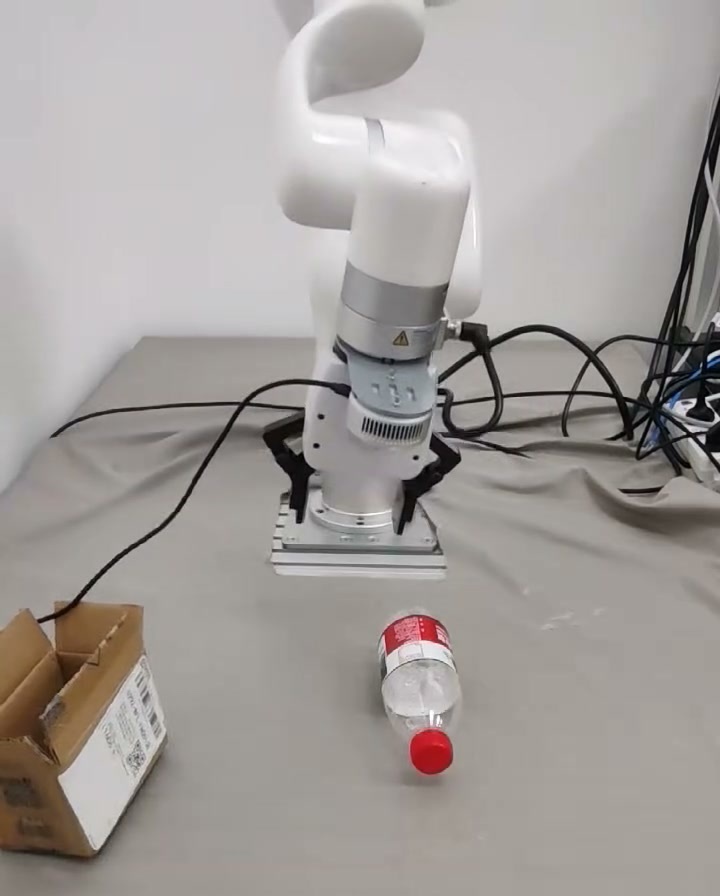}}}
\expandafter\newcommand\csname xarmrow-minivla-ECL\endcsname{\parbox[c][0.064\textheight][c]{0.089\linewidth}{\centering\includegraphics[width=\linewidth]{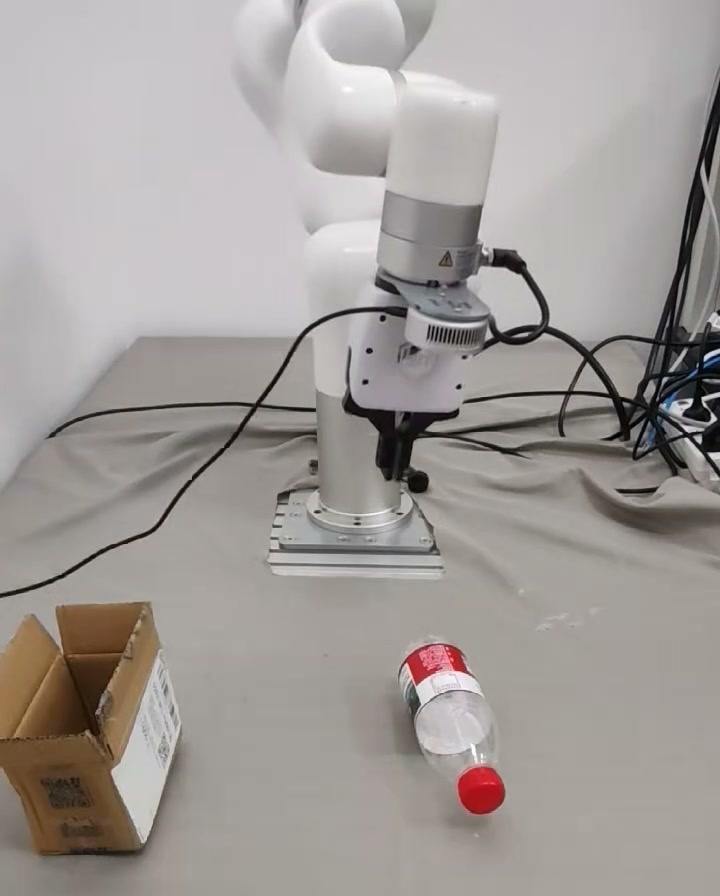}} & \parbox[c][0.064\textheight][c]{0.089\linewidth}{\centering\includegraphics[width=\linewidth]{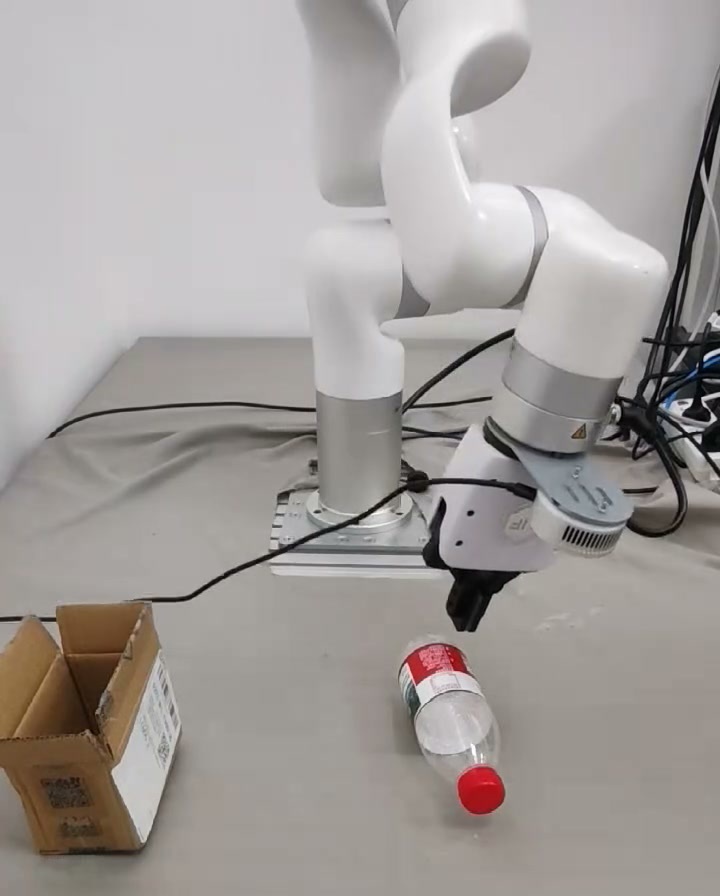}} & \parbox[c][0.064\textheight][c]{0.089\linewidth}{\centering\includegraphics[width=\linewidth]{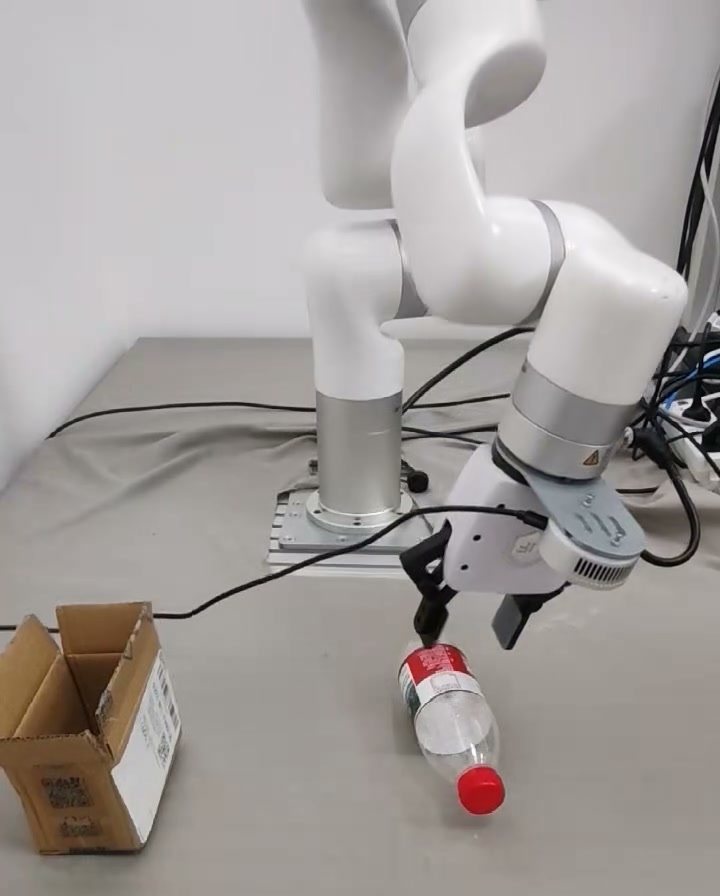}} & \parbox[c][0.064\textheight][c]{0.089\linewidth}{\centering\includegraphics[width=\linewidth]{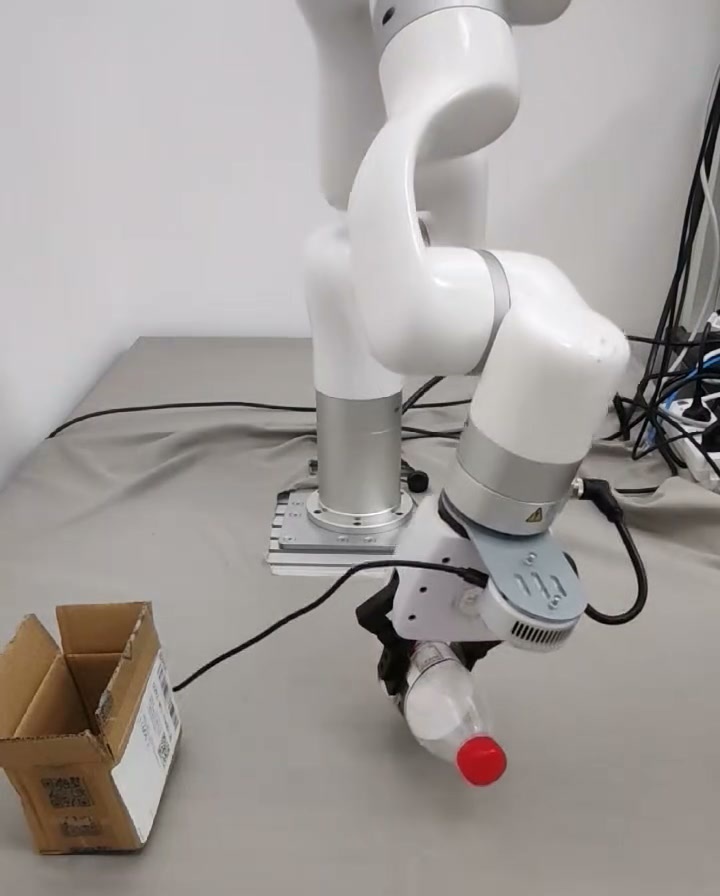}} & \parbox[c][0.064\textheight][c]{0.089\linewidth}{\centering\includegraphics[width=\linewidth]{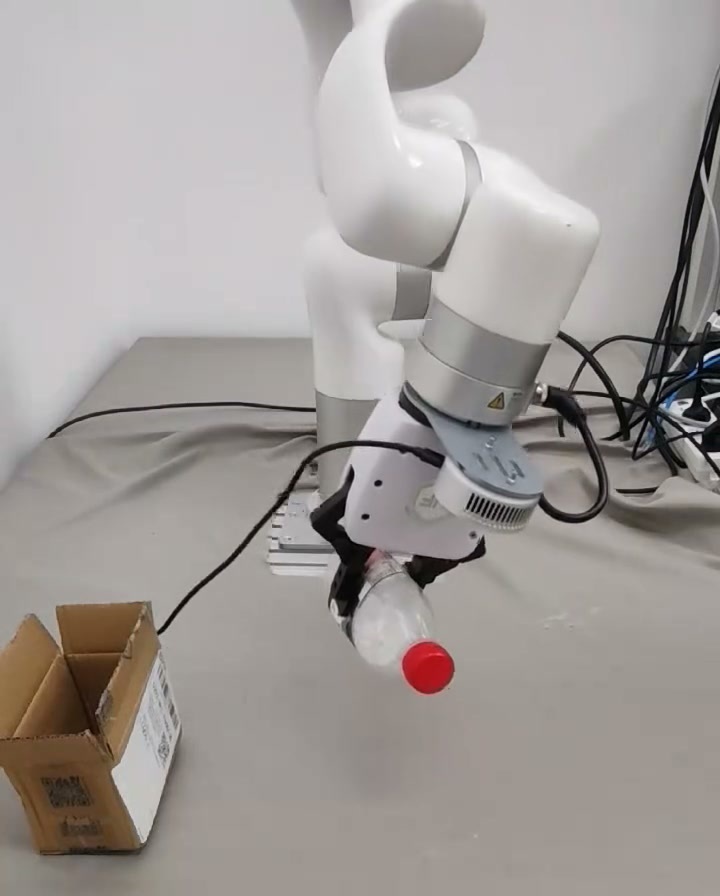}} & \parbox[c][0.064\textheight][c]{0.089\linewidth}{\centering\includegraphics[width=\linewidth]{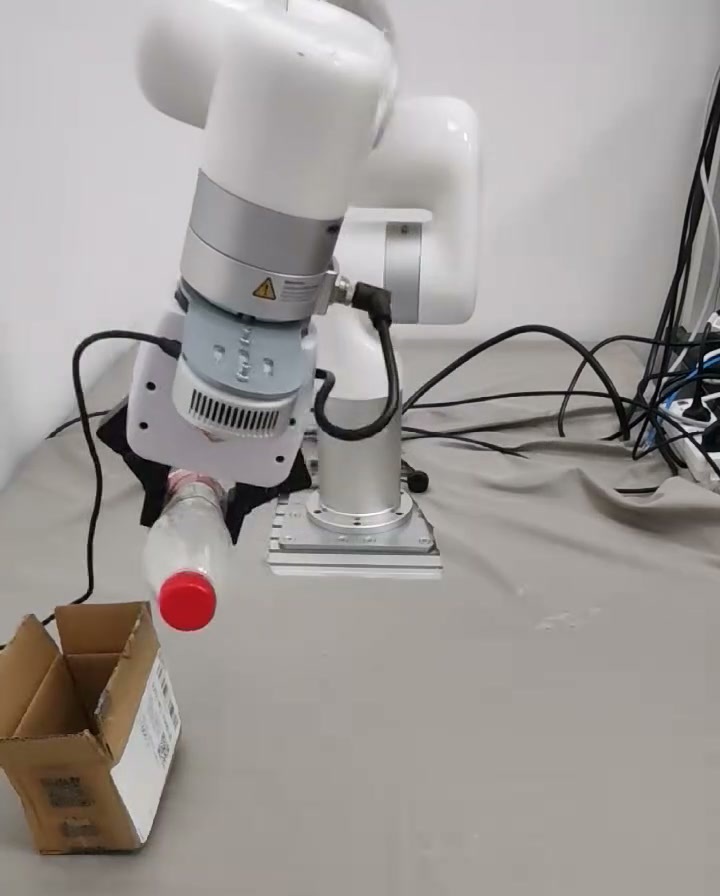}} & \parbox[c][0.064\textheight][c]{0.089\linewidth}{\centering\includegraphics[width=\linewidth]{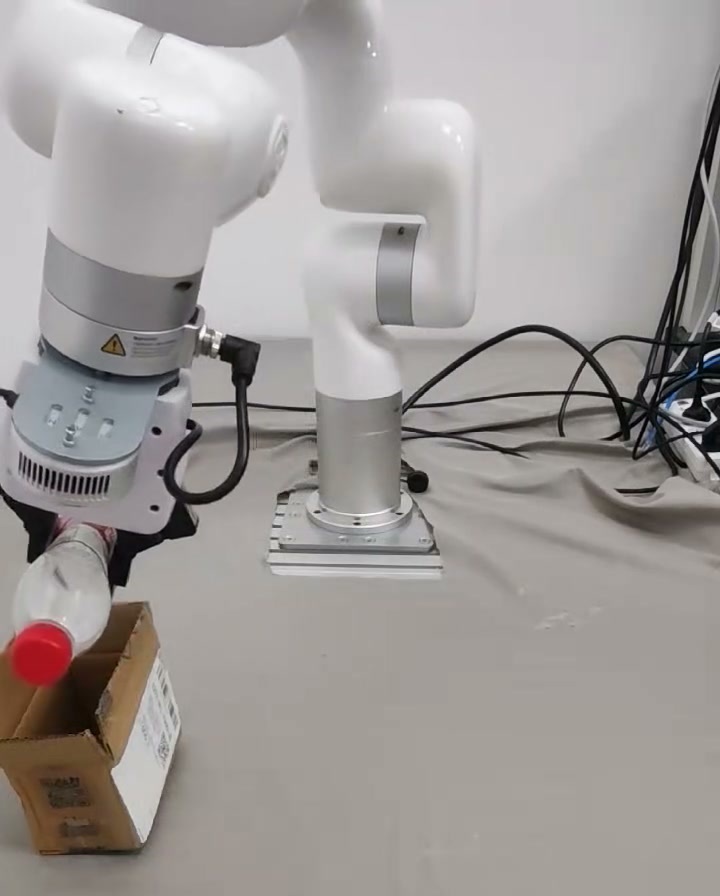}} & \parbox[c][0.064\textheight][c]{0.089\linewidth}{\centering\includegraphics[width=\linewidth]{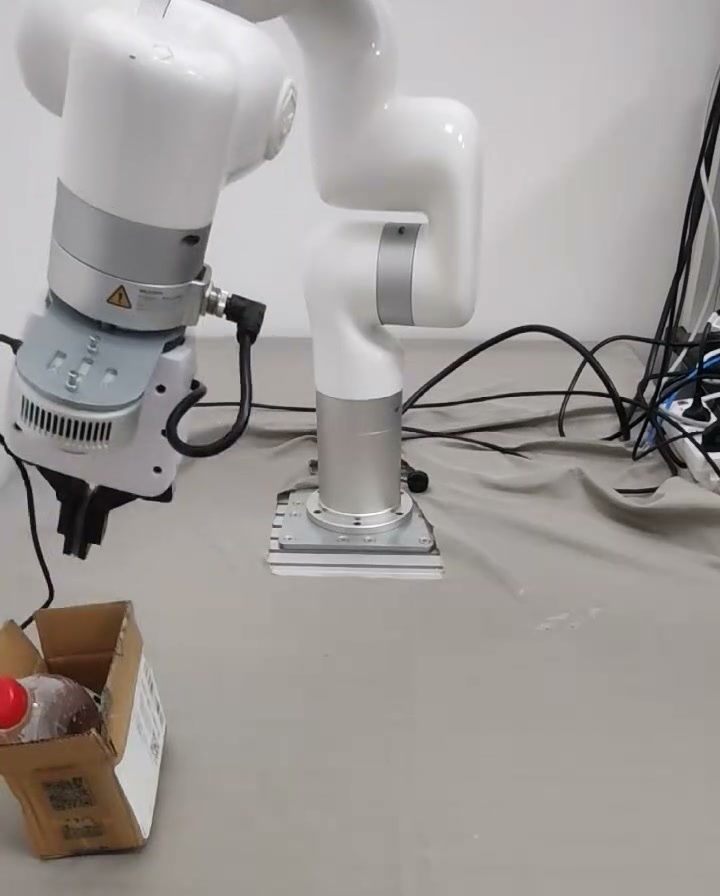}}}
\expandafter\newcommand\csname piperrow-pi05-APA-third\endcsname{\parbox[c][0.038\textheight][c]{0.079\linewidth}{\centering\includegraphics[width=\linewidth]{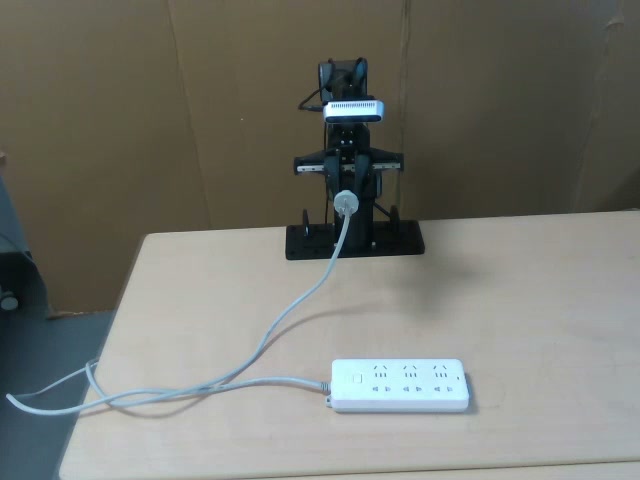}} & \parbox[c][0.038\textheight][c]{0.079\linewidth}{\centering\includegraphics[width=\linewidth]{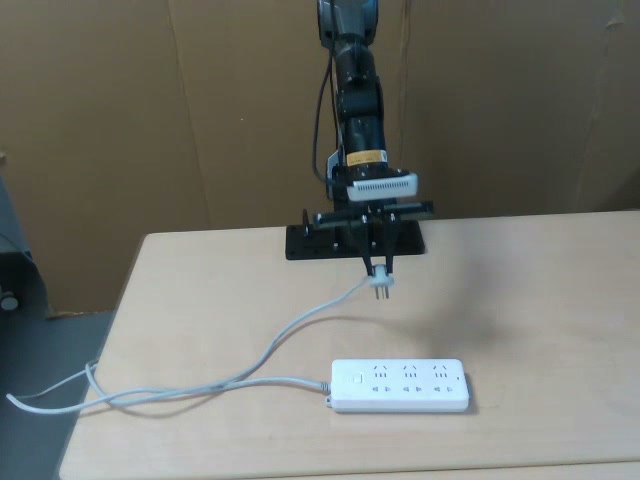}} & \parbox[c][0.038\textheight][c]{0.079\linewidth}{\centering\includegraphics[width=\linewidth]{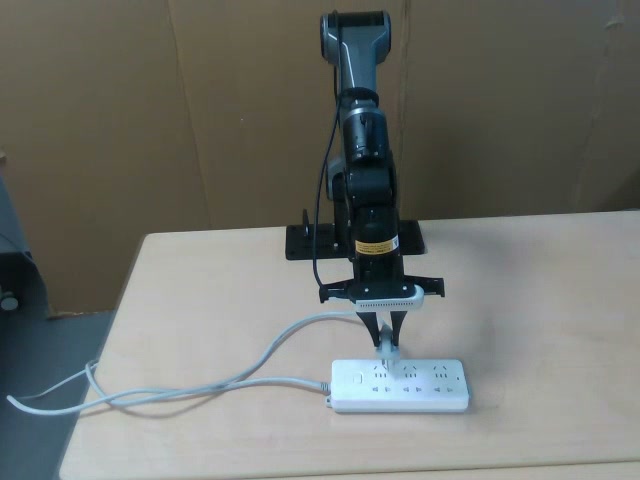}} & \parbox[c][0.038\textheight][c]{0.079\linewidth}{\centering\includegraphics[width=\linewidth]{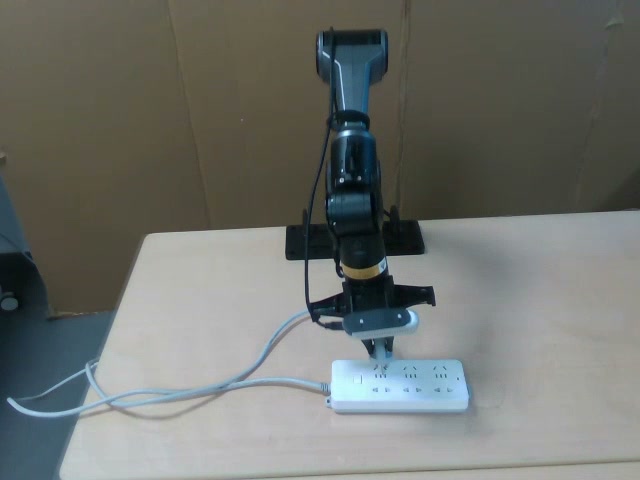}} & \parbox[c][0.038\textheight][c]{0.079\linewidth}{\centering\includegraphics[width=\linewidth]{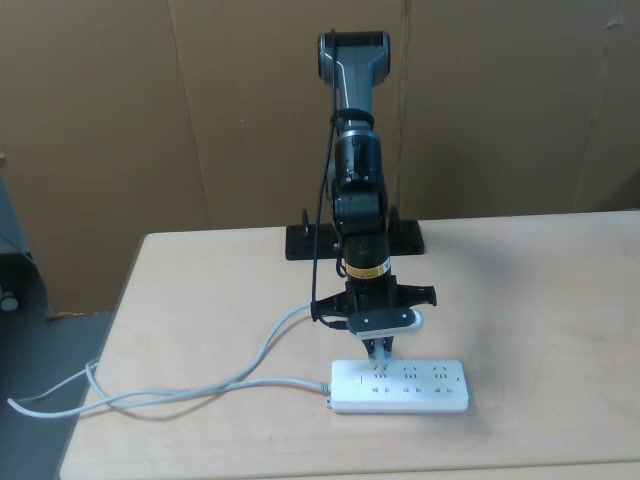}} & \parbox[c][0.038\textheight][c]{0.079\linewidth}{\centering\includegraphics[width=\linewidth]{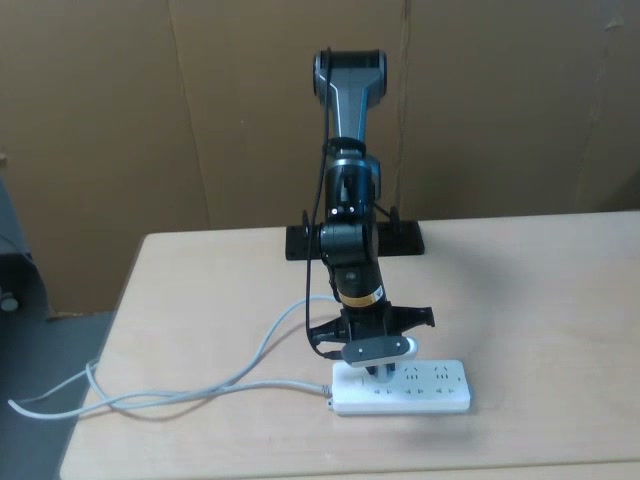}} & \parbox[c][0.038\textheight][c]{0.079\linewidth}{\centering\includegraphics[width=\linewidth]{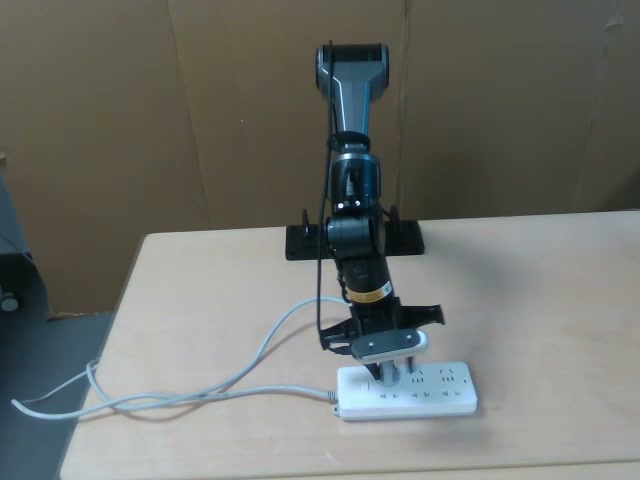}} & \parbox[c][0.038\textheight][c]{0.079\linewidth}{\centering\includegraphics[width=\linewidth]{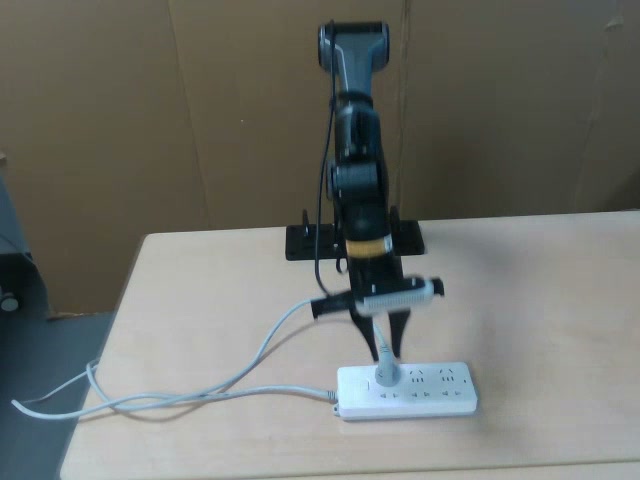}}}
\expandafter\newcommand\csname piperrow-pi05-APA-wrist\endcsname{\parbox[c][0.038\textheight][c]{0.079\linewidth}{\centering\includegraphics[width=\linewidth]{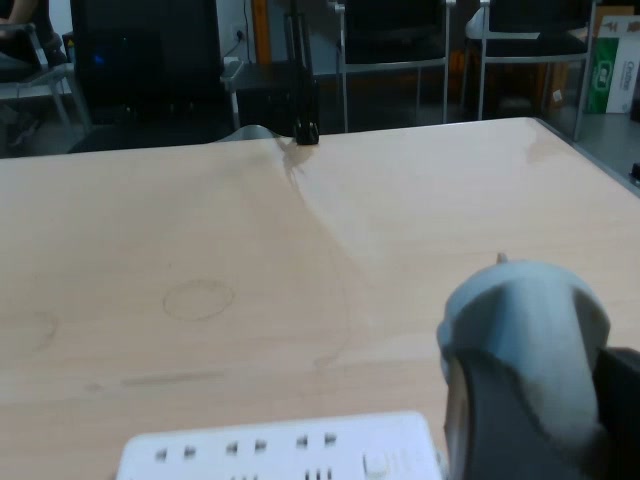}} & \parbox[c][0.038\textheight][c]{0.079\linewidth}{\centering\includegraphics[width=\linewidth]{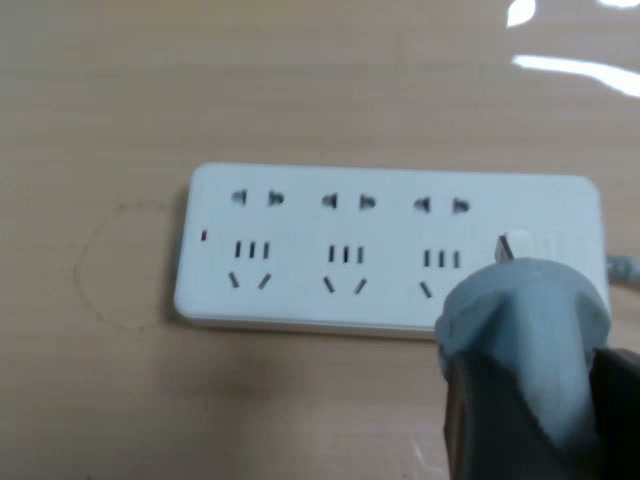}} & \parbox[c][0.038\textheight][c]{0.079\linewidth}{\centering\includegraphics[width=\linewidth]{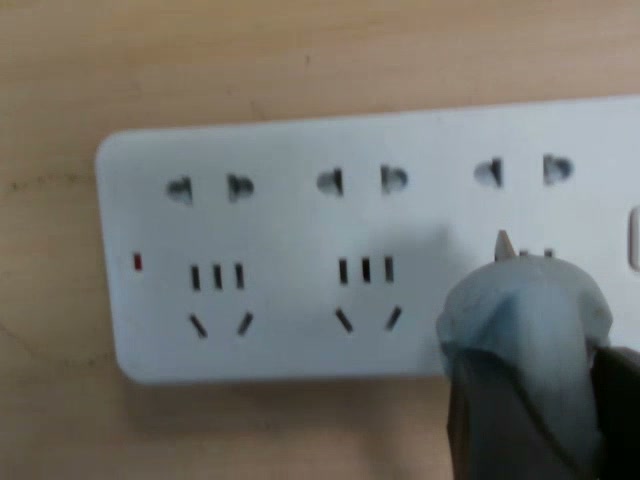}} & \parbox[c][0.038\textheight][c]{0.079\linewidth}{\centering\includegraphics[width=\linewidth]{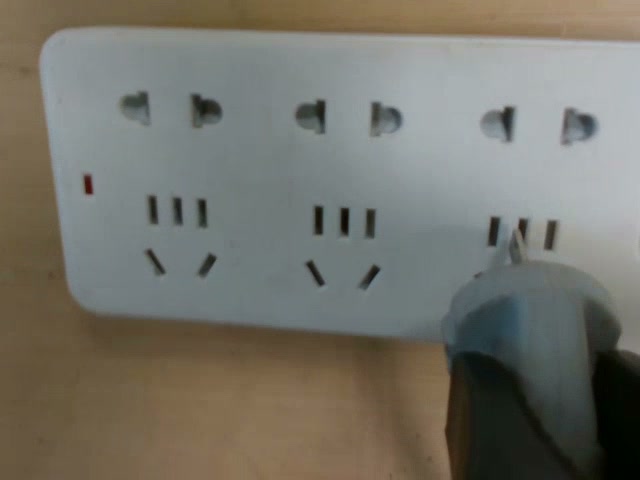}} & \parbox[c][0.038\textheight][c]{0.079\linewidth}{\centering\includegraphics[width=\linewidth]{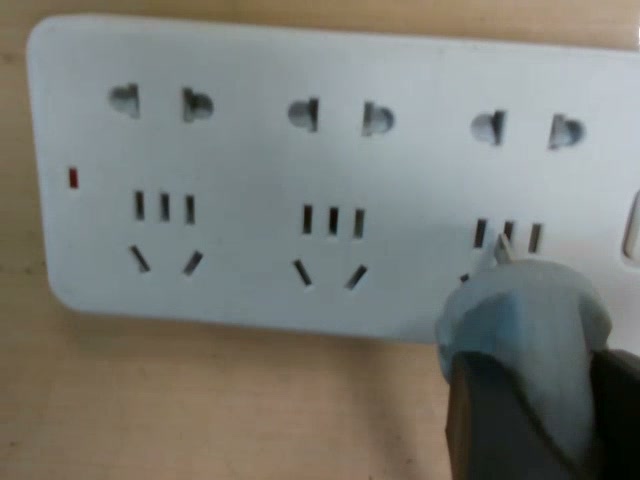}} & \parbox[c][0.038\textheight][c]{0.079\linewidth}{\centering\includegraphics[width=\linewidth]{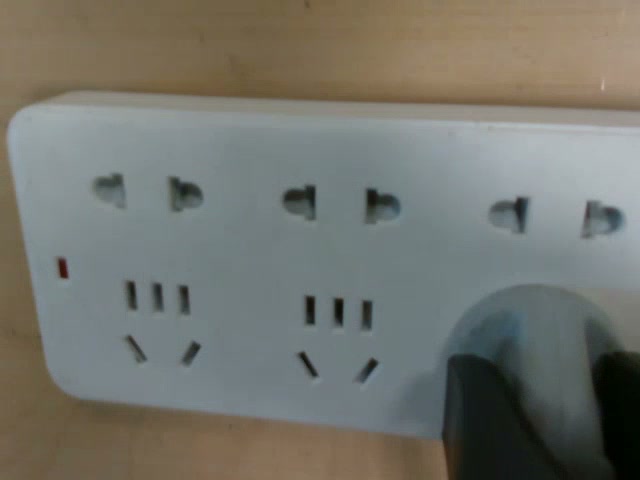}} & \parbox[c][0.038\textheight][c]{0.079\linewidth}{\centering\includegraphics[width=\linewidth]{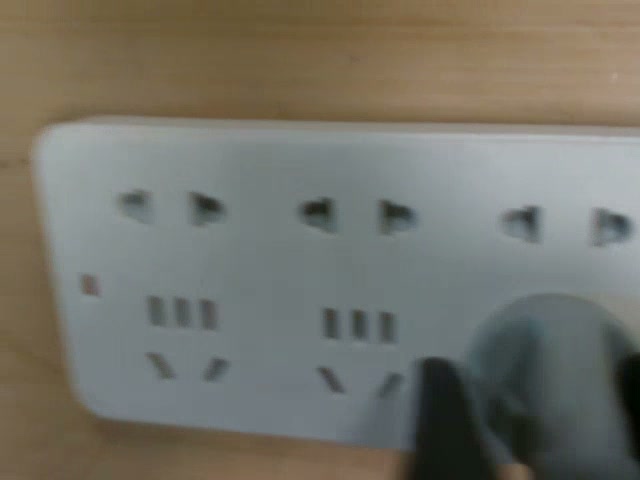}} & \parbox[c][0.038\textheight][c]{0.079\linewidth}{\centering\includegraphics[width=\linewidth]{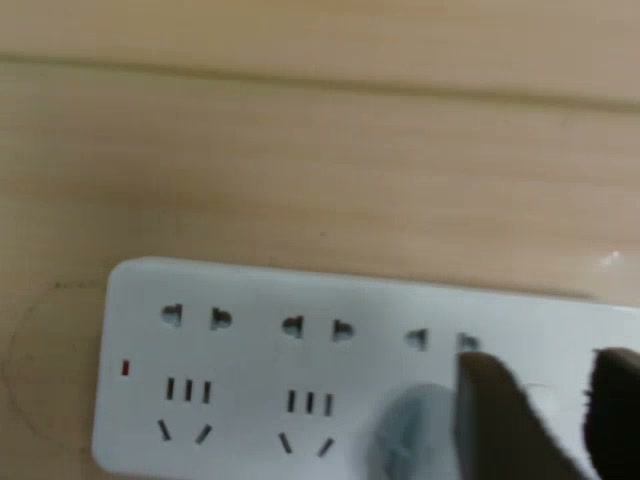}}}
\expandafter\newcommand\csname piperrow-pi05-ECL-third\endcsname{\parbox[c][0.038\textheight][c]{0.079\linewidth}{\centering\includegraphics[width=\linewidth]{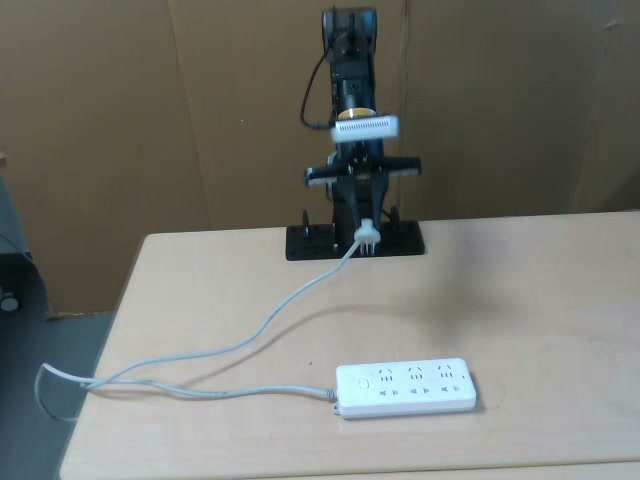}} & \parbox[c][0.038\textheight][c]{0.079\linewidth}{\centering\includegraphics[width=\linewidth]{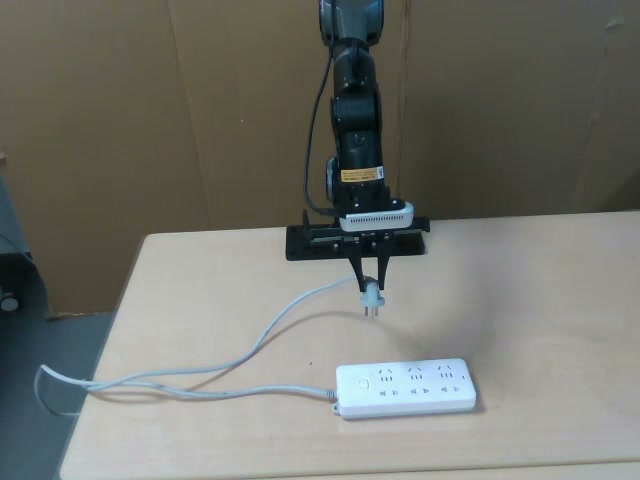}} & \parbox[c][0.038\textheight][c]{0.079\linewidth}{\centering\includegraphics[width=\linewidth]{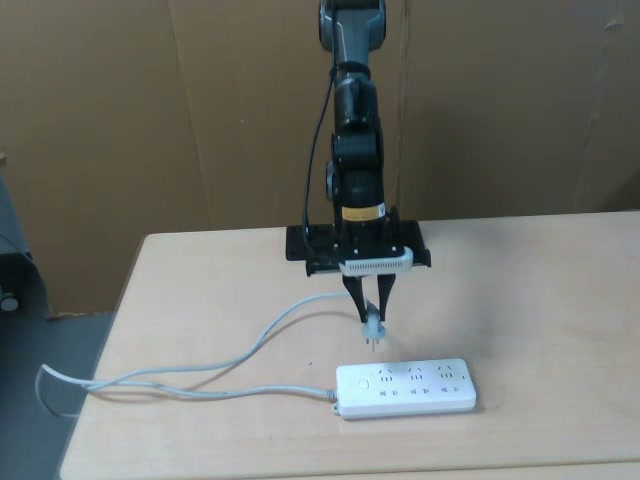}} & \parbox[c][0.038\textheight][c]{0.079\linewidth}{\centering\includegraphics[width=\linewidth]{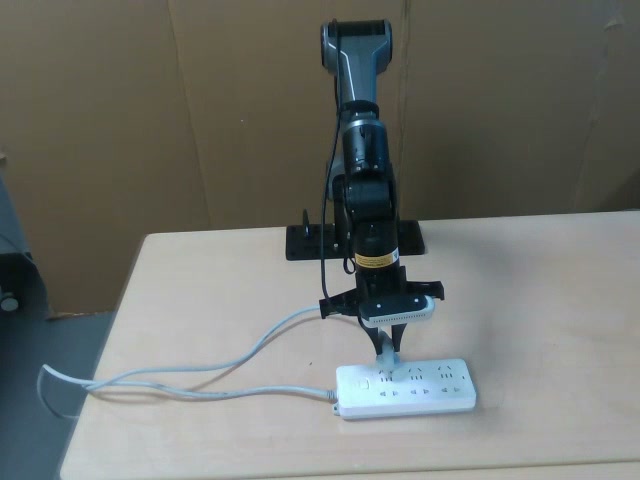}} & \parbox[c][0.038\textheight][c]{0.079\linewidth}{\centering\includegraphics[width=\linewidth]{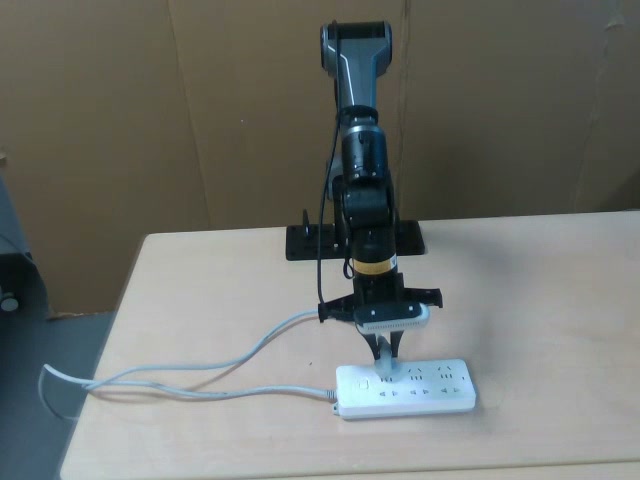}} & \parbox[c][0.038\textheight][c]{0.079\linewidth}{\centering\includegraphics[width=\linewidth]{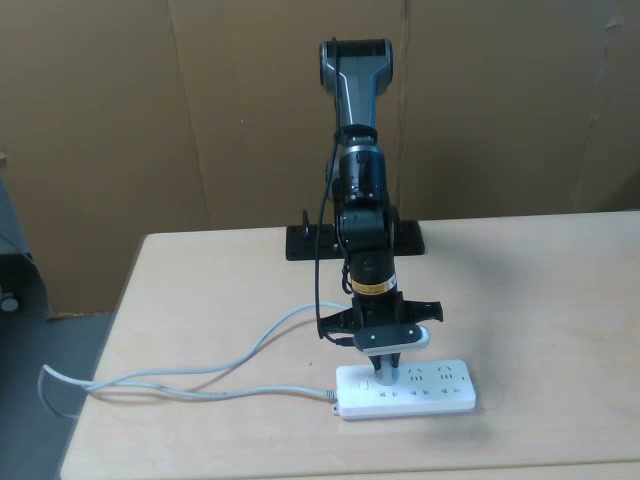}} & \parbox[c][0.038\textheight][c]{0.079\linewidth}{\centering\includegraphics[width=\linewidth]{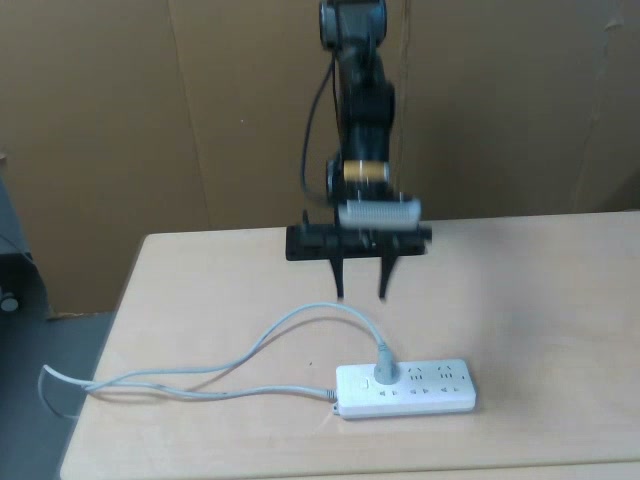}} & \parbox[c][0.038\textheight][c]{0.079\linewidth}{\centering\includegraphics[width=\linewidth]{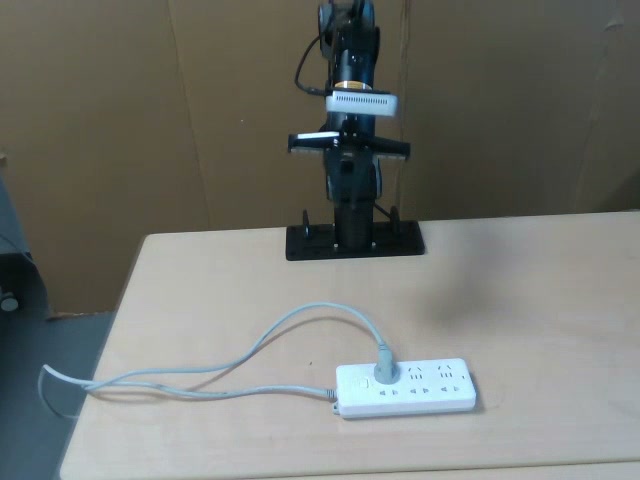}}}
\expandafter\newcommand\csname piperrow-pi05-ECL-wrist\endcsname{\parbox[c][0.038\textheight][c]{0.079\linewidth}{\centering\includegraphics[width=\linewidth]{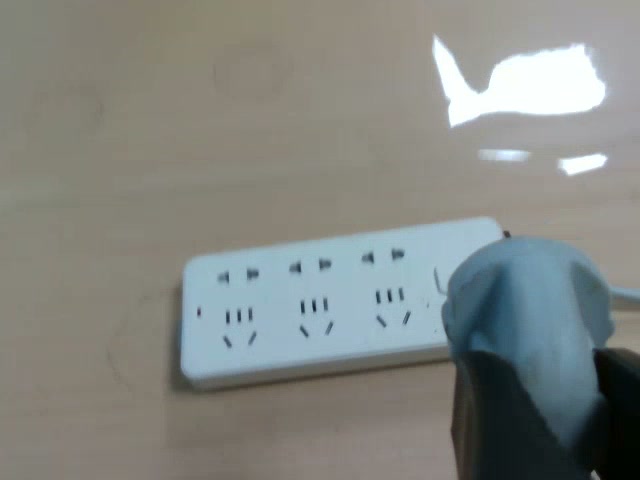}} & \parbox[c][0.038\textheight][c]{0.079\linewidth}{\centering\includegraphics[width=\linewidth]{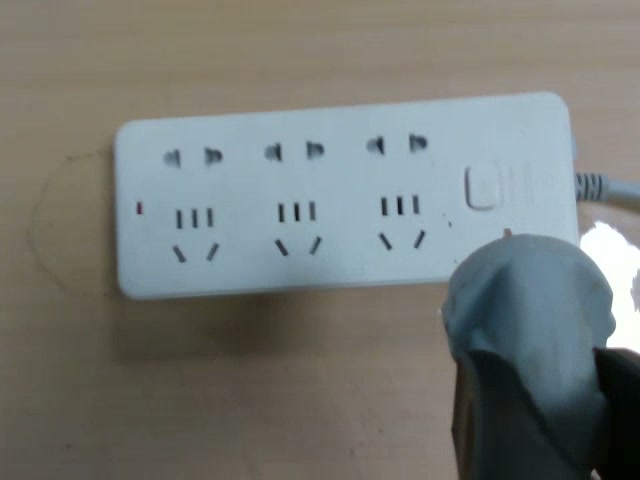}} & \parbox[c][0.038\textheight][c]{0.079\linewidth}{\centering\includegraphics[width=\linewidth]{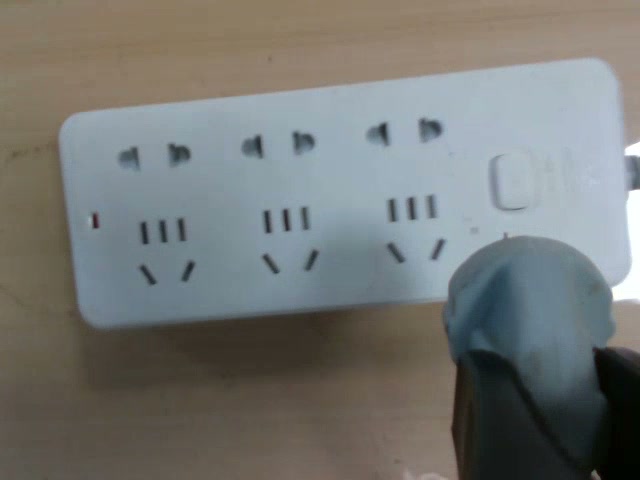}} & \parbox[c][0.038\textheight][c]{0.079\linewidth}{\centering\includegraphics[width=\linewidth]{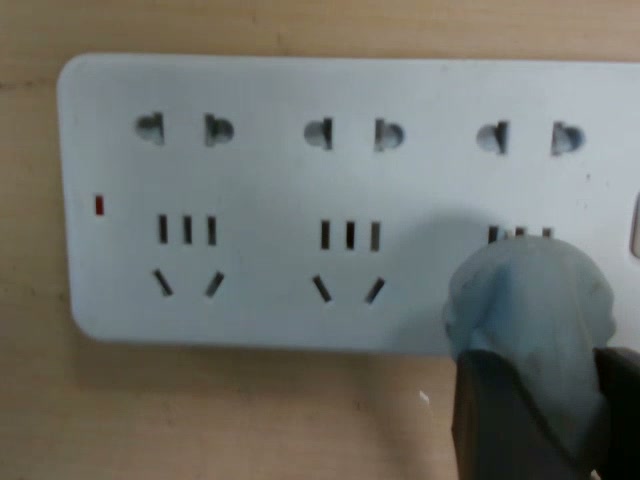}} & \parbox[c][0.038\textheight][c]{0.079\linewidth}{\centering\includegraphics[width=\linewidth]{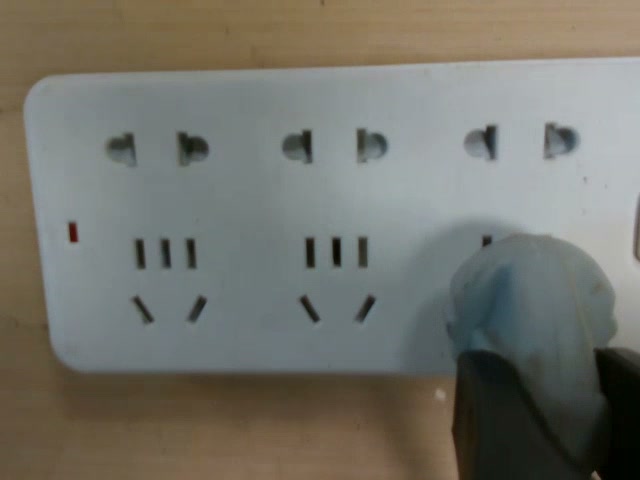}} & \parbox[c][0.038\textheight][c]{0.079\linewidth}{\centering\includegraphics[width=\linewidth]{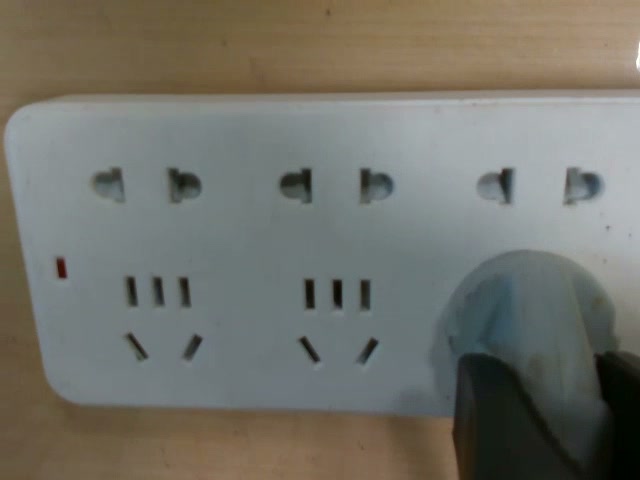}} & \parbox[c][0.038\textheight][c]{0.079\linewidth}{\centering\includegraphics[width=\linewidth]{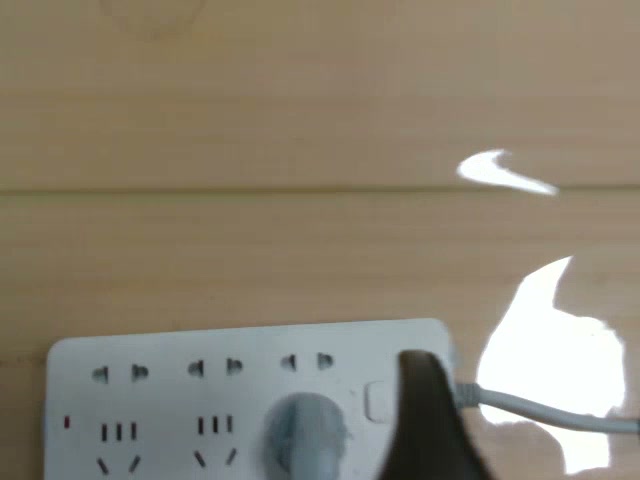}} & \parbox[c][0.038\textheight][c]{0.079\linewidth}{\centering\includegraphics[width=\linewidth]{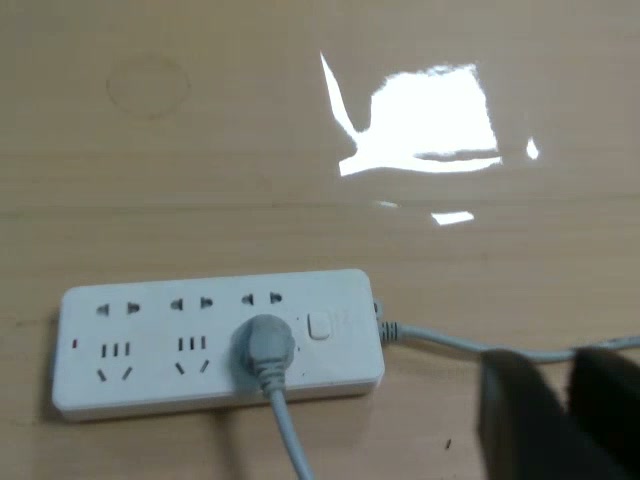}}}
\expandafter\newcommand\csname xarmrow-pi05-APA\endcsname{\parbox[c][0.064\textheight][c]{0.089\linewidth}{\centering\includegraphics[width=\linewidth]{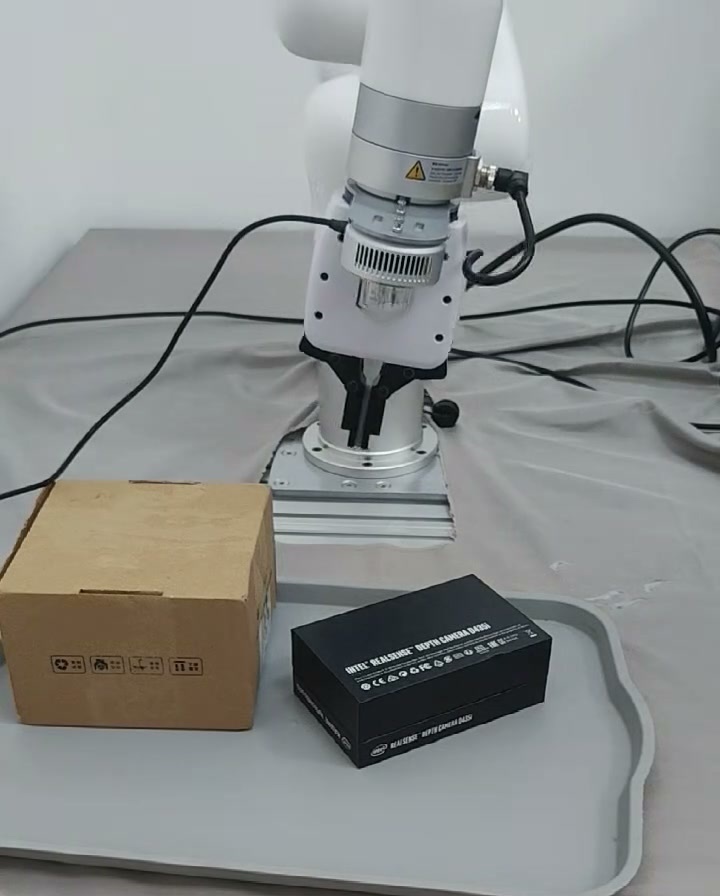}} & \parbox[c][0.064\textheight][c]{0.089\linewidth}{\centering\includegraphics[width=\linewidth]{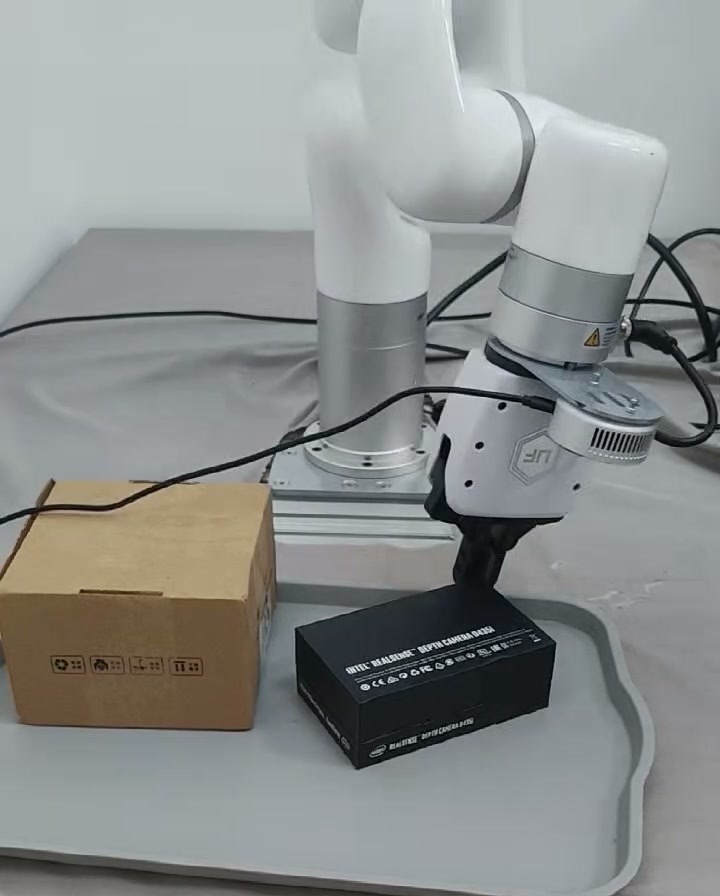}} & \parbox[c][0.064\textheight][c]{0.089\linewidth}{\centering\includegraphics[width=\linewidth]{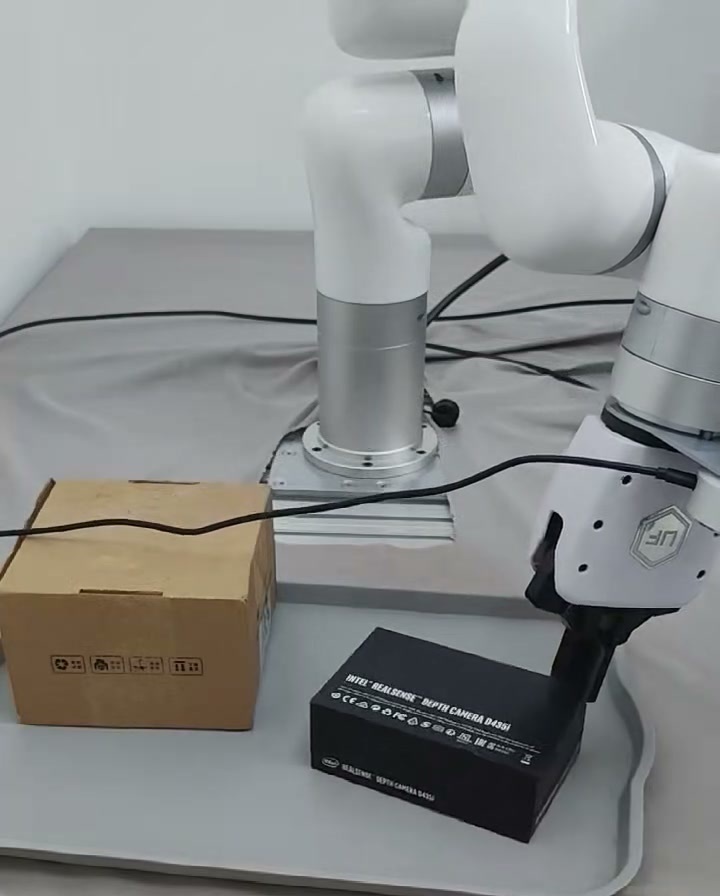}} & \parbox[c][0.064\textheight][c]{0.089\linewidth}{\centering\includegraphics[width=\linewidth]{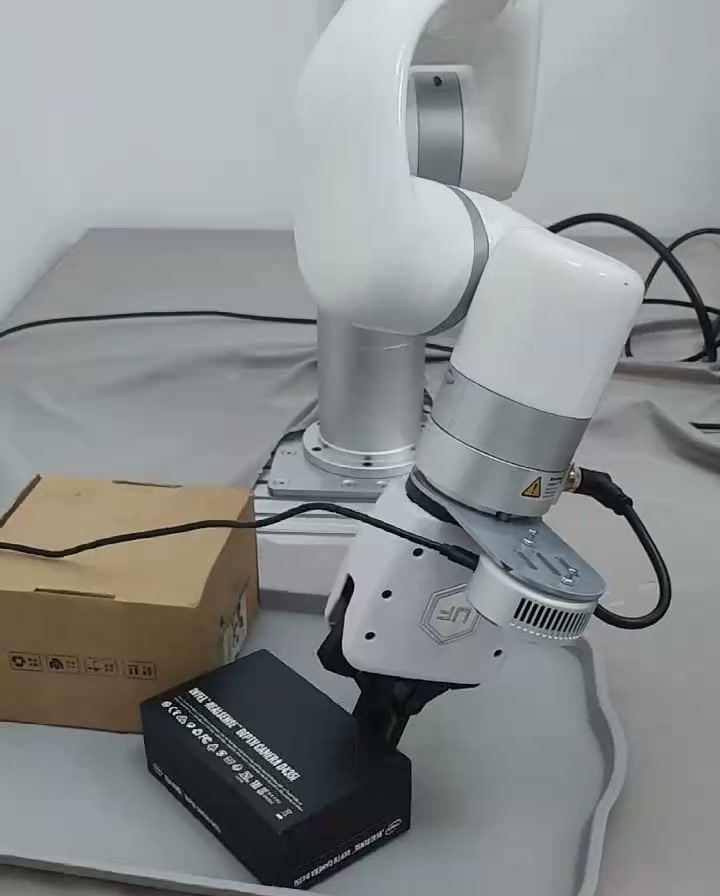}} & \parbox[c][0.064\textheight][c]{0.089\linewidth}{\centering\includegraphics[width=\linewidth]{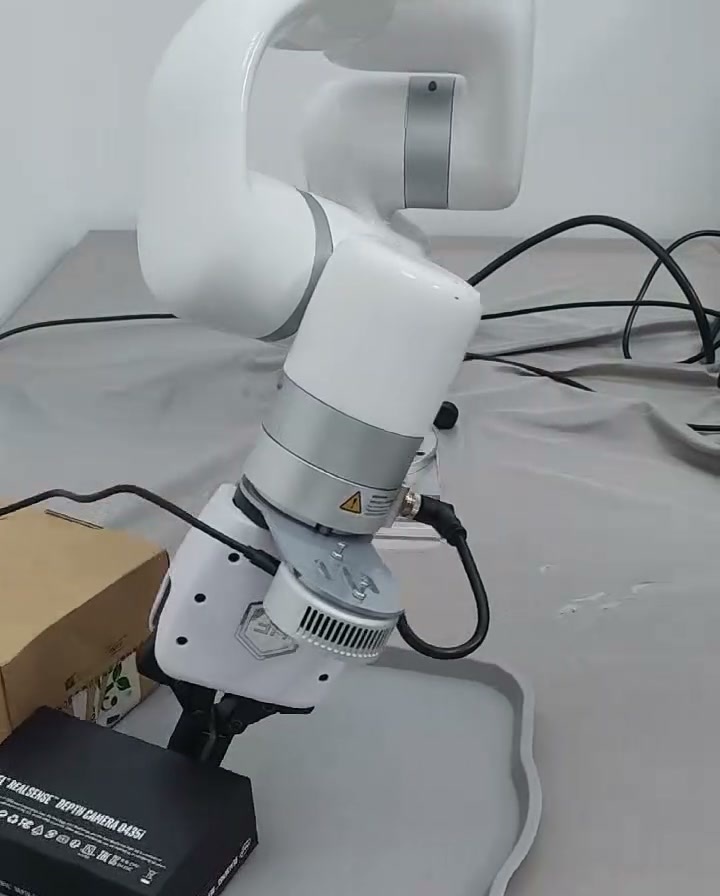}} & \parbox[c][0.064\textheight][c]{0.089\linewidth}{\centering\includegraphics[width=\linewidth]{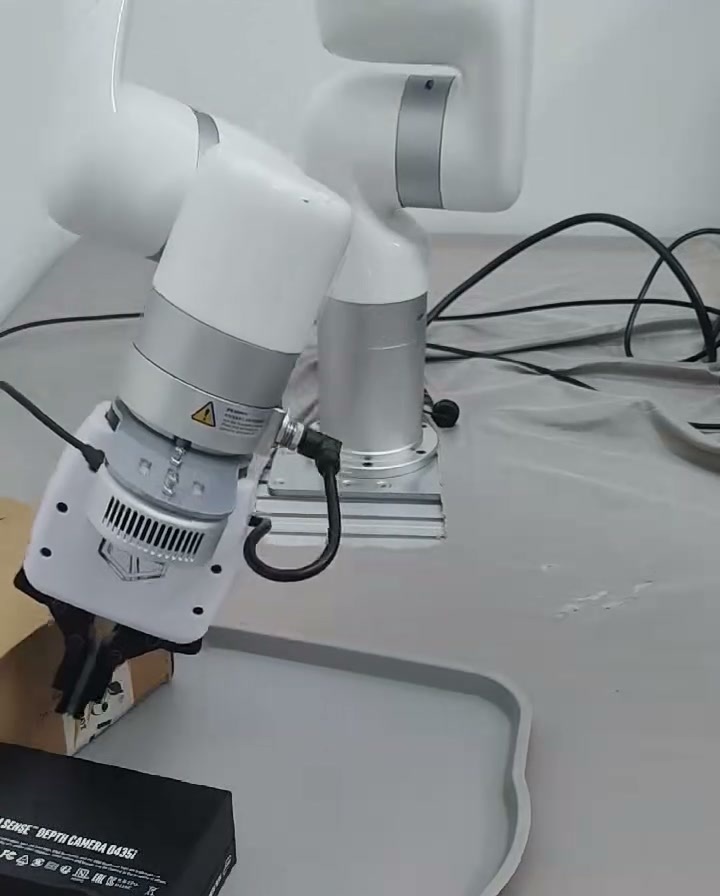}} & \parbox[c][0.064\textheight][c]{0.089\linewidth}{\centering\includegraphics[width=\linewidth]{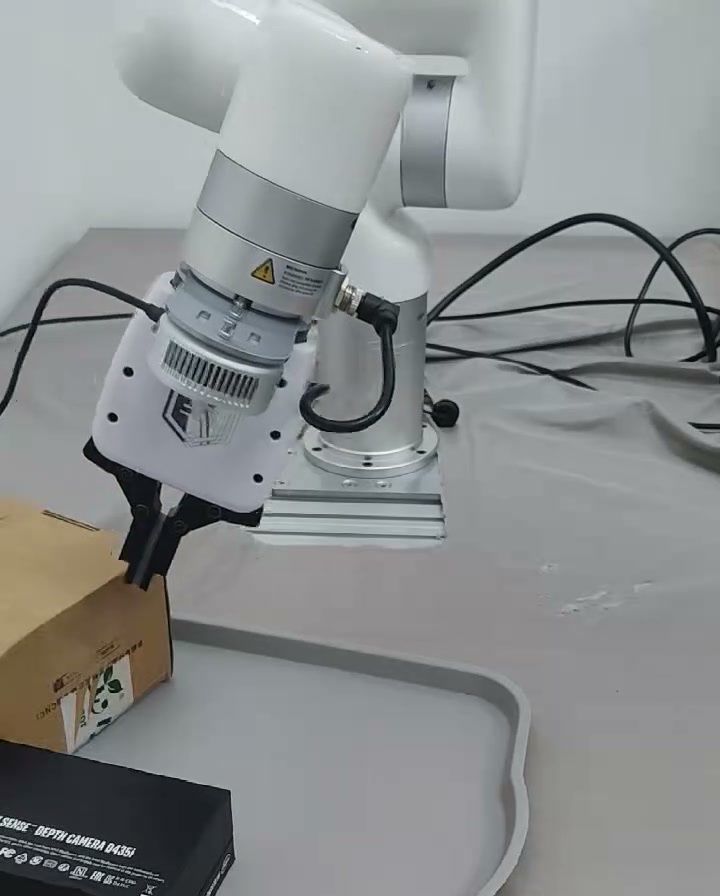}} & \parbox[c][0.064\textheight][c]{0.089\linewidth}{\centering\includegraphics[width=\linewidth]{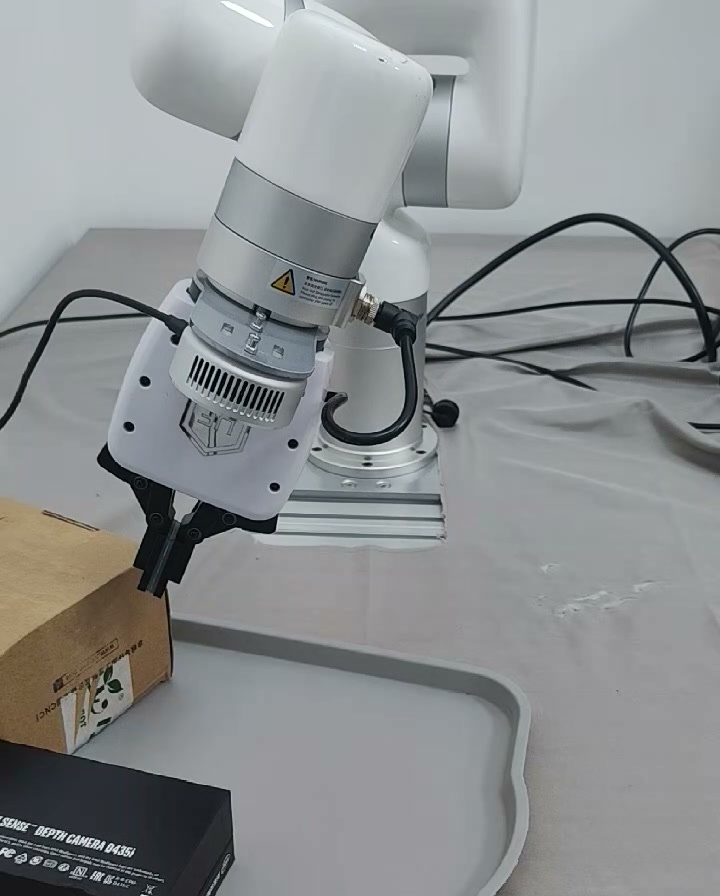}}}
\expandafter\newcommand\csname xarmrow-pi05-ECL\endcsname{\parbox[c][0.064\textheight][c]{0.089\linewidth}{\centering\includegraphics[width=\linewidth]{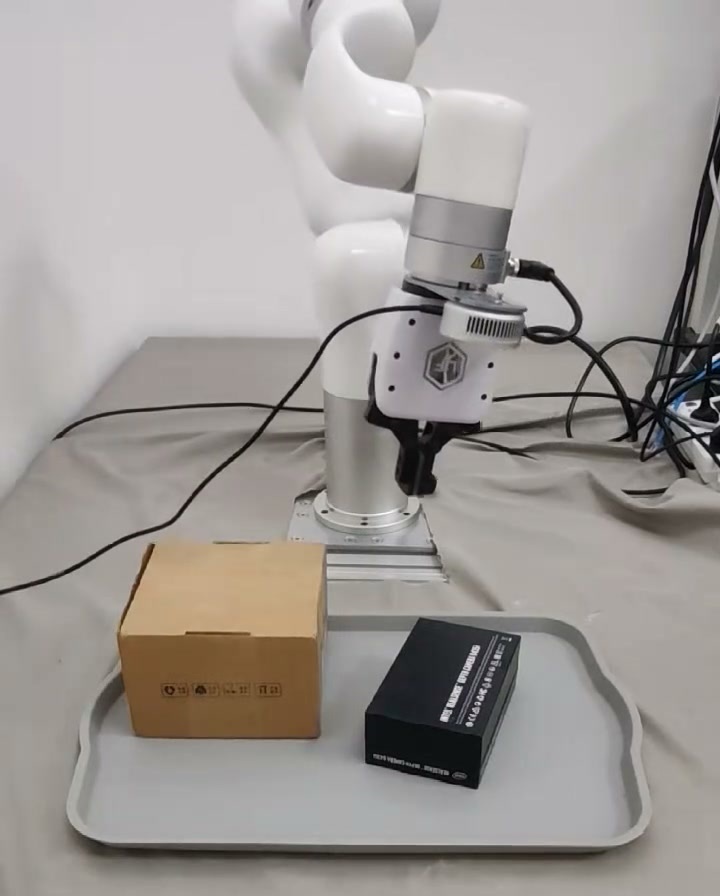}} & \parbox[c][0.064\textheight][c]{0.089\linewidth}{\centering\includegraphics[width=\linewidth]{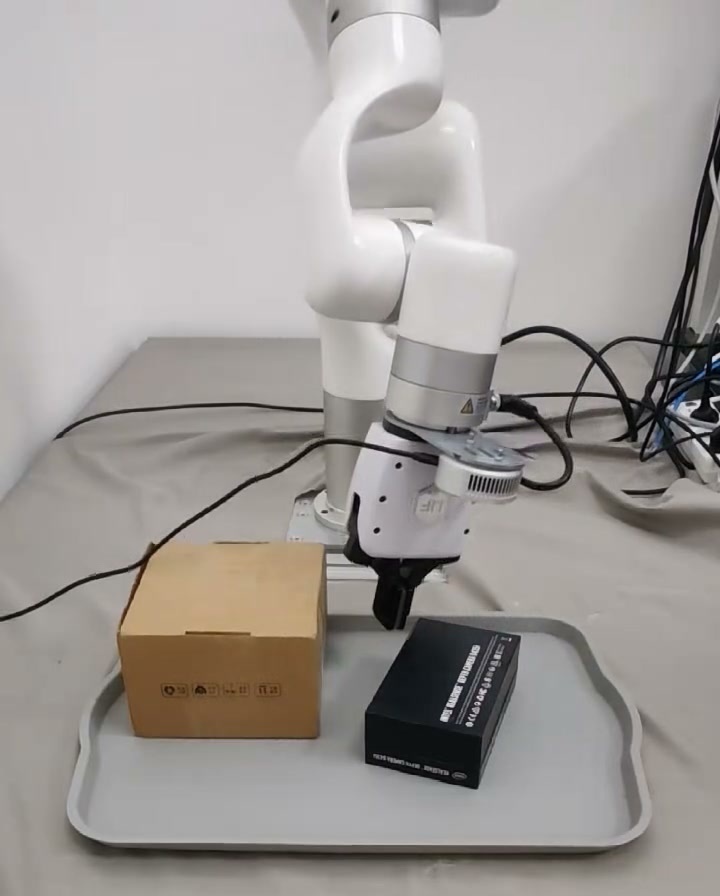}} & \parbox[c][0.064\textheight][c]{0.089\linewidth}{\centering\includegraphics[width=\linewidth]{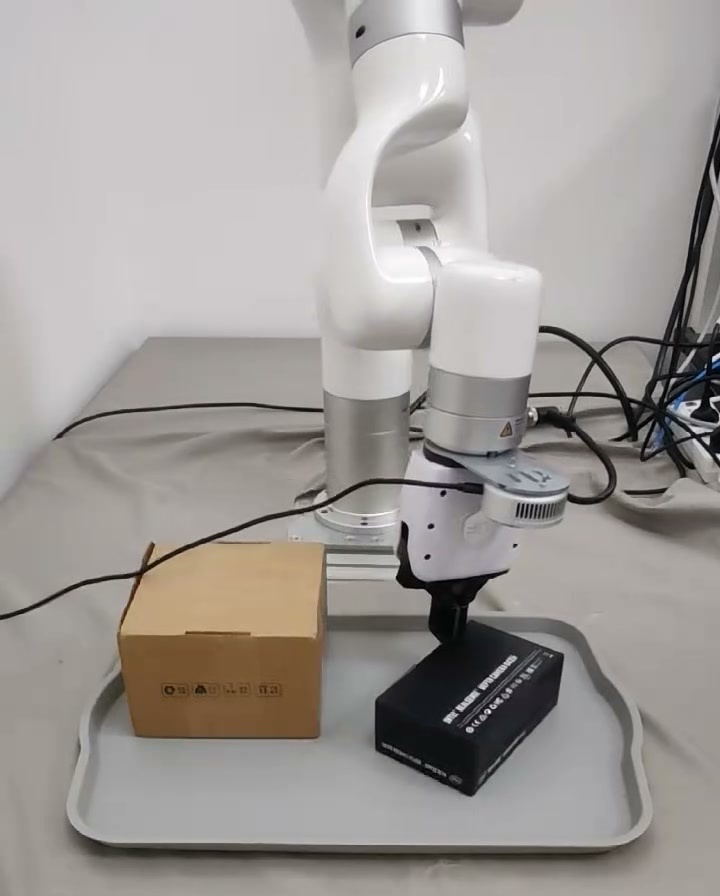}} & \parbox[c][0.064\textheight][c]{0.089\linewidth}{\centering\includegraphics[width=\linewidth]{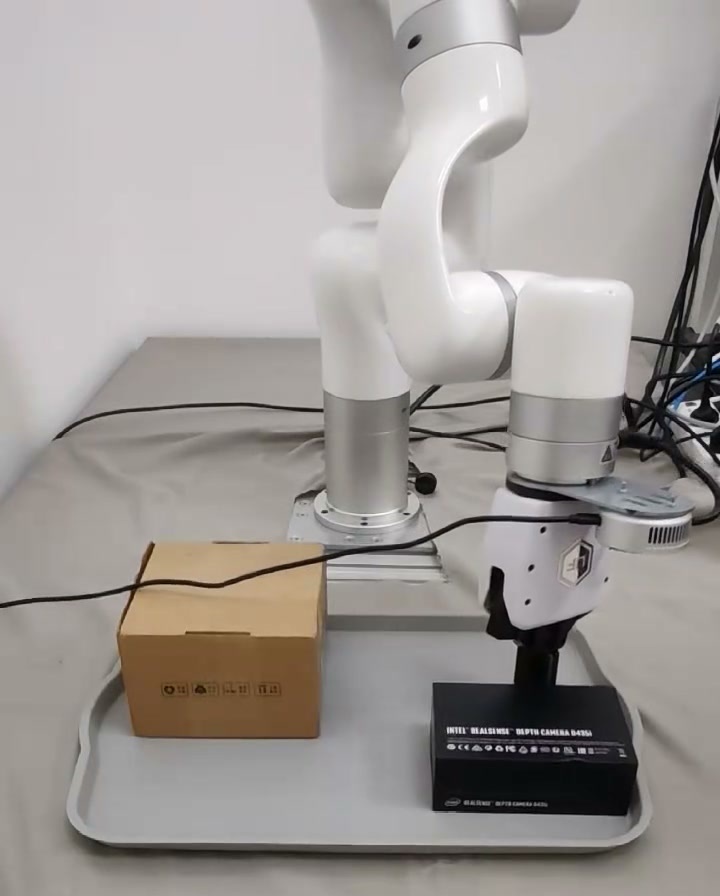}} & \parbox[c][0.064\textheight][c]{0.089\linewidth}{\centering\includegraphics[width=\linewidth]{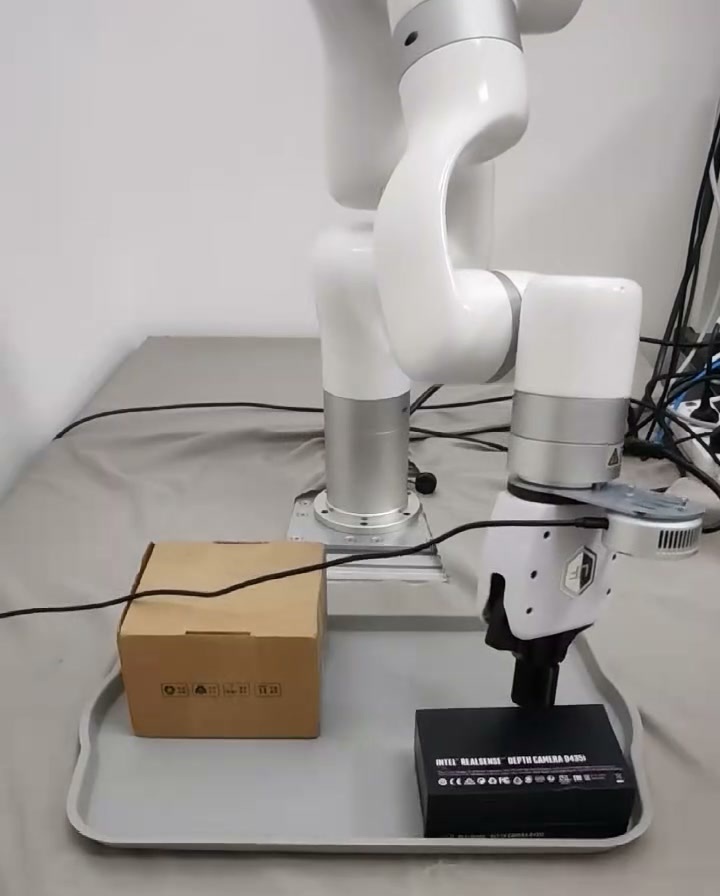}} & \parbox[c][0.064\textheight][c]{0.089\linewidth}{\centering\includegraphics[width=\linewidth]{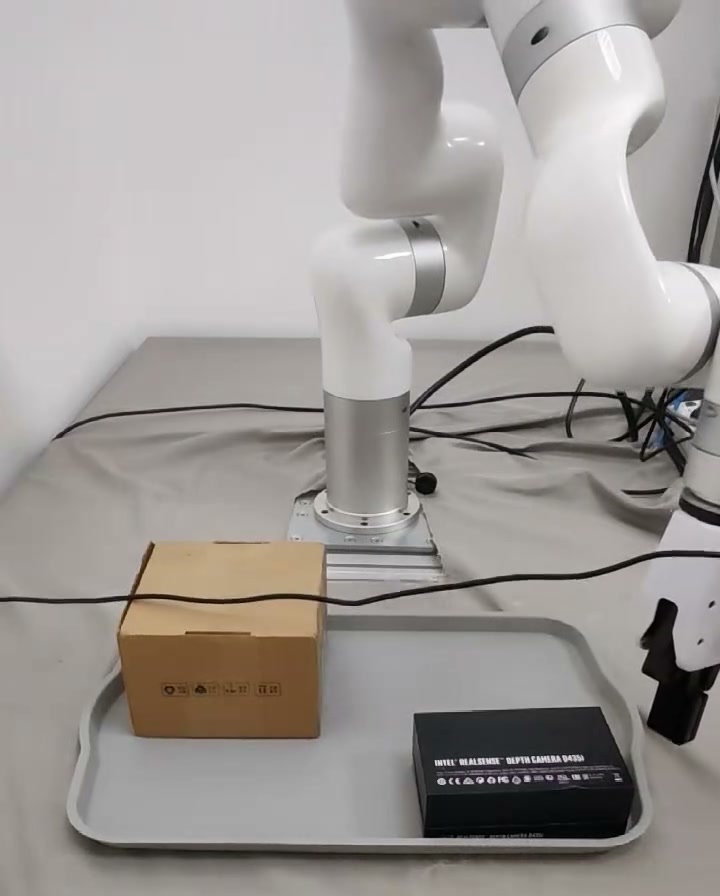}} & \parbox[c][0.064\textheight][c]{0.089\linewidth}{\centering\includegraphics[width=\linewidth]{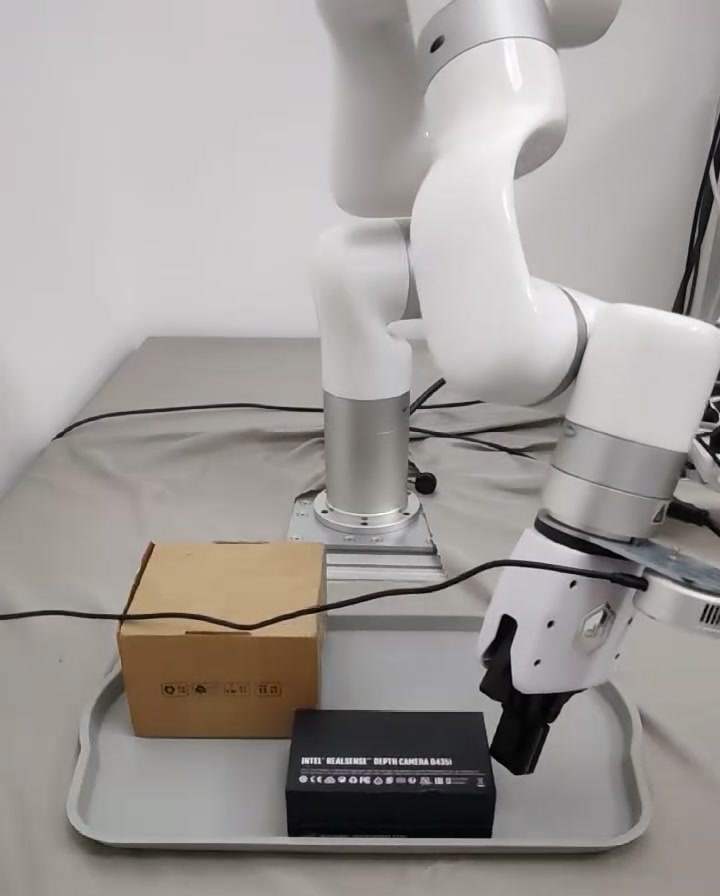}} & \parbox[c][0.064\textheight][c]{0.089\linewidth}{\centering\includegraphics[width=\linewidth]{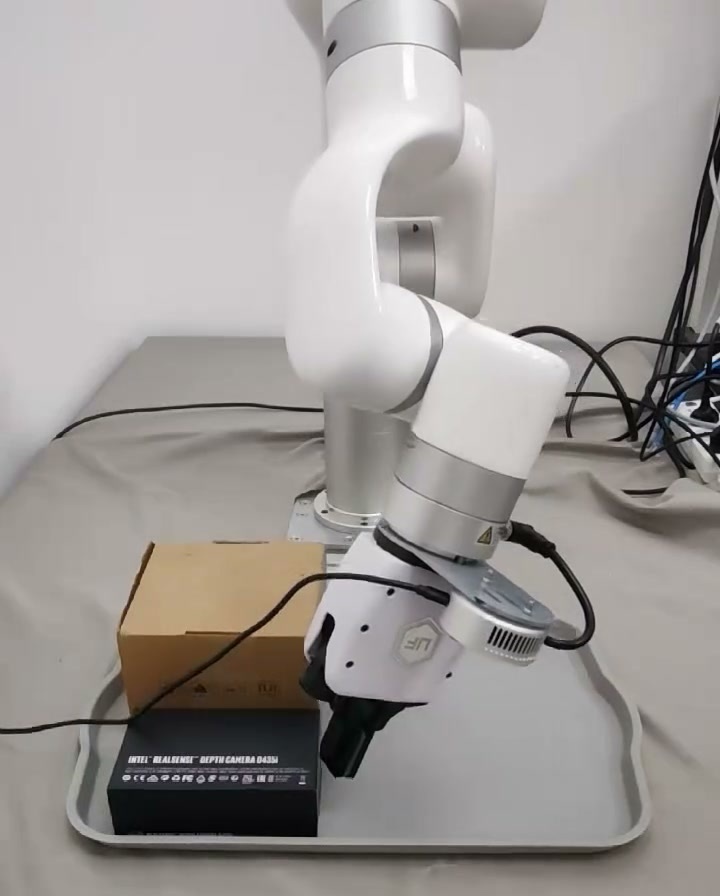}}}

\subsection{Simulation Rollouts}
\label{app:simulation_rollouts}

\paragraph{Evaluation configuration.}
Table~\ref{tab:app_sim_rollout_protocol} fixes the protocol used for both
backbones. Each task is evaluated for 675 episodes (225 episodes for each of
three seeds), and every episode terminates at success, failure, or the fixed
evaluation horizon. For $\pi_{0.5}$, we use the official OpenPI LIBERO configuration: a
10-step action horizon, batch size 256, learning rate $5\times10^{-5}$, 10k
warm-up steps, 30k optimization steps, and EMA decay 0.999.
The evaluation budget, seeds, and stopping rule are identical for MiniVLA and
$\pi_{0.5}$ so that the rollout comparisons use the same protocol.

\begin{table}[H]
\centering
\caption{Configuration for the simulation rollout evaluation.}
\label{tab:app_sim_rollout_protocol}
\small
\setlength{\tabcolsep}{5pt}
\renewcommand{\arraystretch}{1.08}
\begin{tabular}{lccccc}
\toprule
Backbone & Rollout limit & Episodes/task & Seeds & Episodes/seed & Action horizon \\
\midrule
MiniVLA & 350 steps & 225 & 7, 14, 21 & 75 & --- \\
$\pi_{0.5}$ & 350 steps & 225 & 7, 14, 21 & 75 & 10 actions \\
\bottomrule
\end{tabular}
\end{table}

\begin{table}[H]
\centering
\caption{MiniVLA rollout for Task 3, ``Pick up the ketchup and place it in the basket.''}
\label{tab:app_sim_minivla_head}
\scriptsize\setlength{\tabcolsep}{1pt}
\begin{tabular}{@{}>{\centering\arraybackslash}m{0.10\linewidth}*{8}{c}@{}}
\toprule
Method & \multicolumn{8}{c}{Simulation step} \\
\cmidrule(lr){2-9}
& 0 & 50 & 100 & 150 & 200 & 250 & 300 & 350 \\
\midrule
FULL & \raisebox{-0.5\height}{\includegraphics[width=0.087\linewidth]{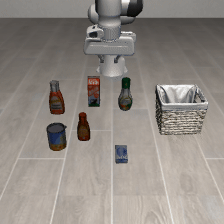}} & \raisebox{-0.5\height}{\includegraphics[width=0.087\linewidth]{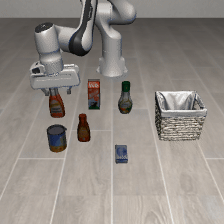}} & \raisebox{-0.5\height}{\includegraphics[width=0.087\linewidth]{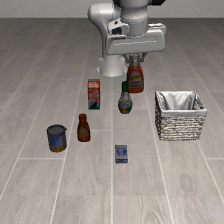}} & \raisebox{-0.5\height}{\includegraphics[width=0.087\linewidth]{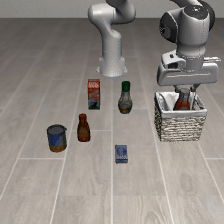}} & \raisebox{-0.5\height}{\includegraphics[width=0.087\linewidth]{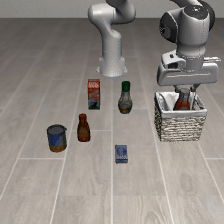}} & \raisebox{-0.5\height}{\includegraphics[width=0.087\linewidth]{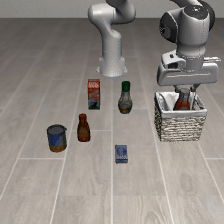}} & \raisebox{-0.5\height}{\includegraphics[width=0.087\linewidth]{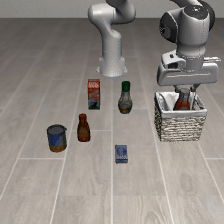}} & \raisebox{-0.5\height}{\includegraphics[width=0.087\linewidth]{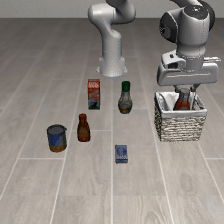}} \\
LT & \raisebox{-0.5\height}{\includegraphics[width=0.087\linewidth]{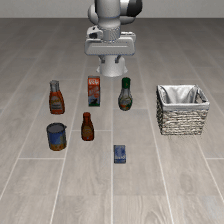}} & \raisebox{-0.5\height}{\includegraphics[width=0.087\linewidth]{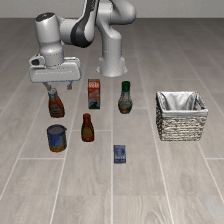}} & \raisebox{-0.5\height}{\includegraphics[width=0.087\linewidth]{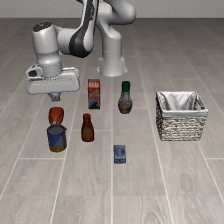}} & \raisebox{-0.5\height}{\includegraphics[width=0.087\linewidth]{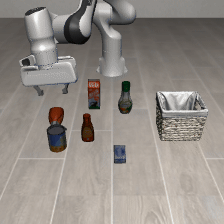}} & \raisebox{-0.5\height}{\includegraphics[width=0.087\linewidth]{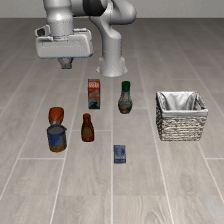}} & \raisebox{-0.5\height}{\includegraphics[width=0.087\linewidth]{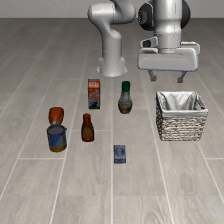}} & \raisebox{-0.5\height}{\includegraphics[width=0.087\linewidth]{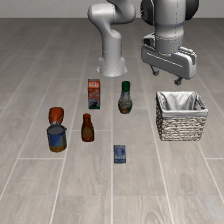}} & \raisebox{-0.5\height}{\includegraphics[width=0.087\linewidth]{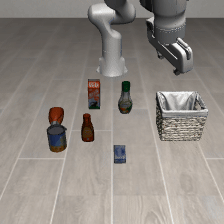}} \\
APA & \raisebox{-0.5\height}{\includegraphics[width=0.087\linewidth]{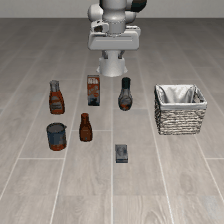}} & \raisebox{-0.5\height}{\includegraphics[width=0.087\linewidth]{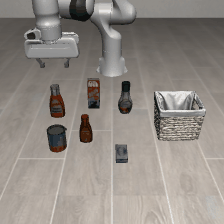}} & \raisebox{-0.5\height}{\includegraphics[width=0.087\linewidth]{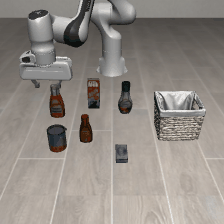}} & \raisebox{-0.5\height}{\includegraphics[width=0.087\linewidth]{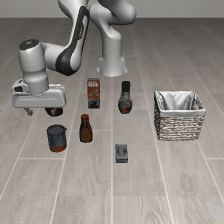}} & \raisebox{-0.5\height}{\includegraphics[width=0.087\linewidth]{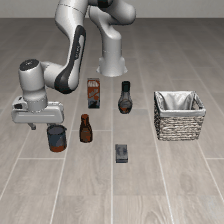}} & \raisebox{-0.5\height}{\includegraphics[width=0.087\linewidth]{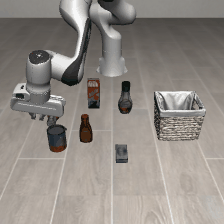}} & \raisebox{-0.5\height}{\includegraphics[width=0.087\linewidth]{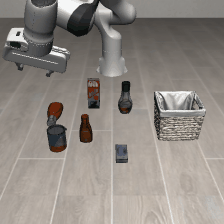}} & \raisebox{-0.5\height}{\includegraphics[width=0.087\linewidth]{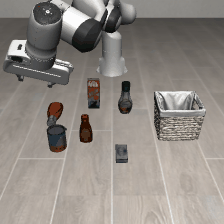}} \\
\methodname{} & \raisebox{-0.5\height}{\includegraphics[width=0.087\linewidth]{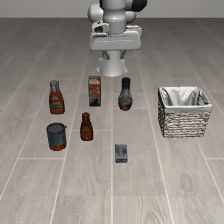}} & \raisebox{-0.5\height}{\includegraphics[width=0.087\linewidth]{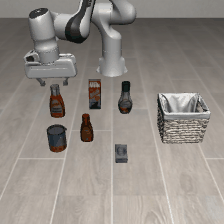}} & \raisebox{-0.5\height}{\includegraphics[width=0.087\linewidth]{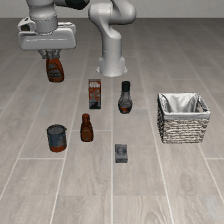}} & \raisebox{-0.5\height}{\includegraphics[width=0.087\linewidth]{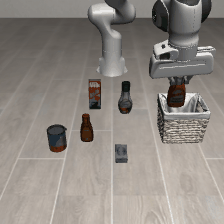}} & \raisebox{-0.5\height}{\includegraphics[width=0.087\linewidth]{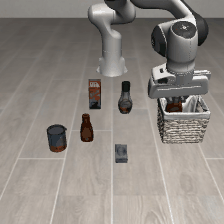}} & \raisebox{-0.5\height}{\includegraphics[width=0.087\linewidth]{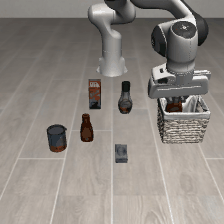}} & \raisebox{-0.5\height}{\includegraphics[width=0.087\linewidth]{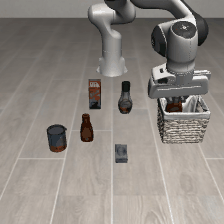}} & \raisebox{-0.5\height}{\includegraphics[width=0.087\linewidth]{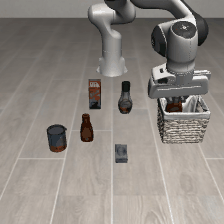}} \\
\bottomrule
\end{tabular}
\end{table}
\vspace{-3pt}
\noindent{\scriptsize\textit{Source:}
\href{https://github.com/Physical-Intelligence/openpi/blob/main/src/openpi/training/config.py}{Official OpenPI $\pi_{0.5}$ LIBERO training configuration}.}\par

\clearpage
\begin{table}[H]
\centering
\caption{MiniVLA rollout for Task 6, ``Push the plate to the front of the stove.''}
\label{tab:app_sim_minivla_tail}
\scriptsize\setlength{\tabcolsep}{1pt}
\begin{tabular}{@{}>{\centering\arraybackslash}m{0.12\linewidth}*{8}{c}@{}}
\toprule
Method & \multicolumn{8}{c}{Simulation step} \\
\cmidrule(lr){2-9}
& 0 & 50 & 100 & 150 & 200 & 250 & 300 & 350 \\
\midrule
FULL & \raisebox{-0.5\height}{\includegraphics[width=0.087\linewidth]{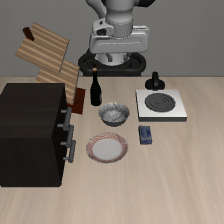}} & \raisebox{-0.5\height}{\includegraphics[width=0.087\linewidth]{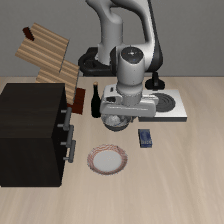}} & \raisebox{-0.5\height}{\includegraphics[width=0.087\linewidth]{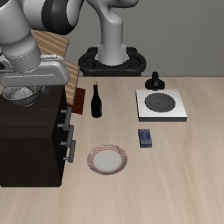}} & \raisebox{-0.5\height}{\includegraphics[width=0.087\linewidth]{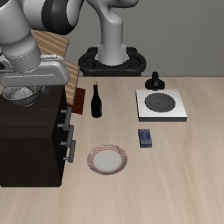}} & \raisebox{-0.5\height}{\includegraphics[width=0.087\linewidth]{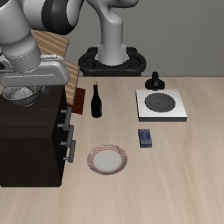}} & \raisebox{-0.5\height}{\includegraphics[width=0.087\linewidth]{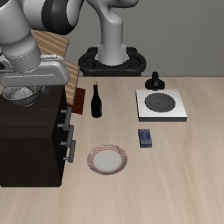}} & \raisebox{-0.5\height}{\includegraphics[width=0.087\linewidth]{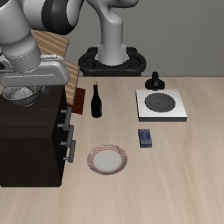}} & \raisebox{-0.5\height}{\includegraphics[width=0.087\linewidth]{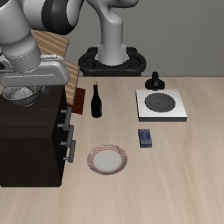}} \\
LT & \raisebox{-0.5\height}{\includegraphics[width=0.087\linewidth]{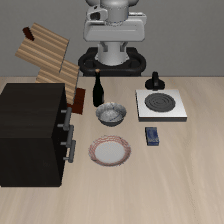}} & \raisebox{-0.5\height}{\includegraphics[width=0.087\linewidth]{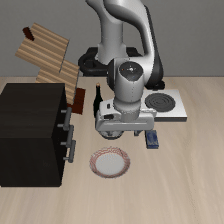}} & \raisebox{-0.5\height}{\includegraphics[width=0.087\linewidth]{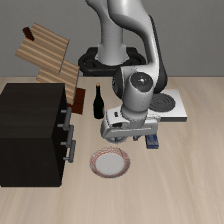}} & \raisebox{-0.5\height}{\includegraphics[width=0.087\linewidth]{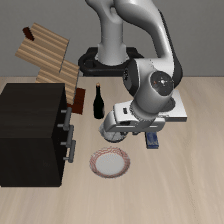}} & \raisebox{-0.5\height}{\includegraphics[width=0.087\linewidth]{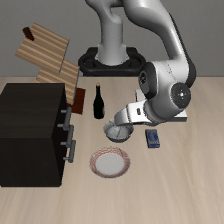}} & \raisebox{-0.5\height}{\includegraphics[width=0.087\linewidth]{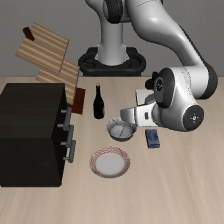}} & \raisebox{-0.5\height}{\includegraphics[width=0.087\linewidth]{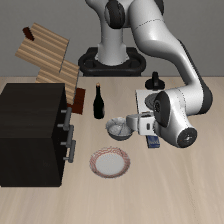}} & \raisebox{-0.5\height}{\includegraphics[width=0.087\linewidth]{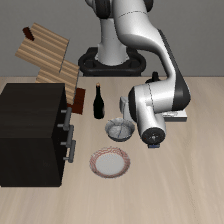}} \\
APA & \raisebox{-0.5\height}{\includegraphics[width=0.087\linewidth]{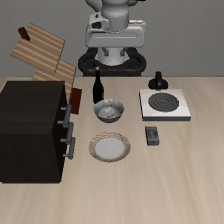}} & \raisebox{-0.5\height}{\includegraphics[width=0.087\linewidth]{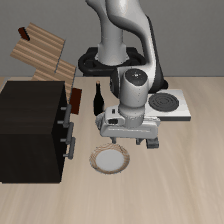}} & \raisebox{-0.5\height}{\includegraphics[width=0.087\linewidth]{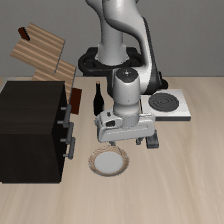}} & \raisebox{-0.5\height}{\includegraphics[width=0.087\linewidth]{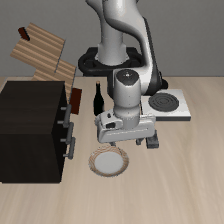}} & \raisebox{-0.5\height}{\includegraphics[width=0.087\linewidth]{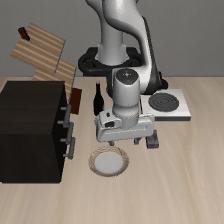}} & \raisebox{-0.5\height}{\includegraphics[width=0.087\linewidth]{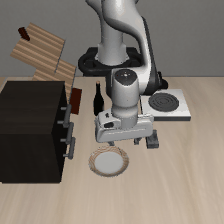}} & \raisebox{-0.5\height}{\includegraphics[width=0.087\linewidth]{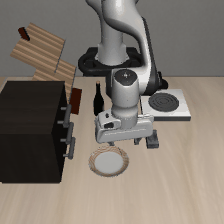}} & \raisebox{-0.5\height}{\includegraphics[width=0.087\linewidth]{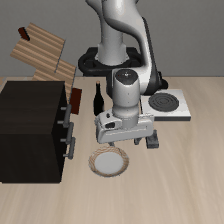}} \\
\methodname{} & \raisebox{-0.5\height}{\includegraphics[width=0.087\linewidth]{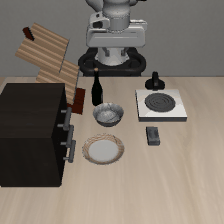}} & \raisebox{-0.5\height}{\includegraphics[width=0.087\linewidth]{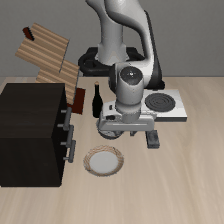}} & \raisebox{-0.5\height}{\includegraphics[width=0.087\linewidth]{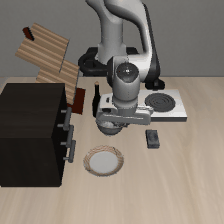}} & \raisebox{-0.5\height}{\includegraphics[width=0.087\linewidth]{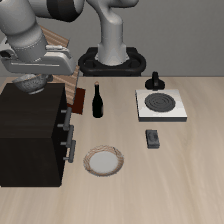}} & \raisebox{-0.5\height}{\includegraphics[width=0.087\linewidth]{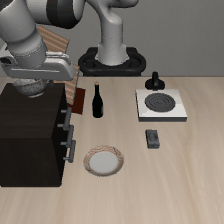}} & \raisebox{-0.5\height}{\includegraphics[width=0.087\linewidth]{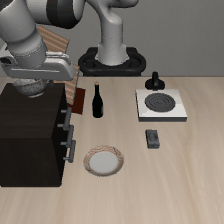}} & \raisebox{-0.5\height}{\includegraphics[width=0.087\linewidth]{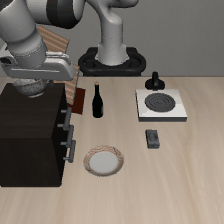}} & \raisebox{-0.5\height}{\includegraphics[width=0.087\linewidth]{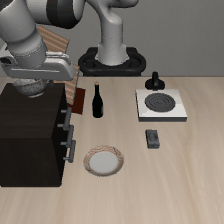}} \\
\bottomrule
\end{tabular}
\end{table}
\vspace{-8pt}

\begin{table}[H]
\centering
\caption{$\pi_{0.5}$ rollout for Task 8, ``Put the wine bottle on top of the cabinet.''}
\label{tab:app_sim_pi05_head}
\scriptsize\setlength{\tabcolsep}{1pt}
\begin{tabular}{@{}>{\centering\arraybackslash}m{0.12\linewidth}*{8}{c}@{}}
\toprule
Method & \multicolumn{8}{c}{Simulation step} \\
\cmidrule(lr){2-9}
& 0 & 50 & 100 & 150 & 200 & 250 & 300 & 350 \\
\midrule
FULL & \raisebox{-0.5\height}{\includegraphics[width=0.087\linewidth]{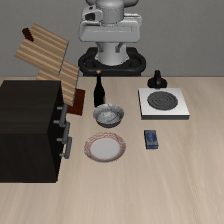}} & \raisebox{-0.5\height}{\includegraphics[width=0.087\linewidth]{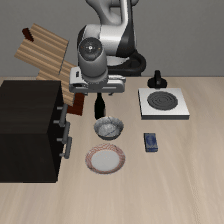}} & \raisebox{-0.5\height}{\includegraphics[width=0.087\linewidth]{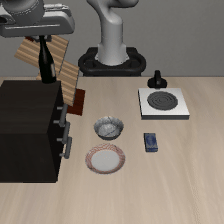}} & \raisebox{-0.5\height}{\includegraphics[width=0.087\linewidth]{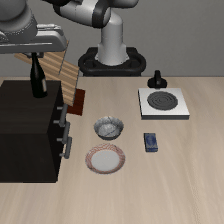}} & \raisebox{-0.5\height}{\includegraphics[width=0.087\linewidth]{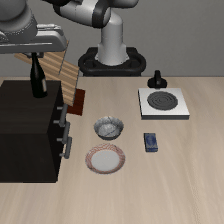}} & \raisebox{-0.5\height}{\includegraphics[width=0.087\linewidth]{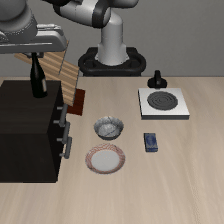}} & \raisebox{-0.5\height}{\includegraphics[width=0.087\linewidth]{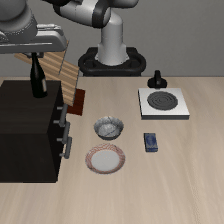}} & \raisebox{-0.5\height}{\includegraphics[width=0.087\linewidth]{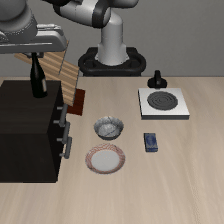}} \\
LT & \raisebox{-0.5\height}{\includegraphics[width=0.087\linewidth]{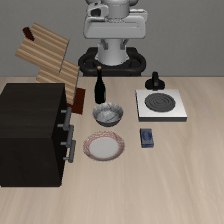}} & \raisebox{-0.5\height}{\includegraphics[width=0.087\linewidth]{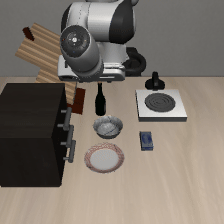}} & \raisebox{-0.5\height}{\includegraphics[width=0.087\linewidth]{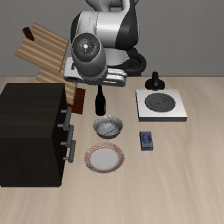}} & \raisebox{-0.5\height}{\includegraphics[width=0.087\linewidth]{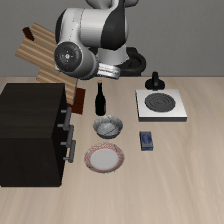}} & \raisebox{-0.5\height}{\includegraphics[width=0.087\linewidth]{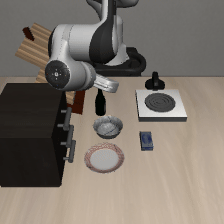}} & \raisebox{-0.5\height}{\includegraphics[width=0.087\linewidth]{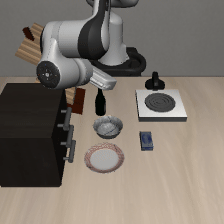}} & \raisebox{-0.5\height}{\includegraphics[width=0.087\linewidth]{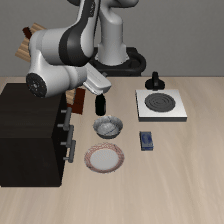}} & \raisebox{-0.5\height}{\includegraphics[width=0.087\linewidth]{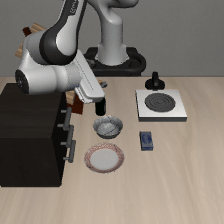}} \\
APA & \raisebox{-0.5\height}{\includegraphics[width=0.087\linewidth]{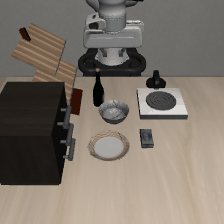}} & \raisebox{-0.5\height}{\includegraphics[width=0.087\linewidth]{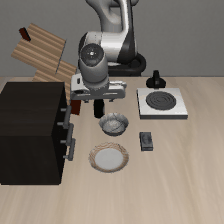}} & \raisebox{-0.5\height}{\includegraphics[width=0.087\linewidth]{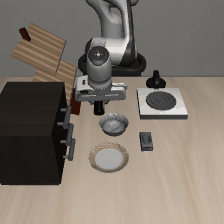}} & \raisebox{-0.5\height}{\includegraphics[width=0.087\linewidth]{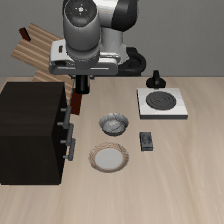}} & \raisebox{-0.5\height}{\includegraphics[width=0.087\linewidth]{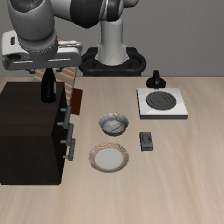}} & \raisebox{-0.5\height}{\includegraphics[width=0.087\linewidth]{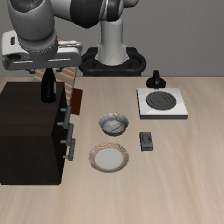}} & \raisebox{-0.5\height}{\includegraphics[width=0.087\linewidth]{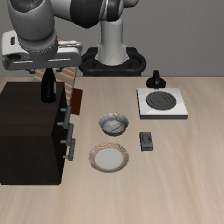}} & \raisebox{-0.5\height}{\includegraphics[width=0.087\linewidth]{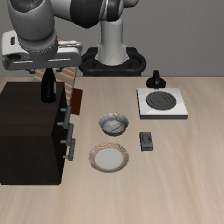}} \\
\methodname{} & \raisebox{-0.5\height}{\includegraphics[width=0.087\linewidth]{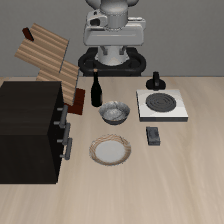}} & \raisebox{-0.5\height}{\includegraphics[width=0.087\linewidth]{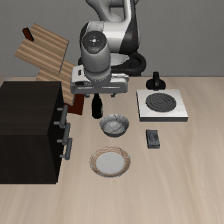}} & \raisebox{-0.5\height}{\includegraphics[width=0.087\linewidth]{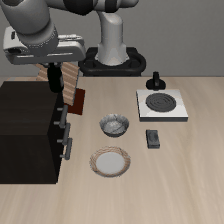}} & \raisebox{-0.5\height}{\includegraphics[width=0.087\linewidth]{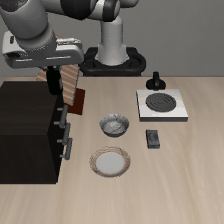}} & \raisebox{-0.5\height}{\includegraphics[width=0.087\linewidth]{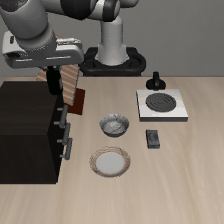}} & \raisebox{-0.5\height}{\includegraphics[width=0.087\linewidth]{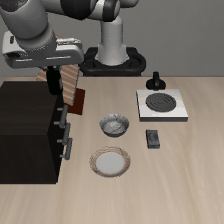}} & \raisebox{-0.5\height}{\includegraphics[width=0.087\linewidth]{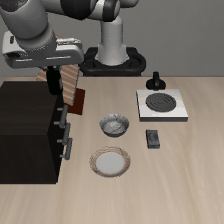}} & \raisebox{-0.5\height}{\includegraphics[width=0.087\linewidth]{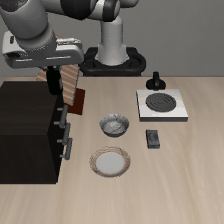}} \\
\bottomrule
\end{tabular}
\end{table}
\vspace{-8pt}

\begin{table}[H]
\centering
\caption{$\pi_{0.5}$ rollout for Task 9, ``Put the wine bottle on the rack.''}
\label{tab:app_sim_pi05_tail}
\scriptsize\setlength{\tabcolsep}{1pt}
\begin{tabular}{@{}>{\centering\arraybackslash}m{0.12\linewidth}*{8}{c}@{}}
\toprule
Method & \multicolumn{8}{c}{Simulation step} \\
\cmidrule(lr){2-9}
& 0 & 50 & 100 & 150 & 200 & 250 & 300 & 350 \\
\midrule
FULL & \raisebox{-0.5\height}{\includegraphics[width=0.087\linewidth]{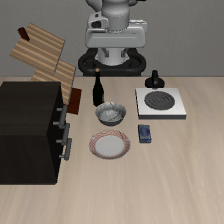}} & \raisebox{-0.5\height}{\includegraphics[width=0.087\linewidth]{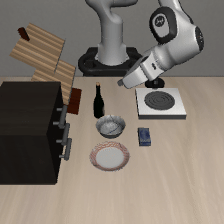}} & \raisebox{-0.5\height}{\includegraphics[width=0.087\linewidth]{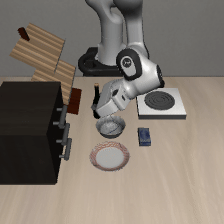}} & \raisebox{-0.5\height}{\includegraphics[width=0.087\linewidth]{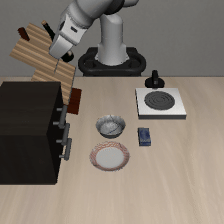}} & \raisebox{-0.5\height}{\includegraphics[width=0.087\linewidth]{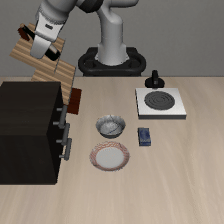}} & \raisebox{-0.5\height}{\includegraphics[width=0.087\linewidth]{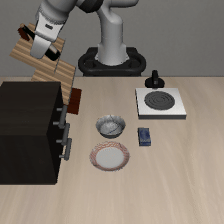}} & \raisebox{-0.5\height}{\includegraphics[width=0.087\linewidth]{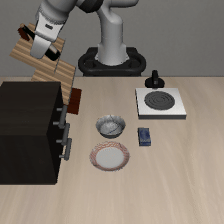}} & \raisebox{-0.5\height}{\includegraphics[width=0.087\linewidth]{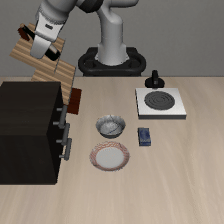}} \\
LT & \raisebox{-0.5\height}{\includegraphics[width=0.087\linewidth]{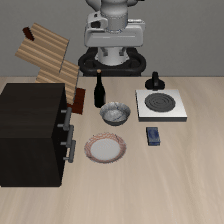}} & \raisebox{-0.5\height}{\includegraphics[width=0.087\linewidth]{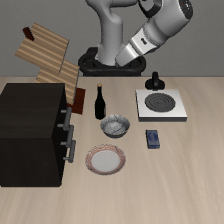}} & \raisebox{-0.5\height}{\includegraphics[width=0.087\linewidth]{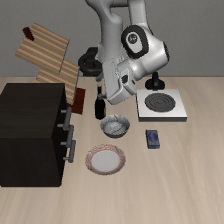}} & \raisebox{-0.5\height}{\includegraphics[width=0.087\linewidth]{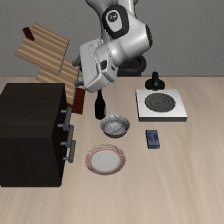}} & \raisebox{-0.5\height}{\includegraphics[width=0.087\linewidth]{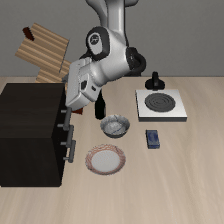}} & \raisebox{-0.5\height}{\includegraphics[width=0.087\linewidth]{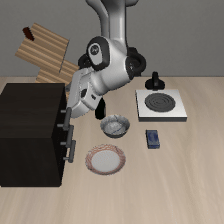}} & \raisebox{-0.5\height}{\includegraphics[width=0.087\linewidth]{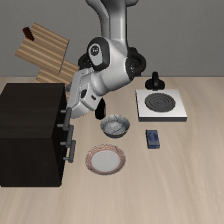}} & \raisebox{-0.5\height}{\includegraphics[width=0.087\linewidth]{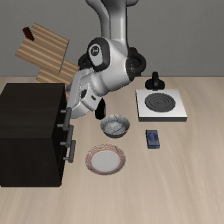}} \\
APA & \raisebox{-0.5\height}{\includegraphics[width=0.087\linewidth]{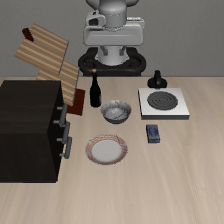}} & \raisebox{-0.5\height}{\includegraphics[width=0.087\linewidth]{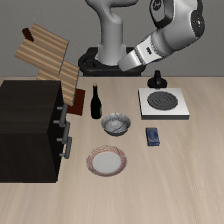}} & \raisebox{-0.5\height}{\includegraphics[width=0.087\linewidth]{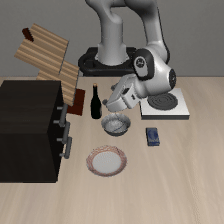}} & \raisebox{-0.5\height}{\includegraphics[width=0.087\linewidth]{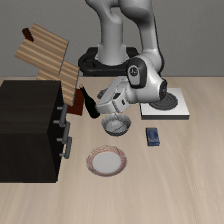}} & \raisebox{-0.5\height}{\includegraphics[width=0.087\linewidth]{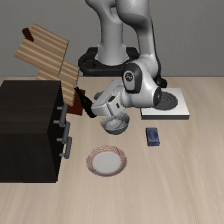}} & \raisebox{-0.5\height}{\includegraphics[width=0.087\linewidth]{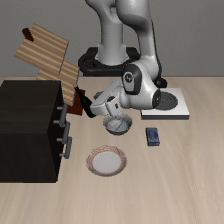}} & \raisebox{-0.5\height}{\includegraphics[width=0.087\linewidth]{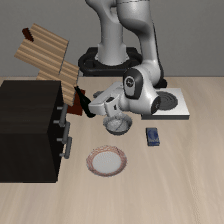}} & \raisebox{-0.5\height}{\includegraphics[width=0.087\linewidth]{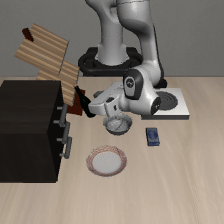}} \\
\methodname{} & \raisebox{-0.5\height}{\includegraphics[width=0.087\linewidth]{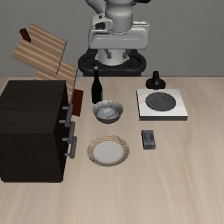}} & \raisebox{-0.5\height}{\includegraphics[width=0.087\linewidth]{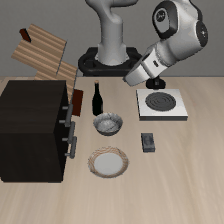}} & \raisebox{-0.5\height}{\includegraphics[width=0.087\linewidth]{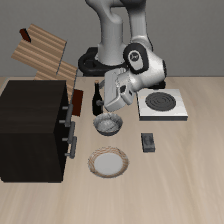}} & \raisebox{-0.5\height}{\includegraphics[width=0.087\linewidth]{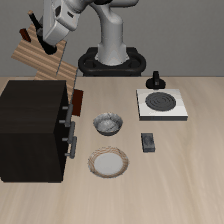}} & \raisebox{-0.5\height}{\includegraphics[width=0.087\linewidth]{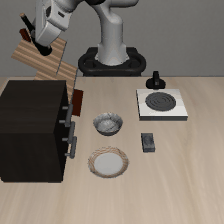}} & \raisebox{-0.5\height}{\includegraphics[width=0.087\linewidth]{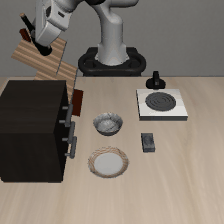}} & \raisebox{-0.5\height}{\includegraphics[width=0.087\linewidth]{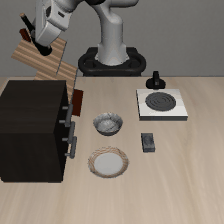}} & \raisebox{-0.5\height}{\includegraphics[width=0.087\linewidth]{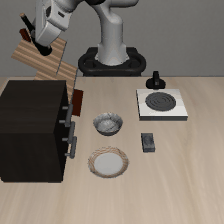}} \\
\bottomrule
\end{tabular}
\end{table}

\clearpage
\subsection{Real-World Rollouts}
\label{app:real_rollouts}

\paragraph{Evaluation configuration.}
We evaluate MiniVLA and $\pi_{0.5}$ separately on AGILEX PIPER and xArm6. Each
policy--robot pair is tested for 50 trials per task with randomized initial
object poses. PIPER uses synchronized global-camera and wrist-camera inputs,
whereas xArm6 uses a global camera. Consistent with the real-world study
in Section~\ref{sec:real_world_experiments}, this part compares only APA and
\methodname{}. Each real-world rollout uses eight frames sampled uniformly over
its normalized duration, from 0\% (start) to 100\% (final observation). This
axis supports comparison across robots even when their control rates differ.

\begin{table}[H]
\centering
\caption{Configuration for the real-world rollout evaluation.}
\label{tab:app_real_rollout_protocol}
\small
\setlength{\tabcolsep}{7pt}
\renewcommand{\arraystretch}{1.08}
\begin{tabular}{lccc}
\toprule
Robot & Visual observations & Backbones & Comparison / trials \\
\midrule
AGILEX PIPER & Global camera + wrist & MiniVLA, $\pi_{0.5}$ & APA vs. ECL / 50 per task \\
xArm6 & Global camera & MiniVLA, $\pi_{0.5}$ & APA vs. ECL / 50 per task \\
\bottomrule
\end{tabular}
\end{table}

\newcommand{\piperrollouttable}[3]{%
\begin{table}[H]
\centering
\caption{AGILEX PIPER rollout with #1 (global-camera and wrist-camera views).}
\label{#3}
\scriptsize\setlength{\tabcolsep}{1pt}
\renewcommand{\arraystretch}{0.90}
\begin{tabular}{@{}>{\centering\arraybackslash}m{0.085\linewidth}>{\centering\arraybackslash}m{0.10\linewidth}*{8}{c}@{}}
\toprule
Method & View & \multicolumn{8}{c}{Normalized rollout progress} \\
\cmidrule(lr){3-10}
& & $0\%$ & $14\%$ & $29\%$ & $43\%$ & $57\%$ & $71\%$ & $86\%$ & $100\%$ \\
\midrule
\multirow[c]{2}{*}[-0.022\textheight]{\textbf{APA}} & Global camera & \csname piperrow-#2-APA-third\endcsname \\
& Wrist & \csname piperrow-#2-APA-wrist\endcsname \\
\midrule
\multirow[c]{2}{*}[-0.022\textheight]{\textbf{\methodname{}}} & Global camera & \csname piperrow-#2-ECL-third\endcsname \\
& Wrist & \csname piperrow-#2-ECL-wrist\endcsname \\
\bottomrule
\end{tabular}
\end{table}}

\newcommand{\xarmrollouttable}[3]{%
\begin{table}[H]
\centering
\caption{xArm6 rollout comparison with #1.}
\label{#3}
\scriptsize\setlength{\tabcolsep}{1pt}
\renewcommand{\arraystretch}{0.90}
\begin{tabular}{@{}>{\centering\arraybackslash}m{0.12\linewidth}*{8}{c}@{}}
\toprule
Method & \multicolumn{8}{c}{Normalized rollout progress} \\
\cmidrule(lr){2-9}
& $0\%$ & $14\%$ & $29\%$ & $43\%$ & $57\%$ & $71\%$ & $86\%$ & $100\%$ \\
\midrule
\textbf{APA} & \csname xarmrow-#2-APA\endcsname \\
\textbf{\methodname{}} & \csname xarmrow-#2-ECL\endcsname \\
\bottomrule
\end{tabular}
\end{table}}

\subsubsection{AGILEX PIPER}
\piperrollouttable{MiniVLA}{minivla}{tab:app_real_piper_minivla}
\piperrollouttable{$\pi_{0.5}$}{pi05}{tab:app_real_piper_pi05}

\begin{figure}[H]
\centering
\includegraphics[width=0.86\linewidth]{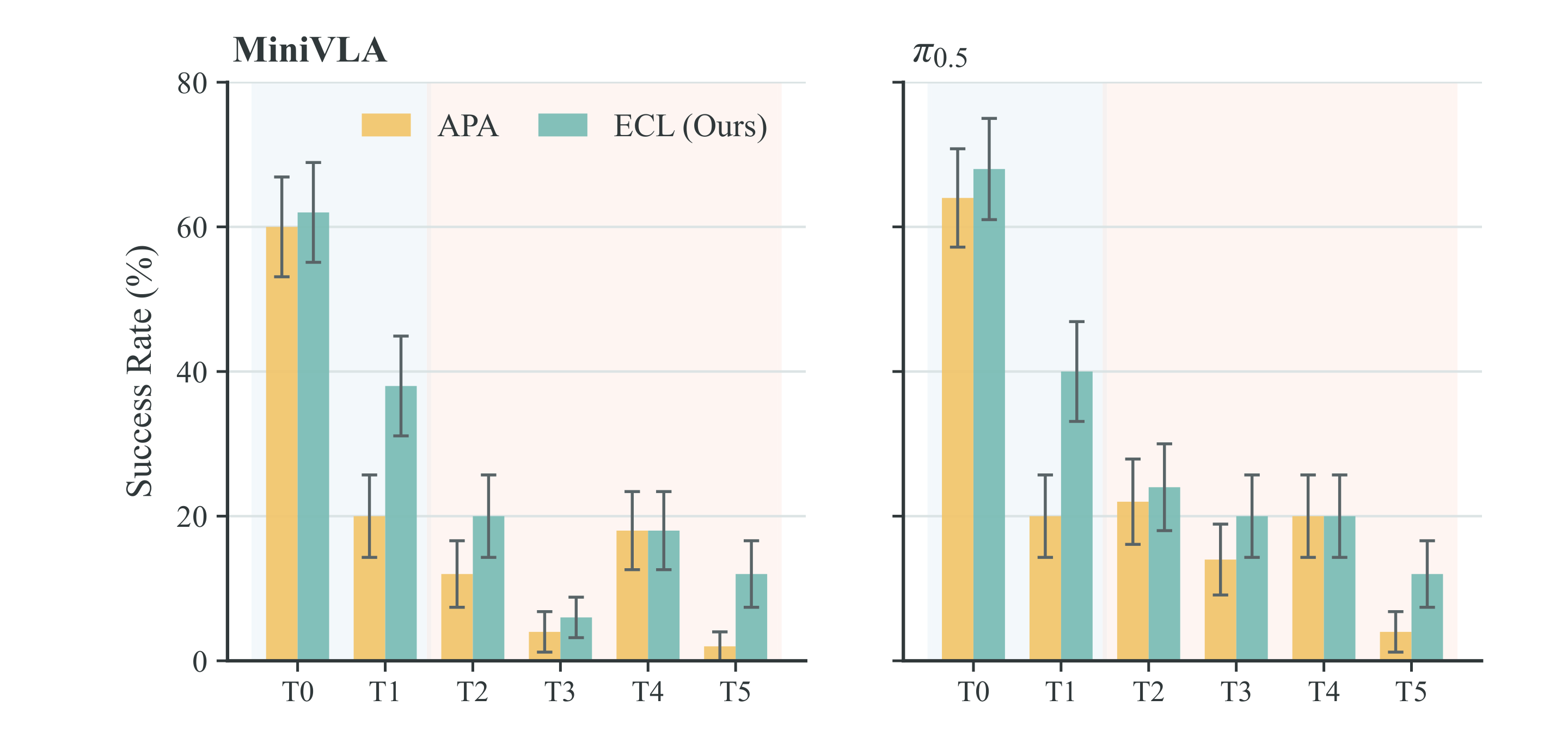}
\caption{xArm6 per-task success rates for MiniVLA and $\pi_{0.5}$, comparing
APA with \methodname{}.}
\label{fig:app_xarm_success}
\end{figure}

\subsubsection{xArm6}
\xarmrollouttable{MiniVLA}{minivla}{tab:app_real_xarm_minivla}

\xarmrollouttable{$\pi_{0.5}$}{pi05}{tab:app_real_xarm_pi05}

Figure~\ref{fig:app_xarm_success} reports the per-task success rates on xArm6.
Both xArm6 rollout tables use the same time axis as the PIPER tables. The final version will pair each rollout with its task instruction and outcome.
These examples illustrate controller behavior; the quantitative claims remain
based on the complete success-rate evaluation.

\clearpage

\section{Comparison of Mask Selection Strategies}
\label{app:mask_selection_visualization}
\begin{figure}[H]
\centering
\includegraphics[width=\linewidth]{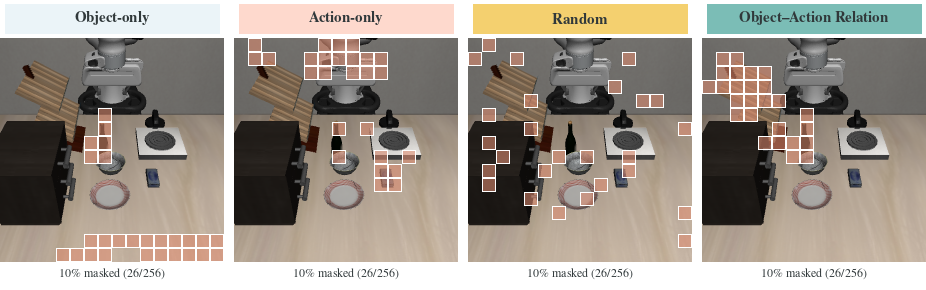}
\caption{\textbf{Illustrative comparison of mask selection strategies.}
From left to right: object-only, action-only, random, and object--action
relation selection. Peach overlays with white borders mark selected
patches; all panels select 26/256 patches. The supplied visualization uses
semantic-mask and saved-attention proxies, as described below, rather
than a direct export of the complete \methodname{} selector.}
\label{fig:app_mask_selection}
\end{figure}

Figure~\ref{fig:app_mask_selection} compares four patch-selection strategies on the same observation from the task ``put the wine bottle on the rack.'' Each panel selects 26 of the $16\times16=256$ visual patches, corresponding to the nominal 10\% masking budget. Keeping the observation and patch budget fixed makes the spatial differences between selection rules visible.

\textbf{Reading the comparison.} Object-only selection uses target-object coverage; action-only selection uses a saved patch-attention map; random selection samples patches without task information. The relational illustration combines object coverage and attention, with an additional goal-region contribution. Consequently, random patches are dispersed across the image, while the relational panel includes regions around the bottle and rack. These panels illustrate why the spatial location of a fixed masking budget matters when constructing an evidence-limited reference.

\section{Effectiveness of Visual Evidence in ECL}
\label{app:ecl_visual_evidence}

The purpose of this analysis is to verify whether \methodname{} can learn the
correct visual evidence for under-represented tail tasks. As discussed in the
Introduction, tail-task failures are closely associated with weak
task-specific visual grounding: without reliable evidence for the current
task, a shared policy is more likely to rely on action patterns learned from
frequent head tasks. We therefore visualize two LT tail tasks---``put the bowl
on top of the cabinet'' and ``put the wine bottle on the rack''---and compare
the LT baseline with \methodname{}.

\begin{figure}[H]
\centering
\setlength{\fboxsep}{0pt}
\begin{tabular}{@{}c@{\hspace{0.03\textwidth}}c@{}}
\includegraphics[width=0.20\textwidth]{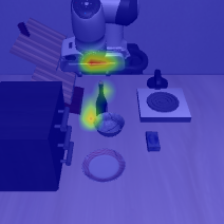} &
\includegraphics[width=0.20\textwidth]{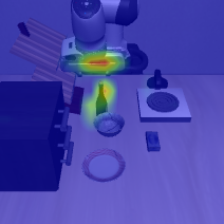}
\\
\noalign{\vskip 0.03\textwidth}
\includegraphics[width=0.20\textwidth]{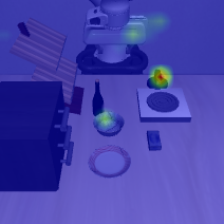} &
\includegraphics[width=0.20\textwidth]{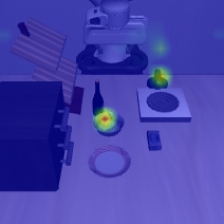}
\end{tabular}\par\vspace{0.7em}
\caption{\textbf{Visual evidence on representative LT tail tasks.} Each square
shows a paired comparison with the LT baseline on the left and ECL on the
right. The top row shows ``put the bowl on top of the cabinet,'' and the bottom
row shows ``put the wine bottle on the rack.'' Relative to the LT baseline, ECL more
consistently grounds its decisions in instruction-relevant objects,
interaction regions, and goal locations, supporting our finding that ECL
effectively learns the correct task-specific visual evidence for tail-task
decision making.}
\label{fig:app_ecl_visual_evidence}
\end{figure}

Figure~\ref{fig:app_ecl_visual_evidence} shows that the LT baseline often
assigns attention to the workspace background or regions unrelated to the
instruction. In contrast, \methodname{} consistently focuses on the visual
evidence required to complete the task, including the instructed object, the
end-effector--object interaction, and the target location. These results show
that \methodname{} does more than improve the final success rate: it enables
the policy to learn task-specific visual evidence from scarce tail-task
demonstrations and use that evidence to support action prediction.

This is the central motivation for \emph{Evidence-Guided Counterfactual
Learning}. Instruction- and action-related signals identify the visual
evidence that guides the counterfactual intervention. Training on the resulting
intact--counterfactual difference then strengthens the association among the
instruction, the correct visual evidence, and the predicted action. In this
way, ECL explicitly encourages tail tasks to recover reliable visual grounding,
which also reduces their dependence on head-task action bias.

\section{Training Cost Details}

\textbf{Recorded training timings.} Table~\ref{tab:app_compute_training} summarizes the supplied training logs. For a log containing $M$ timing records $t_j$, we report the arithmetic mean $\bar{t}=M^{-1}\sum_{j=1}^{M}t_j$, the median, and the accumulated recorded time $T_{\mathrm{log}}=\sum_{j=1}^{M}t_j/3600$ in hours. All valid records are retained, including initialization and unusually slow steps. The timer fields are \texttt{VLA Train/Step Time} for FULL, LT, and APA, and \texttt{seconds} for \methodname{}. Their measurement boundaries cannot be verified from the logs alone, so $T_{\mathrm{log}}$ is not interpreted as end-to-end training time: unrecorded data preparation, checkpointing, evaluation, and restart intervals are not added.

\begin{table}[H]
\centering
\caption{Recorded MiniVLA training cost. GPU counts and batch sizes are taken
from saved configurations. Batch denotes per-device/global batch size.
Logged hours cover only the supplied timing records, which span different
training intervals; these rows are not a controlled overhead comparison.}
\label{tab:app_compute_training}
\small
\setlength{\tabcolsep}{3.2pt}
\begin{tabular}{l r r c r r r r}
\toprule
Method & Data size & GPUs & Batch & Logged steps & Mean (s) & Median (s) & Logged (h) \\
\midrule
FULL & 55,966 & 1 & 20/20 & 100,738 & 0.692 & 0.660 & 18.41 \\
LT (Baseline) & 19,234 & 1 & 20/20 & 34,622 & 0.608 & 0.570 & 5.89 \\
APA & 21,049 & 1 & 20/20 & 37,888 & 0.673 & 0.651 & 7.21 \\
\methodname{} (Ours) & 19,234 & 1 & 20/20 & 34,622 & 1.068 & 1.090 & 10.27 \\
\bottomrule
\end{tabular}
\end{table}

\textbf{Cost of the counterfactual branch.} Within the \methodname{} run, the first 1,923 steps have no active counterfactual samples, consistent with the configured two-epoch warm-up. Their mean recorded time is 0.519\,s/step. The subsequent 32,698 steps activate the counterfactual branch and average 1.100\,s/step, with a median of 1.090\,s/step. The ratio of these phase means is approximately $2.12\times$. This is a within-run observation across different training phases, not a matched measurement against the LT baseline. These operations accompany the original and masked predictions and a separate no-gradient selector pass in the training procedure; their individual timing contributions are not separately logged.

\noindent\begin{minipage}{\linewidth}
\textbf{Data volume, training time, and performance trade-off.} The LT baseline
uses 19,234 samples and has the lowest mean step time (0.608\,s). APA expands
the dataset to 21,049 samples, an increase of 1,815 samples ($9.4\%$), while
also increasing the logged steps from 34,622 to 37,888 and the mean step time
to 0.673\,s. Consequently, its recorded training time rises from 5.89 to 7.21
hours. In addition, producing the augmented APA samples requires object
detection, image inpainting, and data validation; this time-consuming data
construction is not included in the logged hours reported in
Table~\ref{tab:app_compute_training}. ECL uses exactly the same 19,234 samples
and 34,622 logged steps as LT, but its mean step time increases to 1.068\,s and
its recorded training time to 10.27 hours. This increase is explained by the
additional intact and counterfactual predictions and the no-gradient evidence
selection performed during training. Although ECL therefore incurs a higher
per-step training cost, it requires no additional demonstrations or augmented
data, and the additional computation is confined to training. Considering the
success-rate improvements in Table~\ref{tab:per_task_success}, the substantial
time required to construct APA's extra data, and the absence of additional
deployment-time computation, ECL provides the preferable overall balance
between policy performance and training cost.
\end{minipage}

\end{document}